\RequirePackage[2020-02-02]{latexrelease}
\documentclass[]{clv3}
\usepackage{hyperref}
\usepackage{xcolor}
\usepackage{graphicx}
\usepackage{latexsym}
\usepackage{url}
\usepackage[T1]{fontenc}

\usepackage{svg}
\usepackage{amsmath}
\usepackage{comment}
\usepackage{booktabs}
\usepackage{color,soul}
\usepackage{float}
\usepackage{savesym}

\savesymbol{numdef}
\usepackage{csquotes}
\restoresymbol{CSQ}{numdef}

\usepackage{lipsum}
\usepackage{tabularx}
\usepackage[toc,page]{appendix}

\usepackage[pass]{geometry}

\MakeOuterQuote{"}
\definecolor{darkblue}{rgb}{0, 0, 0.5}
\hypersetup{colorlinks=true,citecolor=darkblue, linkcolor=darkblue,urlcolor=darkblue}
\newcommand{\aiscomment}[3]{{\textcolor{#3}{[#1 #2]}}}

\newcommand{\newcite}[1]{\citet{#1}}

\newcommand{\enull}{\mbox{\tt NULL}}
\newcommand{\commentout}[1]{}

\runningtitle{} 
\runningauthor{} 
\jname{}
\jinfo{}

\title{Measuring Attribution in Natural Language Generation Models}

\author{Hannah Rashkin\thanks{\ \ Equal contribution. All authors contributed to all parts of the paper. $\spadesuit$ Led development of the conceptual framework. $\clubsuit$ Led human annotation study. $\diamondsuit$ Contributed to modeling experiments. $\heartsuit$ Provided project leadership and management.
E-mail: \{hrashkin,vitalyn,mrlamm,loraa,mjcollins,dipanjand,slav,gtomar,reitter\}@google.com, iulia@iuliaturc.com}$^{, \clubsuit \diamondsuit}$}
\affil{Google Research}
\author{Vitaly Nikolaev$^{*, \clubsuit \spadesuit}$}
\affil{Google Research}
\author{{Matthew Lamm}$^\spadesuit$}
\affil{Google Research}
\author{{Lora Aroyo}$^\spadesuit$}
\affil{Google Research}
\author{{Michael Collins}$^\spadesuit$}
\affil{Google Research}
\author{{Dipanjan Das}$^{\spadesuit \heartsuit}$}
\affil{Google Research}
\author{{Slav Petrov}$^\heartsuit$}
\affil{Google Research}
\author{{Gaurav Singh Tomar}$^\diamondsuit$}
\affil{Google Research}
\author{{Iulia Turc}$^\diamondsuit$}
\affil{Google Research}
\author{{David Reitter}$^{\spadesuit \heartsuit}$}
\affil{Google Research}

\date{}

\begin{document}
\maketitle
\begin{abstract}
With recent improvements in natural language generation (NLG) models for various applications, it has become imperative to have the means to identify and evaluate whether NLG output is only sharing verifiable information about the external world.
In this work, we present a new evaluation framework entitled \emph{Attributable to Identified Sources} (AIS) for assessing the output of natural language generation models, when such output pertains to the external world. We first define AIS and introduce a two-stage annotation pipeline for allowing annotators to appropriately evaluate model output according to AIS guidelines.
We empirically validate this approach on  generation datasets spanning three tasks (two conversational QA datasets, a summarization dataset, and a table-to-text dataset) via human evaluation studies that suggest that AIS could serve as a common framework for measuring whether model-generated statements are supported by underlying sources.
We release guidelines for the human evaluation studies. 
\end{abstract}

\section{Introduction}\label{sec:intro}
Large, pretrained neural models have advanced Natural Language Generation (NLG) performance across a variety of use cases, including text summarization, translation, and dialogue.
Yet, generative neural models are  known to hallucinate often, lacking faithfulness to underlying sources, for example in summarization or in grounded dialogue systems. Accurate evaluation with respect to these issues is important.

In this paper, we develop a framework for the evaluation of {\em attribution}, by which we mean the accurate use of source documents to support generated text. Attribution is closely related to issues of hallucination and faithfulness (see \S\ref{sec:background} for discussion). As a key motivating example, consider a dialog with a system that generates responses to a user’s sequence of questions:

\begin{small}
\begin{tabbing}
\hspace{0.2cm} \= \textsc{User:} what was George Harrison’s first solo album?\\
\> \textsc{System:} \= it was “Wonderwall Music”, released in November 1968.\\
\> \textsc{User:} how old was he when it was released?\\
\> \textsc{System:} he was 25 years old
\end{tabbing}
\end{small}
\noindent

If such a system, in addition to generating responses, could attribute its statements to source documents, that is, provide sufficient and concise evidence for its claims, system designers and users alike could more readily ascertain the extent to which the information it provides is supported by underlying sources.
Prior work in NLG spanning diverse use cases such as summarization, dialogue response generation and data-to-text generation have investigated issues of faithfulness and  “hallucination”, but have not provided a uniform and formally expressed framework to measure these errors.  We discuss the relationship of our work to related work in \S\ref{sec:background}.

In \S\ref{sec:definitions}, we introduce our evaluation framework, Attributable to Identified Sources (AIS), that can be used to assess whether statements in natural language made by a system are derivable from a given underlying source. 
The definition of AIS (see \S\ref{sec:ais-definition}) formalizes the meaning of a sentence $s$ in context using the notion of explicatures~\cite{carston1988explicature,wilson2002relevance}\footnote{For example, in the above dialogue the explicature of “he was 25 years old” is “George Harrison was 25 years old when `Wonderwall Music' was released”: the latter explicature is evaluated for attribution. Note that this use of explicatures is closely related to prior work on decontextualization \cite{choi2021decontext}, see  \S\ref{sec:background} for more discussion.}, and defines attribution to some background information source $P$ in terms of an intuitive test, asking whether ``According to \textit{P}, \textit{s}''. 
It also accommodates system outputs whose meaning is uninterpretable.
AIS can be used as a pre-condition or in tandem with other metrics or evaluation frameworks to assess overall quality.
For example, characteristics of the underlying source (such as ``source quality''), the fluency of the generated text, and so forth, can be measured using complementary metrics that are out of scope in this work.

We propose specific instantiations of AIS for three NLG tasks (\S\ref{sec:human-study}): 
response generation in a conversational QA setting (as in the example above; responses must be attributable to a provided answer document),
text summarization (where the summary must be attributable to the source article), and description generation from structured tables, or table-to-text (where the description must be attributable to the source table and associated metadata).
Each domain involves a number of challenges: for example, in dialogue systems a key challenge is that the meaning of system responses is highly contextually dependent.

Next, we establish the feasibility of AIS evaluations via an empirical study through conducting human evaluation experiments.
We train annotators to evaluate output text from multiple models per task using task-specific instantiations of AIS.
We show that in our human evaluation studies, it is possible to achieve a moderate to high degree of inter-annotator agreement (see \S\ref{sec:human-study} for more details).  We're also able to observe differences in model outputs' AIS scores, following generally expected trends.
As part of this work, we release the detailed guidelines for human evaluation.
We believe that AIS as a framework would be essential for the evaluation of system-generated utterances across NLG tasks.
\section{Background}\label{sec:background}
\textbf{Hallucations in NLG.} As alluded to in \S\ref{sec:intro}, past work has identified the issue of hallucination in neural generation models.
\newcite{wiseman-etal-2017} presented challenges in data-to-text generation where neural models generate hallucinated content not supported by source data;  they proposed an automatic information extraction-based metric to evaluate generated text for that particular scenario and conducted a small human evaluation study examining whether summaries are supported by source data.
More recently, \newcite{parikh-etal-2020-totto} presented a larger human evaluation study in the context of a data-to-text generation dataset entitled ToTTo, and measured hallucinations in terms of \textit{faithfulness} with respect to a source data table.

Hallucination has been a salient subject of investigation in text summarization. \newcite{maynez2020faithfulness} presented an extensive characterization of hallucinations, discussed behavior of models that generate content that are present in larger corpora beyond a given source and conducted a significant human study. Additional automatic QA-based methods for detecting hallucinations have been proposed by \newcite{wang-etal-2020-asking}, \newcite{nan-etal-2021-improving}, among others.
One of the most relevant papers, \newcite{durmus-etal-2020-feqa}, involved both a human evaluation and the introduction of an automatic question--answer based evaluation method. Their human evaluation of summary sentences is similar to our two-stage annotation pipeline where they evaluate sentences in two steps --- first for whether it is understandable, and second, if so, for faithfulness to the underlying source (their instructions to annotators are: ``If the information conveyed by the sentence is not expressed in the source, select `unfaithful'.'').

In the case of response generation for dialogue, especially in scenarios that involve the system responding about the real world, research has focused on measuring the responses' consistency to prior conversational history or their groundedness to some external evidence, that we deem to be very close to the topic of hallucination.
These have been measured via dialogue-specific natural language inference methods, often via human studies and data creation \cite{Welleck2019DialogueNL,mehri-eskenazi-2020-usr,Gupta2021DialFactAB,Honovich2021Q2EF,Dziri2021EvaluatingGI,santhanam2021rome}.

Despite a significant amount of work pertaining to hallucination spanning multiple NLG problems, there is no unified approach to evaluate whether system generated statements are supported by underlying source documents.
Human evaluation studies are varied from paper to paper and detailed, reproducible annotation instructions are unavailable \cite{belz2020disentangling}. Likewise, the use of terminology for describing and defining evaluation criteria also lacks consistency and further complicates reproducibility \cite{howcroft2020twenty}.
General-purpose benchmarking across these tasks have gained traction \cite{gehrmann-etal-2021-gem}, but there has not been a standardized treatment of the attribution problem.
Our paper attempts to address this gap by explicitly formalizing the evaluation of attribution as a replicable and extendable conceptual framework.  As part of our definition of attribution, we outline a more formal background for ``information conveyed by the text'' --- in particular through the use of explicatures (see Figure~\ref{fig:explicatures} for examples). Lastly, we demonstrate that AIS can be generalized across multiple NLG tasks in which context, source documents, and generated text can take different forms.

\noindent\textbf{Fact Verification.} A related field of study has dealt with the topic of fact or claim verification \cite[\textit{inter alia}]{thorne-etal-2018-fever,Thorne2018AutomatedFC,DBLP:journals/corr/abs-2104-00640}.
Work in this area has framed the task as retrieving supporting evidence given a claim, and optionally classifying semantic relationships between the claim and the evidence text.
Modeling approaches have overlapped with recent literature examining natural language inference \cite{DBLP:conf/aaai/NieCB19}.
\citet{DBLP:journals/corr/abs-2104-00640} have examined several human annotation tasks for the above family of problems;  however, there are several key differences with this work. First, we evaluate the quality of a system generated utterance with respect to given evidence source (a fundamentally different end goal); we utilize the notion of explicatures in defining attribution; finally, we avoid absolute judgments regarding ``factuality'' of utterances.
As mentioned in \S\ref{sec:intro}, rather than making  factuality judgments, we deem that complementary evaluation methods such as ``source quality'' in tandem with AIS would be required to evaluate the factuality of utterances.
As a corollary, we assume the source is a reference, and that an actual system may select sources for their trustworthiness.

\noindent \textbf{Decontextualization.} \citet{choi2021decontext} introduce the task of decontextualization, that is, the problem of taking a sentence in context and rewriting it in a way that it's meaning is preserved, and it can be interpreted out of context. This is directly related to the idea of explicatures, which are also used in the current paper. 

\section{A Formal Definition of Attributable to Identified Sources}
\label{sec:definitions}

This section gives a formal definition of AIS, attempting to give a clear and precise definition of attribution. We  first  give  a  definition  of  AIS  for  a  simple case, where the utterance from a system is a standalone proposition.  In spite of the simplicity of this  setting,  it  is  highly  informative,  and  forms the  basis  for  the  full  definition  of  AIS. We then describe  how  this  definition extends to a much larger set of system utterances, in particular giving a treatment of \emph{interpretability}\footnote{We acknowledge that the term "interpretability" has come to signify "model interpretability" in the NLP and ML community (as established in \citet{harrington1985harvey}, \citet{ribeiro2016model}). The term in our use represents how interpretable system output is for a human annotator. The choice of terminology is intended to be more conceptually transparent when used by annotators: unlike other terms like "meaningful"/"nonsensical" \cite{durmus-etal-2020-feqa}, or "sensibleness"~\cite{adiwardana2020towards}, "interpretability" more readily alludes to the significance of the propositions in system generated output in relationship to context. Finally, the annotators are typically not familiar with the "model interpretability" usage of the term.}, and \emph{contextual effects}. A key idea in our model of meaning in context is the notion of explicatures~\cite{carston1988explicature,wilson2002relevance,choi2021decontext}. In a final subsection, we describe how key aspects of the AIS definition naturally lend it to operationalization, while also pointing out how certain idealizations (e.g., the notion of a ``generic speaker”) must be relaxed to accommodate the practical realities of implementation.

\subsection{An Initial Definition of AIS: Attribution of Standalone Propositions}

We now give a definition of AIS for a simple but important case, where the text in question is a {\em standalone proposition}. We in general assume a setting where AIS is to be determined for a string whose meaning is ascertained relative to a context. In the following treatment we assume that time is the only non-linguistic aspect of context relevant to determining textual meaning, modeling a setting where two generic speakers communicate over a text-based channel, with no additional prior information about each other.\footnote{Extensions of AIS to more complex settings may require a more elaborate notion of non-linguistic context.}

We define standalone propositions as follows:

\begin{definition}[Standalone Propositions]
A standalone proposition is a declarative sentence that is interpretable once a time $t$ has been specified.
\end{definition}
\noindent 
To illustrate the definition of standalone propositions, consider the following examples:
\begin{description}
\item[Example S1:] George Harrison was 25 years old when his album `Wonderwall Music' was released.

\item[Example S2:] He was 25 years old.

\item[Example S3:] George Harrison was 25 years old.

\item[Example S4:] George Harrison died over 15 years ago.
\end{description}

\noindent 
All four examples are declarative sentences. S1 is a standalone proposition. S4 is a standalone proposition, as it is interpretable once the time $t$ is specified. 
S2 is however not a standalone proposition, as it cannot be interpreted without additional contextual information: It is unclear what "He" refers to.
More subtly, S3 is also not a standalone proposition, because it lacks details of historical context.

The definition of AIS for standalone propositions is as follows:

\begin{definition}[AIS for standalone propositions]
A pair $(s, t)$ consisting of a standalone proposition $s$ and a time $t$ is Attributable to Identified Sources (AIS) iff the following conditions hold:
\begin{enumerate}
    \item The system provides a set of parts $P$ of some underlying corpus $K$, along with $s$.
    \item $(s, t)$ is attributable to $P$. 
\end{enumerate}

\noindent
A pair $(s, t)$ is {\bf attributable} to  a set of parts $P$ of some underlying corpus $K$ iff:
A generic hearer will, with a chosen level of confidence, affirm the following statement:
 “According to $P$, $s$'', where $s$ is interpreted relative to time $t$.
\end{definition}

\noindent 
Here, the corpus $K$ could be a set of web pages, and the parts $P$ could be pointers to paragraphs or sentences within $K$; or the corpus $K$ could be a knowledge graph, with $P$ as parts of the underlying knowledge graph; other examples are no doubt possible. 

As an example, consider standalone proposition S1 given above, assume that the corpus $K$ is all of Wikipedia, $t_0$ is the present time (specifically, noon on December 21st 2021), and assume that the set $P$ consists of a single paragraph from Wikipedia, as follows:

\begin{description}
\item[Example P1:] George Harrison (25 February 1943 --- 29 November 2001) was an English musician, singer--songwriter, and music and film producer who achieved international fame as the lead guitarist of the Beatles. His debut solo album was
`Wonderwall Music', released in November 1968.
\end{description}
\noindent 
Under this definition, it would be correct for a hearer to judge ``$(S1, t_0)$ is attributable to P1'', because the ``according to'' test in the AIS definition holds. That is, it is reasonable to say according to P1, S1'' where S1 is interpreted at time $t_0$: ``according to P1, George Harrison was 25 years old when his album `Wonderwall Music' was released.''

Note that in some cases the system may provide multiple parts. The standalone proposition $S$ may also be justified by certain forms of multi-hop reasoning (e.g., arithmetic processes) over that set of parts. The above example requires reasoning about dates and age.

\begin{table*}
    \small
    \caption{AIS examples illustrating challenges in AIS judgements. These types of examples may be difficult to assess for AIS because they need extra reasoning, or assumptions about shared knowledge.  These examples are purely illustrative (not from real data examples).}
    \begin{tabularx}{\textwidth}{X X X}
    \toprule
    Evidence & Proposition Candidate & Challenges\\ 
    \midrule
George   Harrison   (25   February 1943  —  29 November 2001) was  an  English musician\dots  His debut solo album was `Wonderwall Music’, released in November 1968. & George Harrison was 25 years old when his album 'Wonderwall Music' was released. & \textit{Common sense and cultural knowledge is required to interpret the information in the proposition as it requires inferring that ``his'' is still referring to George Harrison. "George" is typically a male name in English; musicians release albums; therefore, "his album" likely refers to George Harrison, but not another unattested entity.} \\
    \midrule
The runtime of the theatrical edition of ``The Fellowship of the Ring'' is 178 minutes, the runtime of ``The Two Towers'' is 179 minutes, and the runtime of ``The Return of the King'' is 201 minutes. & The full run-time of ``The Lord of the Rings'' trilogy is 558 minutes. & \textit{Evaluating this requires numerical reasoning, and it also requires knowing that ``The Lord of the Rings'' trilogy consists of the three films mentioned (background knowledge that may vary from person to person).  Additionally, it requires assumptions that the runtime is consistently referring to the theatrical edition of these movies.} \\
    \bottomrule
    \end{tabularx}
    \label{tab:implicit_examples}
\end{table*}

\subsection{Extending AIS: Attribution of Sentences in Context}
We now extend the previous definition of AIS to cover sentences that that go beyond standalone propositions. To do so, we will need to consider multi-sentence cases, and cases with non-empty linguistic contexts. We will also cover cases that are uninterpretable. 

We first define the notion of ``utterance'':
\begin{definition}[Utterance]
An utterance is a sequence of one or more {\em sentences} produced by a system or user, where a sentence may be a declarative, a question, a command, an exclamation, or a fragment. The $i$th system utterance is  $s_i = s_{i,1} \ldots s_{i, |s_i|}$, where $s_{i,j}$ is the $j$th sentence within system utterance $s_i$, and similarly the $i$th user utterance is $u_i = u_{i,1} \ldots u_{i, |u_i|}$.
\end{definition}

To briefly illustrate our approach to non-empty linguistic contexts, consider the following interaction between a user and system (originally given in the introduction; repeated here for convenience):

\begin{small}
\begin{tabbing}
\hspace{0.2cm} \= $u_1$: what was George Harrison’s first solo album?\\
\> $s_1$: \= it was “Wonderwall Music”, released in November 1968.\\
\> $u_2$: how old was he when it was released?\\
\> $s_2$: he was 25 years old.
\end{tabbing}
\end{small}

\noindent
The system utterance $s_2 = \hbox{\em he was 25 years old}$ is clearly not a standalone proposition. As such, it cannot be evaluated for AIS given our previous definition. However, given the previous context in the interaction, intuitively the meaning of $s_2$ is something similar to the standalone proposition ``George Harrison was 25 years old when his album ``Wonderwall Music'' was released''. This latter ``paraphrase'' of $s_2$'s meaning is a standalone proposition, and can be evaluated using the AIS definition for standalone propositions.

We will make this notion of ``paraphrase'' of the meaning of an utterance in context more formal, through the introduction of {\em explicatures}. The explicature of $s_2$ in context of the previous utterances $u_1, s_1, u_2$ is $e$ = {\em George Harrison was 25 years old when his album ``Wonderwall Music'' was released}. Once explicatures have been defined in this way, they can be evaluated for AIS in exactly the same way as standalone propositions.

\subsubsection{Definition of Interactions and Linguistic Context}

We will use the following definition of {\em interaction} throughout the paper:
\begin{definition}[Interactions]
An {\em interaction}  consists of: 1) a sequence $u_1 \ldots u_m$ of $m \geq 0$ user utterances; 2) a sequence $s_1 \ldots s_n$ of $n \geq 0$ system utterances; 3) a strict total order over the $m + n$ user and system utterances.\footnote{For example, the order might be specified by functions $U: \{1 \ldots m\} \rightarrow \{1 \ldots (m + n)\}$ and $S: \{1 \ldots n\} \ldots (m+n)\}$ where $U(i)$ (respectively $S(i)$) is the position of utterance $u_i$ (respectively $s_i$) in the total ordering. The notational details will not be important for this paper.
}
\end{definition}

\noindent 
This setting is intended to be quite general, including a broad class of applications where systems generate utterances. In conversational QA systems we typically have alternating user and system utterances, where $m = n$, and the total ordering is $u_1, s_1, u_2, s_2, \ldots u_n, s_n$. In summarization tasks we have a simplified setting where $m = 0$, $n = 1$, and $s_1$ is equal to the summary generated by the system. Table-to-text tasks are similar to summarization in that $m = 0$, $n = 1$, while $s_1$ is the description of the table generated by the system.

Each sentence has an associated linguistic context:

\begin{definition}[Linguistic Context for Sentences]
We define the linguistic context for system sentence $s_{i,j}$ to be $c_{i,j}$, where $c_{i,j}$ is the ordered sequence of sentences (with speaker identities, user or system) that precedes $s_{i,j}$ in the total ordering. 
We define the linguistic context for user sentence $u_{i,j}$ to be $c'_{i,j}$, where $c'$ is defined in a similar way.\footnote{An equally plausible definition would be to define $c_{i,j}$ to also include the following sentences within utterance $s_i$, that is, $s_{i,j-1}, s_{i,j+1} \ldots s_{i, |s_i|}$ (and an analogous definition for $c'_{i,j}$). That is, the context would be extended to include sentences that follow $s_{i,j}$ in the utterance $s_{i}$. This would allow instances of cataphora, for example, to be handled in the definitions of explicatures and attribution.}
\end{definition}

\noindent 
Here the definition of "sentence" is intended to be quite broad. A sentence could be a declarative sentence, a question, or a fragment (such as the string "25 years old"). Under the above definitions, the context for a user or system sentence is
simply the sequence of user and system sentences that precedes it. To illustrate these definitions consider the following example:

\begin{small}
\begin{tabbing}
\hspace{0.2cm} \= $u_1$: what was George Harrison’s first solo album?\\
\> $s_1$: \= it was “Wonderwall Music”, released in November 1968.\\
\> $u_2$: how old was he when it was released?\\
\> $s_2$: He was 25 years old. It was the first solo album by a member of the beatles.
\end{tabbing}
\end{small}

\noindent
Here the system utterance $s_2$ consists of two sentences, $s_{2,1} =$ {\em He was 25 years old} and $s_{2,2} =$ {\em It was the first solo album by a member of the Beatles}.

\subsection{Explicatures}\label{sec:explicatures}

A key goal in this section is to define AIS for sentences $s_{i,j}$ in linguistic contexts $c_{i,j}$ which are non-empty (i.e., which contain previous sentences in the discourse). To do this it will be critical to formalize what is meant intuitively by "the meaning of $s_{i, j}$ in context $c_{i, j}$". To do this we introduce {\em explicatures} (this definition is closely related to definition 1 in \citet{choi2021decontext}):

\begin{definition}[Explicatures]
Define the context $c$ to be $(c_l, t)$, where $c_l$ is the linguistic context and $t$ is the time. Define  $\bar{c}$ to be the context $(\epsilon, t)$ where $\epsilon$ is the linguistically empty context: that is, $\bar{c}$ is a copy of $c$ but with $c_l$ replaced by $\epsilon$.
The set of {\em explicatures} $E(c, x)$ of a sentence $x$ in a context $c$ is a set  that satisfies the following conditions: 1) each $e \in E(c, x)$ is a declarative sentence or question that is interpretable in context $\bar{c}$; 2) each $e \in E(c, x)$ has the same truth-conditional meaning in $\bar{c}$ as the meaning of sentence $x$ in context $c$. 
\end{definition}
\noindent 
Note that the sentence $x$ will most often in this paper be a system sentence $s_{i, j}$ in linguistic context $c_{i,j}$, but can also be a user sentence $u_{i, j}$ in linguistic context $c'_{i,j}$.

Thus, each $e \in E(c, x)$ is a paraphrase of $x$ that is interpretable in the linguistically empty context and that preserves the truth-conditional meaning of $x$ in context $c$. Note that $E(c, x)$ is a set because there may be multiple ways of paraphrasing $x$, which are equivalent in meaning. Given an equivalence relation between sentences that identifies whether any two sentences are equal in meaning or not, we can think of a single member of $E(c, x)$ as a representative of the entire set $E(c, x)$. Following this, in a slight abuse of terminology we will henceforth often write "the explicature of $x$ in context $c$ is $e$" as if there is a single unique explicature $e$, with the understanding that $e$ represents the entire set $E(c, x)$. We will also write $E(c, x) = e$ as shorthand for $E(c, x)$ being equal to the set of all sentences whose meaning is the same as that of $e$.

In addition, we define interpretability as follows:
\begin{definition}[Interpretability]
A sentence $x$ in context $c$ is {\em uninterpretable} if the truth-conditional meaning of $x$ in context $c$ is unclear. In this case we write $E(c, x) = \enull$.
\end{definition}

Figure~\ref{fig:explicatures} shows several examples illustrating these definitions. Some key points are as follows:

\hspace{1ex}

\noindent
{\em Remark 1:}
    In example E1, the system response is a direct answer to a question, $s_{2,1} = $ {\em 25 years old}. $s_{2, 1}$ itself is not a declarative sentence, but given the context (in particular the question it is answering), its explicature is the standalone proposition {\em George Harrison was 25 years old when "Wonderwall Music" was released}. This type of example --- where a direct answer to a question is an entity, noun-phrase, or some other fragment, but its explicature is a standalone proposition --- is important and frequent. As another example consider the following:
    
\begin{tabbing}
\hspace{0.2cm} \= $u_1$: What was George Harrison’s first solo album?\\
\> $s_1$: Wonderwall Music\\
\>$E(c_{1,1}, s_{1,1}) = $ \= {\em George Harrison's first solo album em was "Wonderwall Music"}
\end{tabbing}

\noindent
{\em Remark 2:} In Example E3, the system segment is a sequence of two declarative sentences. Each sentence has an explicature that is a standalone proposition. This type of case is again frequent and important.

\hspace{1ex}

\noindent
{\em Remark 3:} In Example E4 the system utterance is uninterpretable, because it is not clear what "the band" is referring to. Example E5 contains disfluencies that make it difficult to reliably interpret: "it" is not the expected pronominal reference; in this context "25" becomes too ambiguous to interpret as referring to the age of a human entity. 

\hspace{1ex}

\noindent
{\em Remark 4:} Examples E6 and E7 contain questions in the system and user utterance respectively. These examples illustrate that single questions (E7) or questions within multi-sentence utterances (E6) have well-defined explicatures.

\begin{figure}[!t]
\begin{small}
\begin{tabbing}
{\bf Example E1}\\
\hspace{0.2cm} \= $u_1$: what was George Harrison’s first solo album?\\
\> $s_1$: it was “Wonderwall Music”, released in July 2006.\\
\> $u_2$: how old was he when it was released?\\
\> $s_2$: 25 years old\\
\> $E(c_{2,1}, s_{2,1})$ = \= {\em George Harrison was 25 years old when "Wonderwall Music" was released}
\end{tabbing}

\begin{tabbing}
{\bf Example E2}\\
\hspace{0.2cm} \= $u_1, s_1, u_2$ as in Example E1\\
\> $s_2$: he was 25 years old\\
\> $E(c_{2,1}, s_{2,1})$ = \= {\em George Harrison was 25 years old \em when "Wonderwall Music" was released}
\end{tabbing}

\begin{tabbing}
{\bf Example E3}\\
\hspace{0.2cm} \= $u_1, s_1, u_2$ as in Example E1\\
\> \=$s_2$: He was 25 years old. It was the first solo \\
\> \> album by a member of the Beatles.\\
\> $E(c_{2,2}, s_{2,2})$ = \=  {\em Wonderwall Music was the first solo album by a member of the Beatles.}
\end{tabbing}

\begin{tabbing}
{\bf Example E4}\\
\hspace{0.2cm} \= $u_1, s_1, u_2$ as in Example E1\\
\> $s_2$: the band was The Beatles\\
\> $E(c_{2,2}, s_{2,2})$ = \= $\enull$
\end{tabbing}

\begin{tabbing}
{\bf Example E5}\\
\hspace{0.2cm} \= $u_1, s_1, u_2$ as in Example E1\\
\> $s_2$: it was 25\\
\> $E(c_{2,2}, s_{2,2})$ = \= $\enull$
\end{tabbing}

\begin{tabbing}
{\bf Example E6}\\
\hspace{0.2cm} \= $u_1, s_1, u_2$ as in Example E1\\
\> \=$s_2$: He was 25 years old. Have you heard that album? \\
\> $E(c_{2,2}, s_{2,2})$ = \= {\em Have you heard the album "Wonderwall Music"?}
\end{tabbing}

\begin{tabbing}
{\bf Example E7}\\
\hspace{0.2cm} \= $u_1$: what was George Harrison’s first solo album?\\
\> $s_1$: it was “Wonderwall Music”, released in July 2006.\\
\> $u_2$: how old was he when it was released?\\
\> $E(c'_2, u_2)$ = \= {\em how old was George Harrison when "Wonderwall Music" was released?}
\end{tabbing}

\end{small}
\caption{Examples of utterances in context, and their explicatures.}
\label{fig:explicatures}
\end{figure}

\label{sec:ais}

\subsubsection{The Full Definition of AIS}
\label{sec:ais-definition}

With this background, we can now give the full definition of AIS:

\begin{definition}[AIS, full definition]
\label{defn:ais-full}
A pair $(s, c)$, where $s$ is a sentence and $c=(c_{l}, t)$ is a pair consisting of a linguistic context and a time, is {Attributable to Identified Sources (AIS)} iff the following conditions hold:
\begin{enumerate}
    \item The system provides a set of parts $P$ of some underlying corpus $K$, along with $s$.
    \item $s$ in the context $c$ is interpretable (i.e., $E(c, s) \neq \enull$).
    \item The explicature $E(c, s)$ is a standalone proposition.
    \item The pair $(E(c, s), t)$ is attributable to $P$. 
\end{enumerate}

\noindent
The pair $(E(c, s), t)$ is {\bf attributable} to  a set of parts $P$ of some underlying corpus $K$ iff:
A generic hearer will, with a chosen level of confidence, affirm the following statement:
 “According to $P$, $E(c, s)$”, where $E(c, s)$ is interpreted relative to time $t$.
\end{definition}
\noindent 
The definition is very similar to the earlier definition of AIS for standalone propositions, but with checks for interpretability, and with attribution applied to explicatures of system sentences. Note that AIS can only hold for system sentences that have an explicature that is a standalone proposition  (condition 3). For example, the explicature in Example E6 in Figure~\ref{fig:explicatures} is not a standalone proposition, as it is a question. We leave the treatment of cases such as these to future work (we might for example evaluate attribution for declarative sentences within the explicature, excluding questions; or we might evaluate presuppositions within the questions themselves).

\subsubsection{Attribution of Entire Utterances}

In the previous sections we have described AIS for the individual sentences $s_{i,1} \ldots s_{i,|s_i|}$ within a system utterance $s_i$. This assumes that such a segmentation of the utterance into sentences is available, for example, it is provided by the system. An alternative is to evaluate entire utterances $s_i$ for AIS, in a "single-shot" annotation. AIS applied at the utterance level could potentially have the advantages of simplicity, and the avoidance of segmenting utterances into sentence boundaries. It has the potential disadvantage of being coarser grained, not allowing AIS judgments at the sentence level. The choice of sentence-level vs. utterance-level AIS will depend on the exact application of AIS.

It should be relatively straightforward to extend the full definition of AIS (Section~\ref{sec:ais-definition}) to apply to multi-sentence utterances. The definition of explicatures would need to be extended to multi-sentence utterances; the definition of standalone propositions would also have to be extended to apply to multiple sentences; the definition of "attributable" would also need to be extended.

\subsection{Towards Operationalization of AIS}
\label{sec:operationalization}
In the above definition of AIS, three definitions are of key importance: 1) the     ``according to'' test for standalone propositions; 2) the definition of interpretability; 3) the definition of explicatures, which are related to the interpretation of utterances in non-empty linguistic contexts. 
Note that it is not necessary for annotators to explicitly wield all of these definitions, or come to understand any of them in entirely formal terms, in order to provide AIS judgments. 
In developing human annotation guidelines for annotators who do not necessarily have background in these concepts, we relay the ``according to'' test and interpretability in a way that leverages natural speaker intuitions. We convey explicature through the more intuitive idea of a sentence paraphrase with respect to linguistic context.
Annotators are instructed to apply the ``according to'' test strictly, without making further assumptions beyond what is conveyed in the text. 

The formal definition of AIS makes several idealizing assumptions that must be relaxed in practical settings. 
In lieu of the posited ``generic hearer'', the judgments of actual annotators will naturally be influenced by the particulars of their interpretive capacities, stemming from differences, for example, in cultural background and domain expertise.
Table~\ref{tab:implicit_examples} lists several instances where such differences could conceivably affect judgments.
These effects are, to some extent, inherent to implementing AIS using human judgments.
\section{Human Evaluation Study}
\label{sec:human-study}
We evaluate the feasibility of human AIS assessment for three NLG tasks: conversational question answering, summarization, and table-to-text generation. To quantify the significance of human judgements, we present evaluators with the output of different models for each of the tasks.

The set-up for these annotation tasks is to ask annotators to rate the AIS quality of $s$, some model produced output given some attributed source $P$.  In the conversational QA and summarization settings, $P$ is a document or passage from a document, while in the table-to-text setting $P$ is a table and its description. For conversational QA, annotators are also provided with a context $c$, which is the set of previous conversation turns.  $c$ is used to help annotators understand the contextualized meaning of the model output, what we formally define as explicature in \S\ref{sec:explicatures}.

Because this is a challenging task with many possible edge cases (such as those discussed in Table~\ref{tab:implicit_examples}), we ask five annotators to judge each example.  In our results section, we compare to the consensus answer (if there is one) for simplicity.  In future work, researchers who wish to use AIS for evaluating systems might find use in distinguishing between cases that are more clear cut (i.e., unanimous) versus those where there may be some inherent ambiguity.

\subsection{Task Design}
\label{sec:task-design}
We break the annotation task into two stages described in \S\ref{sec:ann:int} and \S\ref{sec:ann:ais}, which mirrors the formal steps in the AIS definition (\S\ref{sec:ais-definition}) First, the annotators are asked if they are able to understand and identify the information being shared in the model output without seeing the source document (i.e., whether it is {\it interpretable} on its own).  Then, if the output is deemed interpretable, the annotators are shown the ``attributed source'' \textit{P} and asked whether all of the information that is shared in $S$ can be attributed to $P$ (i.e., whether it is {\it AIS}).  As described in the results sections, the splitting of the task into these two steps helps annotators to first filter out outputs that are badly formed (e.g., ungrammatical to the point of impeded intelligibility) or too ambiguous (e.g., unclear pronouns) to appropriately evaluate the attribution.  In the results, we report scores based on the annotator consensus (i.e., majority vote): the percent of total examples marked as interpretable ({\it Int} in Tables) and the percent of interpretable examples that were marked as AIS ({\it AIS}). In some datasets, certain examples were flagged as difficult to annotate due to legibility-related issues (see \S\ref{sec:ann:flag}).  For those cases, we separately report the percentage of examples that were flagged ({\it Flag}) and thus excluded from the interpretability and AIS scores.

\subsubsection{Interpretability Rating}
\label{sec:ann:int}
In the initial stage of the annotation task, we show the annotators the model output $s$ and any preceding context $c$ without showing the source. 
We ask them to identify the interpretability by posing a yes/no question. For example, in the summarization task the annotators are asked:

\emph{Is \textbf{all} of the information relayed by the system summary interpretable to you?}

Note that context $c$ is populated with preceding turns of the system--user interaction\footnote{Some interactions may contain no previous turns.} in the conversational QA task, whereas in summarization and table-to-text tasks it is always empty. In the instructions, the context $c$ is explicitly called out to be used in interpreting output $s$ in the conversational QA task.

Here, the goal is to tease out if the model-generated output $s$ contains any potential ambiguity that would prevent or misguide establishing attribution to its source $P$. Anaphora resolution is the main source of this type of ambiguity, where deictic elements do not have clear antecedents within $s$ or its context  --- for example, pronominal usage with an unclear or broken coreference chain or definite noun phrases as first mentions. Additionally, syntactic ambiguity or disfluency may also result in diminished interpretability of $s$ (see Examples E4, E5 in Figure~\ref{fig:explicatures}).

We acknowledge potential anthropomorphizing effects on how annotators interpret the system output \cite{gopnik1992child}. Because cooperative meaning co-construction between interlocutors is the default communicative strategy of inter-human interaction \cite{grice1975logic}, when faced with ambiguities and slight discrepancies in the system output, annotators may be ``forgiving'' of diminished interpretability, especially if the underlying source is present and can help recover missing context.

In our experiments we have found that not presenting the source at this stage is crucial for ensuring that evaluators are strict in their assessment of interpretability of the system output 
(see Figures~\ref{fig:AIS_QA_UI_1}, \ref{fig:AIS_Summ_UI_1}, and \ref{fig:AIS_T2T_UI_1} for how it was implemented in the task interface).

\subsubsection{AIS Rating}
\label{sec:ann:ais}
If an annotator selects ``yes'' for the interpretability question, we show them the source $P$ and ask them whether \textbf{all} of the information relayed in the output $s$ can be supported by $P$. For example, in the conversational QA task the annotators are asked:

\emph{Is \textbf{all} of the information provided by the system response {\it fully} supported by the source document?}

Note that the $P$ for the conversational QA task is the retrieved document that serves as the source of the system output $s$. In the summarization task $P$ is the original news article from which the summary in $s$ was derived. In the table-to-text task $P$ is the original table, highlighted cells, and table metadata (table title,  section title, and section text) from which the textual table description is generated.

In the instructions, we tell annotators to first think about all of the information that is contained the output including: what's directly stated in the output sentence verbatim as well as any explicatures that can be made from the output with respect to the context, such as inferring pronoun references from the conversational history.

Annotators are instructed to only mark output as attributable if it is clear that all parts can be directly inferred from the source. The instructions specifically call out to utilize the paraphrase test:

\emph{In determining this question, ask yourself whether it is accurate to say “the provided news article says\dots” or “according to the news article\dots” with the system summary following this phrase.}

If the output is misrepresenting information from the source because it is misleadingly worded, missing important context, or even changing only slight details, these cases are all counted as ``not fully attributable''.

\subsubsection{Flag Rating}
\label{sec:ann:flag}
A special rating is reserved for flagging items that would be disqualified from the the task altogether because they flout the range of possible relationships between the utterance, its context, and the source defined in \ref{sec:ais-definition}.

In practical terms, these are tasks that are too malformed for annotators to perform judgements on. This category includes tasks with rendering issues in the interface (missing task elements, e.g., empty utterance), corrupted text resulting in non-communicative utterances (bad text encoding, HTML artifacts), underspecified source (the source document itself is ambiguous because it is too short and may contain unresolved reference chains), or a source that is difficult to understand because it requires expert-level knowledge.

Once a task is flagged, it is disqualified from the rating queue of the annotator who flagged it. Other annotators may choose not to flag this item; cumulative ratings and interannotator agreements are calculated for all non-flagged ratings of a task (see the flag sections in the annotator guidelines for \href{sec:instructions_qa_flag}{conversational QA}, \href{sec:instructions_summ_flag}{summarization}, and \href{sec:instructions_t2t_flag}{table-to-text}) Section~\ref{sec:instructions_t2t_flag}.

\subsubsection{Limitations}
By asking yes/no questions, we can greatly reduce the complexity of this task for annotators.  However, for some applications of AIS measures, it may be useful to have more fine-grained measures.  Additionally, we ask annotators to evaluate the entire output (rather than sentences or specific spans) under the reasoning that if even one span within the model output is not AIS, then the whole output is not AIS (cf. \citet{maynez2020faithfulness}, \citet{durmus-etal-2020-feqa}).

We also acknowledge that there are other aspects of model output quality (e.g., relevance, non-redundancy, etc.) not evaluated here.  We focus on the separate evaluation of AIS as part of a focused effort towards quantifying the attribution itself, disentangled from other desirable generation qualities.

\subsection{Human Evaluation Procedure}
The ratings were performed by a group of nine paid full-time annotators under the guidance and supervision of a project manager. The annotator team is based in Hyderabad, India; the annotators are native speakers of the Indian dialect of the English language. The annotators do not have a background in linguistics. They were trained for this specific task.

Three separate user interfaces were developed for performing the evaluation in this study: one for the conversational QA tasks evaluating the output of models trained on QReCC and WoW datasets,  another for summarization tasks evaluating the output of models trained on the CNN/DM dataset and lastly one for table-to-text tasks evaluating the output of models trained on the ToTTo dataset. The interfaces share many fundamental design elements with task dependent modifications. For example, the conversation QA interface contains a devoted element for displaying the conversational history. All three interfaces explicitly hide the source document/table at the stage when interpretability of the system output is evaluated (see the Appendix for the interface layouts and annotator prompts (Figures~\ref{fig:AIS_QA_UI_1}, \ref{fig:AIS_QA_UI_2}, \ref{fig:AIS_Summ_UI_1}, \ref{fig:AIS_Summ_UI_2}, \ref{fig:AIS_T2T_UI_1}, and \ref{fig:AIS_T2T_UI_2}).

The annotators were trained on the tasks in a series of stages. First, a pilot study of 50--100 items was conducted with the first iteration of the annotator instructions. As part of the pilot, all ratings were required to have written justifications elaborating the reasoning for the provided rating. The results of the pilot were analyzed by the authors to identify common errors patterns; collected justifications were helpful in understanding the reasoning annotators used to arrive at their ratings. The results of the review were communicated back to the annotators, and the instructions were modified to emphasize areas leading to common ratings errors.

Next, a portion of the ratings was inspected by the authors for persistent error patterns and the feedback communicated to annotators. Additionally, the annotators collected edge cases where they found it difficult to make judgements. These edge cases were adjudicated by the authors; recurring complex patterns were used to expand the annotator guidelines (see the Appendix for full final instructions for \href{app:instructions_qa}{conversational QA}, \href{sec:instructions_summarization}{summarization}, and \href{sec:instructions_t2t}{table-to-text}).

Finally, the annotator team performed internal audits on a subset of completed tasks.

Annotators were initially trained on the conversational QA tasks; other tasks and training were introduced subsequently.

\section{Experiments}
\label{sec:task-results}
In the following section, we demonstrate the utility of the AIS templates by showing how it can be applied to three different tasks (conversational QA, summarization, and table-to-text generation) in which the model output is --- by design --- always meant to be attributable to some source document.  We  instantiated the AIS annotation template for four datasets in these domains (see Table~\ref{tab:tasks}) and performed human evaluation studies on generated outputs from multiple models. In order to show the applicability of AIS in detecting nuanced differences between different types of model outputs, we specifically chose models for each dataset that would represent a range of different types of outputs rather than just selecting a set of state-of-the-art models.  We also annotated a selection of gold references from each dataset to better understand the AIS quality of existing datasets in these areas.  We end with analysis of how effectively humans can annotate AIS as well as a discussion of various interpretability and AIS patterns that we found in the resulting annotations.
\begin{table*}
    \caption{Summary of tasks used in human annotation study.}
    \begin{tabularx}{\textwidth}{X >{\raggedright}X X >{\raggedright\arraybackslash}X X}
        \toprule
        Task & Dataset & $C$ & $P$ & $S$ \\
        \midrule
        Conversational QA & QReCC \cite{qrecc} & Conversational History & Retrieved Document & Response \\
        \midrule
        Conversational QA & Wizard of Wikipedia \cite{dinan2019wizard} & Conversational History & Retrieved Fact & Response \\
        \midrule
        Summarization & CNN/DM \cite{cnndm} & N/A & Source Article & Summary \\
        \midrule
        Table-To-Text & ToTTo \cite{parikh-etal-2020-totto} & N/A & Table, Table Description & Caption \\
        \bottomrule
    \end{tabularx}
    \label{tab:tasks}
\end{table*}

\subsection{QReCC Answer Generation}
\paragraph{Set-up} We use the QReCC dataset \cite{qrecc}, a collection of multi-turn conversational QA interactions that extends conversations coming from NaturalQuestions \cite{NQppr}, QUAC \cite{quacppr}, and CAST-19 \cite{castppr}.  In this task, a model is given a conversational history and generates a contextualized response.  We use a task set-up where the document passage containing the answer to the current query has already been retrieved (using the oracle retrieved document passage as the attributed source). We use different variations of T5 models including both base and small size variants.  First, we use the pre-trained T5 models (PT) by themselves by prompting the model (formatted as: ``Query:... Conversation History: ... Document: ... Answer:'').  We also use a version of T5 that has been fine-tuned on QReCC (FT) which uses special tokens to separate the query, context, and document instead of natural-language prompts.  Lastly, to sanity-check the AIS measures, we use a version of the model (no evidence) that only sees the query and conversation history but not the document at generation time.  We expect that the AIS subscores should be much lower in the model that does not use the evidence from document to generate the answer.
 
\begin{table*}
    \caption{Results of human study on 200 examples from QReCC test set (randomly sampled from set of examples where conversation length $\leq5$ turns). PT= pretrained model, FT=fine-tuned on QReCC training data.}
    \begin{tabular*}{\textwidth}{l l | l l}
        \toprule
        Model & Size& Int & AIS  \\
        \midrule
        T5-PT (with Evidence) & Small & 43.0$^*$ & 82.6\\
        & Base & 47.0$^*$ & 69.1$^*$\\
        T5-FT (no Evidence) & Small & 57.8$^*$ & 25.2$^*$\\
        & Base & 59.8$^*$ & 21.8$^*$\\
        T5-FT (with Evidence) & Small & 99.0 & 87.9\\
        & Base & 98.0 & 87.2 \\
        \midrule
        {\it Reference}& & 99.0 & 87.8 \\
        \bottomrule
    \end{tabular*} \\
    \label{tab:qrecc}
    \tabnote{\footnotesize{* Indicates that the result is significantly lower than the \textbf{highest score} in the column (with $p<0.01$).}}
\end{table*}
\paragraph{Results} We show results in Table~\ref{tab:qrecc}. The model outputs' interpretability increases substantially after fine-tuning (by about 50 points).  The AIS subscore is  highest in the fine-tuned model that uses evidence in its input.  As expected, the AIS is drastically lower in the model that does not use the document as input at generation time (the no evidence model) which is both interpretable and AIS only 15\% of the time.  Differences between model sizes (small vs. base) are generally not significant except for the pretrained-only model, though the AIS scores of the smaller versions are typically slightly higher.

\subsection{WoW Answer Generation}
\paragraph{Set-up} We used the seen portion of the test set from Wizard of Wikipedia \cite{dinan2019wizard}.  In this task, a model is given a conversational history and generates a contextualized response based on information from Wikipedia. As with QReCC, we again use a set-up where the Wikipedia sentence has already been retrieved (using the oracle retrieved sentence as the attributed source). To avoid chit--chat style utterances that may not be sharing new information, we sampled 200 examples per model where the previous utterance was a question (contains `?').  We used the models from \citet{rashkinetal}. That paper introduced a controlled T5 model trained on the Wizard of Wikipedia data which uses control tags and re-sampling to target generations that are more faithful to the document (by looking at heuristics such as entailment metrics, lexical precision, and first-person usage). Similar to that paper, we also compared with three models that are seq2seq-style conversation models: the original answer generation system from \citet{dinan2019wizard}, the Dodecadialogue multitask system from \citet{dodeca} and a T5-base model \citep{t5ppr} finetuned on Wizard of Wikipedia data. Because the model from \citet{rashkinetal} was specifically trained to be more faithful to evidence, we expect that it will score higher in the AIS category.

\begin{table*}
    \caption{Results of human study on 200 examples from Wizard of Wikipedia test set \citep{dinan2019wizard} (the seen topic split, using only conversation turns where the previous turn has a question mark).}
    \begin{tabular*}{\textwidth}{l | l l l}
        \toprule
        Model & Flag & Int & AIS   \\
        \midrule
        WoW Baseline \cite{dinan2019wizard} &4.0 & 84.4$^*$ & 19.8$^*$ \\
        Dodeca \cite{dodeca} &  8.5&100.0 & 60.1$^*$\\
        T5 \cite{t5ppr} & 5.5& 98.4 & 39.8$^*$ \\
        T5 (with Controls) \cite{rashkinetal} &7.5& 99.5 & 92.4 \\
        \midrule
        {\it Reference} & 4.0 &100.0 & 15.6$^*$ \\
        \bottomrule
    \end{tabular*}\\
    \tabnote{\footnotesize{* Indicates that the result is significantly lower than the \textbf{highest score} in the column (with $p<0.01$).}}
    \label{tab:wow}
\end{table*}
\paragraph{Results} We show results in Table~\ref{tab:wow}.  Compared to the QReCC data (in which only a few examples were flagged), more examples were flagged with the Wizard of Wikipedia data, which we included as an extra column.  The general trend of results is similar to what was found in the human evaluations of faithfulness and subjectivity in \citet{rashkinetal}.  As expected, the model that has specific controllable inputs for increasing the model's faithfulness to the input document achieves the highest the AIS scores overall.  We also note that the AIS scores of the gold references is lower than the model outputs.  We discuss this more in Section~\ref{sec:goldref}.

\subsection{CNN/DM Summarization}
\paragraph{Set-up} We extend our evaluation framework for a second task, summarization to confirm that AIS can be more broadly applicable.  AIS is crucial in summarization where a generated summary ($S$) must be well-supported by the source article ($P$). In contrast to some of the prior work in hallucination evaluation in summarization \citep{durmus-etal-2020-feqa,maynez2020faithfulness}, the annotators in our task evaluate the full summary for attribution (rather than at a sentence-level or a span-level), in order to account for cases where two individual text spans may be attributable to a source document but --- when composed together --- convey information that is different from the source document (e.g.,  misordered events, pronouns that no longer have the correct references when misordered, etc.). As a first step in applying AIS to summarization, we compare the performance of three different approaches (abstractive vs. extractive vs. hybrid) on 200 examples randomly sampled from the CNN/DM \citep{cnndm} test set.  The source articles in this dataset come from articles in CNN and DailyMail news and the summaries are extracted from bulleted highlights that were included with the article by the journalists.  We expect that high-quality AIS annotations will show a trend where extractive systems achieve higher AIS scores because they are copying directly from the source without adding anything. First, we used MatchSum \citep{matchsum}, a state-of-the-art extractive summarization model.  Because this model is extractive, it is expected that it will be the least prone to hallucinations.  We also used an abstractive summarization system, BigBird \citep{bigbird}.  Lastly, we used Pointer-generator Networks from \citet{pointernetwork} --- a hybrid approach that is uses an abstractive seq2seq model but with an explicit copy mechanism that can extract information from the source document.

\begin{table*}
    \caption{Results of human study on 200 examples from CNN/DM test set (randomly sampled).  Of the three models we tested with, unsurprisingly the more extractive models have higher AIS scores. }
    \begin{tabularx}{\textwidth}{l|l|ll}
    \toprule
        Model & Approach & Int & AIS  \\
        \midrule
        MatchSum \cite{matchsum} & Extractive &90.0 & 99.4\\
        Pointer-Gen \cite{pointernetwork} & Hybrid & 90.0 & 97.8\\
        BigBird \cite{bigbird} & Abstractive & 90.0 & 87.2$^*$ \\
        \midrule
        {\it Reference} &-& 86.0 & 54.1$^*$ \\
    \bottomrule
    \end{tabularx} \\
    \tabnote{\footnotesize{* Indicates that the result is significantly lower than the \textbf{highest score} in the column (with $p<0.01$).}}
    \label{tab:cnndm}
\end{table*}

\paragraph{Results} We show results in Table~\ref{tab:cnndm}. The more extractive approaches generally reach higher AIS subscores. This is a somewhat expected result --- extractive systems are less likely to output hallucinations as they are quoting information verbatim from the documents. As with Wizard of Wikipedia, the AIS scores of the gold reference summaries is surprisingly lower than the model output, which we will discuss more in Section~\ref{sec:goldref}.

\subsection{Table-to-Text ToTTo data}
\paragraph{Set-up} Lastly, we show the utility of extending AIS to a table-to-text task where $P$ is a table rather than a text document.  $S$ is a sentence generated by a model to describe some highlighted portion of the table.  We chose the ToTTo dataset \cite{parikh-etal-2020-totto}, testing with T5 and ByT5 models that were previously used with this data in the GEM benchmark \cite{gehrmann-etal-2021-gem}.  We experiment with two different sizes of ByT5 and three different sizes of the T5 architecture. As before, we sampled the output of 200 examples from the test set.  We also annotated 200 ground-truth references from examples in the dev. set (as the test set does not have gold-truth references publicly available).

\begin{table*}
    \caption{Results of human study on 200 examples from ToTTo test set (model output) and development set (ground-truth references).}
    \begin{tabular*}{\textwidth}{l|lll}
        \toprule
        Model & Flag & Int & AIS  \\
        \midrule
        ByT5-Base & 0.0 & 78.9* & 88.5\\
        ByT5-XL & 0.0 & 79.5* & 86.2\\
        T5-Small & 3.0 & 86.5 &88.6\\
        T5-Base & 5.0 & 91.1 & 86.6 \\
        T5-XL & 6.0 & 89.4 & 85.1*\\
        \midrule
        {\it Reference} &0.0& 83.9 & 91.0 \\
        \bottomrule
    \end{tabular*}\\
    \tabnote{\footnotesize{* Indicates that the result is significantly lower than the \textbf{highest score} in the column (with $p<0.01$)}}
    \label{tab:totto}
\end{table*}
\paragraph{Results} We show results in Table~\ref{tab:totto}. The model with the most ``interpretable'' responses was T5-base, with the ByT5 architectures being significantly less interpretable.  On the other hand, the T5 architecture responses were more likely to be flagged (according to the annotators this was because the flagged responses contained artefacts like unintelligible character encoding errors). Generally, we don't observe statistically significant differences in the AIS subscores though the larger architectures tended to have slightly lower AIS scores (similar to our observations of Table~\ref{tab:qrecc}).

\subsection{Annotation Quality}
In this section we discuss the further implications of the human annotation results.  We focus on two primary questions: (1) can humans reliably annotate AIS? and (2) what do our measured AIS ratings indicate about NLP data and models?

\subsubsection{Interannotator Agreement}
\begin{table*}
    \caption{Annotator agreement measured as interannotator agreement (left half of the table) or as agreement with expert consensus (right half of the table, only measured on QRECC and CNN/DM tasks).  Metrics include --- F1: a F1 measure comparing individual ratings to the consensus rating; PA: pairwise agreement as percentage of individual pairs that agree; $\alpha$: Krippendorff's alpha measure comparing pairs of individual ratings.}
    \begin{tabular*}{\textwidth}{l|lll|lll|lll|lll}
    \toprule
       & \multicolumn{6}{c}{IAA} & \multicolumn{6}{c}{vs. Expert} \\
       & \multicolumn{3}{c}{Int} & \multicolumn{3}{c}{AIS} & \multicolumn{3}{c}{Int} & \multicolumn{3}{c}{AIS} \\
        Task &  F1 & PA & $\alpha$&  F1 & PA & $\alpha$&  F1 & PA & $\alpha$&  F1 & PA & $\alpha$  \\
        \midrule
        CNN/DM & .83&.80&.46&.92&.89&.69&.48&.60&-.04&.81&.86&.61 \\
        QReCC & .97&.96&.91&.93&.89&.76&.77&.81&.54&.77&.78&.54\\
        WoW & .88&.93&.60&.95&.88&.79 &-&-&-&-&-&-\\
        ToTTo & .95 &.95&.84&.92&.92&.74 &-&-&-&-&-&-\\
    \bottomrule
    \end{tabular*}
    \label{tab:iaa}
\end{table*}
We show the interannotator agreement (IAA) for crowd annotators in the left half of Table~\ref{tab:iaa}.  The metrics we used include Krippendorff's alpha comparing individual ratings, pairwise agreement (PA) comparing individual ratings and an F1 score comparing individual ratings to the consensus (majority vote). Agreement results are generally moderate to high, displaying that --- while this is a challenging task --- the annotators are able to be fairly consistent with one another.  The alpha scores are generally lowest on the summarization CNN/DM task, perhaps because the output text is much longer in summarization, increasing the complexity of the rating task.  The F1 scores are similarly high, particularly on the AIS ratings.

\subsubsection{Audits}
\begin{table*}
    \caption{Quality measure on samples of annotations for conversational QA, summarization, and table-to-text tasks. Snapshots represent consecutive annotation sprints with individual annotator judgements (Ann) replicated at 5 per task. A sample (\textit{Smpl}) of each snapshot was evaluated by a project lead on the annotator team. The quality of annotations (\textit{Qual}) was assessed over a varying number of snapshots for each task. The evaluated annotations exclude flagged tasks.}
    \begin{tabular*}{\textwidth}{l|lll|lll|lll}
    \toprule
    &\multicolumn{3}{c|}{Conversational QA}
    &\multicolumn{3}{c|}{Summarization}
    &\multicolumn{3}{c}{Table-to-Text}\\
        Snapshot & Ann & Smpl & Qual & Ann & Smpl & Qual & Ann & Smpl & Qual\\
        \midrule
        1 & 642 & .04 & 1.00 & 88 & .06 & .67 & 261 & .08 & 1.00 \\
        2 & 726 & .05 & 1.00 & 339 & .06 & .87 & 2,518 & .19 & .96 \\
        3 & 1,895 & .03 & 1.00 & 469 & .10 & .94 & 2,463 & .30 & .97 \\
        4 & 2,520 & .04 & .97 & 682 & .08 & 1.00 & 1,151 & .34 & .96 \\
        5 &-&-&-& 608 & .08 & 1.00 & 849 & .30 & .94 \\
        6 &-&-&-& 652 & .08 & 1.00 &-&-&- \\
        7 &-&-&-& 928 & .03 & 1.00 &-&-&- \\
        \midrule
        \emph{Total} & 5,783 & .04 & .99 & 3,766 & .07 & .98 & 7,242 & .26 & .96 \\
        \bottomrule
    \end{tabular*}
    \label{tab:qc_stats}
\end{table*}
Separately, the annotator team also performed  internal audits on the annotation quality where a project lead from the annotator team examined a sample of individual annotator judgements at different points (snapshots) of the annotation process (Table~\ref{tab:qc_stats}). QRECC and WoW annotations were evaluated together as the broader conversational QA annotation task. The overall reported quality is in the high nineties for all three tasks with slight variations. The annotation quality for the conversational QA tasks remains high across all snapshots; we attribute this to the annotators extended experience with the task prior to the annotation of this dataset\footnote{The annotator pool was involved in annotating a series of related tasks for Conversational QA beyond the reported results in this paper.}. The quality of the summarization annotations shows an increase over snapshots, as annotators internalize the guidelines and gain expertise in the task. The quality of the table-to-text annotations fluctuates and is generally the lowest of the three tasks; we attribute this to a much larger sample for which quality was measured. Overall, across the three tasks, the larger the quality evaluated sample, the lower the overall reported quality. Barring genuine task differences that would lead to variations in annotation quality, this suggests that the reported table-to-text quality of annotations is the most representative of all three tasks. 

\subsubsection{Crowd Annotator Performance}
\begin{table*}[th]
    \caption{Average completion times (\textit{ACT}) for ratings tasks in seconds. Conversational QA tasks include evaluation of generated text for QReCC and WoW. Summraziation tasks include evaluation of generated text for CNN/DM. Table-to-text tasks include evaluation of generated text for ToTTo. Justifications were required for all question tasks at the pilot stage, but not at the production stages. Average completion times decrease for all three task types as annotators gain more experiences over the amount of observed tasks (\textit{Tasks}), but are always relatively longer for summarization. Note that the average completion times may be reduced further for summarization with more tasks observed by annotators, a pattern we see in the conversational QA and table-to-text task types.}
    \begin{tabularx}{\textwidth}{l|l|lX|lX|lX}
    \toprule
    & &\multicolumn{2}{c|}{Conversational QA}
    &\multicolumn{2}{c|}{Summarization}
    &\multicolumn{2}{c}{Table-to-Text}\\
        Stage & Justification & Tasks & ACT, secs. & Tasks & ACT, secs. & Tasks & ACT, secs. \\
        \midrule
        Pilot & $+$ & 75 & 375.30 & 50 & 687.02 & 100 & 324.72 \\
        Start & $-$ & 762 & 136.76 & 187 & 308.14 & 136 & 238.40 \\
        Finish & $-$ & 4,022 & 73.19 & 900 & 263.55 & 1,496 & 193.95 \\
    \bottomrule
    \end{tabularx}
    \label{tab:completion_time}
\end{table*}
We also observe average task completion times decrease across all types of tasks as the annotators are exposed to more tasks and internalize the instructions (Table~\ref{tab:completion_time}). Initial pilots for the conversational QA, summarization, and table-to-text tasks included required rating justifications for interpretability and AIS questions as part of annotator training. Once the production annotation started and the justifications were no longer required, completion times decreased significantly for all types of tasks, which we attribute primarily to the annotators no longer typing out detailed justifications of their ratings, but also overall internalization of the guidelines. The effects of annotators internalizing the guidelines are also evident when comparing completion times at production annotation start and finish: average task completion times decrease for all three types of tasks as annotators are exposed to more tasks and gain more experience.

At the same time, the absolute task completion times are consistently and substantially different across the three tasks suggesting their uneven complexity, with conversational QA taking the shortest amount of time to complete, summarization requiring the longest, and table-to-text falling in-between. This pattern follows the trend in the distribution of inter-annotator agreement across the three tasks: tasks with shorter completion times generally have higher interannotator agreement. We postulate that this is primarily due to the difference in the amount of context that is necessary to perform ratings. Although conversational QA tasks may contain several turns of preceding interactions between the system and the user as well as the source document, the amount of information in the source articles in the summarization task is substantially larger. Likewise, source tables in the table-to-text task can be extensive and have the added information complexity of cell highlighting and table metadata. Finally, register and discourse structure effects may be at play here as well. Conversational QA tasks build upon colloquial interactions between the user and the system, setting up the context of the interaction in shorter utterances and helping annotators anticipate the contents of the source document. Likewise, Wikipedia, news articles, and tables package information differently as they serve somewhat different communicative goals, and it is possible that one of these source types is more amenable to inspection required for performing AIS ratings. 

\subsubsection{Expert Ratings}
\begin{table*}
    \small
    \centering
    \caption{Selected conversational QA examples marked as \textbf{AIS} by annotators but \textbf{not AIS} by experts where differences in how entities are named (first name vs. last name) in the response vs. document may cause some ambiguity as to whether the document is accurately describing the same person.  Some people may also have background knowledge about the entities involved that helps them disambiguate whether these documents are referring to the same ``mayer'' or ``adams'' as the question.}
    \begin{tabularx}{\textwidth}{X}
    \toprule
    \textbf{Example 1}\\
    \textbf{Query}\\ 
    what is a charity john mayer works for? \\
    \textbf{Response}\\
in 2002, john mayer created the back to you fund, a nonprofit organization that focuses on fundraising in the areas of health care, education, the arts and talent development. \\
    \textbf{Attributed Document}\\ 
    personal life on his third episode of current mood , mayer revealed he had been sober for two years. he stopped drinking after having what he says was a six-day hangover after drake 's 30th birthday party. relationships mayer has never married, but has been romantically involved with jennifer love hewitt , jessica simpson , minka kelly , jennifer aniston , and katy perry . philanthropy in 2002, mayer created the "back to you" fund, a nonprofit organization that focuses on fundraising in the areas of health care, education, the arts and talent development. the foundation auctions exclusive john mayer items, such as guitar picks, t-shirts and signed cds. \dots \\ 
    \midrule
    \textbf{Example 2}\\
    \textbf{Conversation History}\\
    \textsc{User:} where did john adams practice law? \\
    \textsc{System:} harvard
\\
    \textbf{Query}\\
    what happened after john adams practiced law at harvard? \\
    \textbf{Response}\\
    in 1756, john adams began reading law under james putnam, a leading lawyer in worcester. \\
    \textbf{Attributed Document}\\
    law practice and marriage abigail smith adams – 1766 portrait by benjamin blyth john adams – 1766 portrait also by blyth in 1756, adams began reading law under james putnam, a leading lawyer in worcester. in 1758, he earned an a.m. from harvard, and was soon admitted to the bar, having completed his studies under putnam. he developed an early habit of writing about events and impressions of men in his diary; this included james otis jr. 's 1761 legal argument challenging the legality of british writs of assistance , allowing the british to search a home without notice or reason. \dots \\
    \bottomrule
    \end{tabularx}
    \label{tab:examples:qa:exp}
\end{table*}

\begin{table*}
    \small
    \centering
    \caption{Selected summary examples marked as interpretable by annotators but non-interpretable by experts.  We note that the style of language in these summaries can be vague which may increase the difficulty in leaving a binary interpretability judgement.}
    \begin{tabularx}{\textwidth}{X}
    \toprule
    \textbf{Example 1}\\
    \textbf{Summary} \\
deciding who you will vote for may have more to do with your family than who won the leaders debate ( above ) finds study which looked at the voting habits of twins born in the uk . \newline the aim was to explore how much nature and nurture influence our party political allegiances and potential voting preferences \\
    \midrule
    \textbf{Example 2}\\
    \textbf{Summary} \\
Charlie Stayt was broadcasting live from a primary school in Southampton . \newline He missed out the letter 'c' when he scrawled the word on a whiteboard . \newline Outraged viewers took to Twitter to complain about the spelling error . \newline Stayt later described the gaffe as 'one of those things'\\
    \midrule
    \textbf{Example 3}\\
    \textbf{Summary} \\
university lecturer dr alex russell shares his expert advice . \newline dr russell says that anyone can improve their tasting skills in four hours .\\
    \midrule
    \textbf{Example 4}\\
    \textbf{Summary} \\
in fact , it 's an advert from cosmetics giant revlon for their latest lipstick . \newline the stylish ad is filmed entirely in black and white , with just a slick of pink visible on the woman's lips. \newline revlon uk 's new global tag line , love is on , is the label 's first major relaunch in more than a decade . \\
    \bottomrule
    \end{tabularx}
    \label{tab:examples:summ:int}
\end{table*}
Where AIS is used as a metric for ranking generative models, the internal consistency of crowd annotations is paramount. But, to help illuminate the inherent challenges in calibrating this annotation task, we also compare the crowd ratings with those of expert on a small set of examples.
Due to the challenges of scaling expert evaluations, we limited expert ratings to two tasks (CNN/DM and QReCC) with 50 examples each.  The experts (two co-authors) first annotated the examples separately from each other using the same interface as the crowd annotators and then discussed their answers to reach a consensus. \emph{Expertise} here might be derived from general educational background (a different approach to close reading), the ability to discuss annotations (and to do so carefully at self-guided pace), specialized knowledge, and first-hand familiarity with the evaluation framework. Expertise does not imply that the experts have more experience performing the task than the crowd annotators.

In order to account for natural ambiguity in assigning a rating category, experts marked some cases as ``either option acceptable''. We compare the individual crowd annotator ratings to the expert consensus in the right half of Table~\ref{tab:iaa}. Crowd annotators tend to agree with each other more than they agree with experts, which is expected due to differences in background, incentives, and procedure, although there is still reasonably consistent agreement in most cases.  On closer inspection, we find that most disagreements are cases where there is underlying ambiguity caused by vagueness in the evidence or model output. In these cases experts erred more on the side of being critical of the model and crowd annotators erred more towards being lenient.  In the case of conversational QA, most of the AIS disagreements involved cases where the document and the response do not refer to an entity using the same naming conventions (e.g., using both first and last name; see Table~\ref{tab:examples:qa:exp}) leaving some ambiguity that the document is referring to the same entity as the response. The greatest source of disagreements overall is the interpretability question in the summarization task (see examples in Table~\ref{tab:examples:summ:int}).  The summaries in the CNN/DM dataset were originally crawled from high-level article highlights, and experts observed that --- due to the linguistic style of these highlights --- there were many cases where the language may be vague or ambiguous, making this dimension more challenging. 
Because we use interpretability as a pre-filtering stage for the AIS question, we make allowances for the annotators being more inclusive.  Despite the differences on the interpretability dimension, they generally agreed with experts on most AIS questions, our primary evaluation dimension.

\begin{table}
    \small
    \centering
    \caption{Examples of \textbf{reference} (i.e. gold) text that was marked by annotators as \textbf{interpretable but not AIS}.}
    \begin{tabular}{lp{13cm}}
    \toprule

\textbf{Example 1. (CNN/DM)}\\
\textbf{Summary} \textit{(Reference)}\\ \multicolumn{2}{p{13cm}}{thomas piermayr has been training with blackpool this week . \newline austrian defender is a free agent after leaving mls side colorado rapids . \newline blackpool are bottom of the championship and look set to be relegated . 
.}\\
\textbf{Attributed Document}\\ \multicolumn{2}{p{13cm}}{Blackpool are in talks to sign Austria defender Thomas Piermayr. \newline The 25-year-old has been training with the Championship club this week and they are keen to get him on board for what is expected to be confirmed as a campaign in League One next season. \newline Piermayr is a free agent and had been playing for Colorado Rapids. \newline The former Austria U21 international had a spell with Inverness Caledonian Thistle in 2011. \newline Thomas Piermayr (left, in action for the Colorado Rapids) tries to tackle Obafemi Martins last year .
.}\\
\multicolumn{2}{p{13cm}}{{\bf Explanation:} The article doesn't mention Blackpool being the bottom of the championship but it is included in the gold summary.  Possibly the journalist who wrote the article highlights (that the summary data was extracted from) included that sentence because they were assuming that the sports section audience might already be aware of the standings.}
\\\midrule
       \textbf{Example 2. (Wizard of Wikipedia)}\\ \textbf{Query}\\ \multicolumn{2}{p{13cm}}{
i really love kentucky, i was born and raised here, have you ever been there? }
\\
        \textbf{Response} \textit{(Reference)}\\ \multicolumn{2}{p{13cm}}{
my parents are from kentucky! i have only been to east, south-central region state a few times.
        } \\
\textbf{Attributed Document}\\ \multicolumn{2}{p{13cm}}{kentucky (, ), officially the commonwealth of kentucky, is a state located in the east south-central region of the united states.}\\
\multicolumn{2}{p{13cm}}{{\bf Explanation:} the original interlocutor relayed some information from the document and also embellished with their own personal experiences (that are not verifiable) to make a more engaging conversation.}\\\midrule
        \textbf{Example 3. (QReCC)}\\ \textbf{Query}\\ \multicolumn{2}{p{13cm}}{
what musical has the song you'll never walk alone }
\\
        \textbf{Response} \textit{(Reference)}\\ \multicolumn{2}{p{13cm}}{
 you'll never walk alone is a show tune from the 1945 rodgers and hammerstein musical carousel.
        } \\
\textbf{Attributed Document}\\ \multicolumn{2}{p{13cm}}{liverpool is one of the best supported clubs in the world. \newline
the club states that its worldwide fan base includes more than 200 officially recognised club of the lfc official supporters clubs in at least 50 countries. notable groups include spirit of shankly. \newline
the club takes advantage of this support through its worldwide summer tours, which has included playing in front of 101,000 in michigan, u.s., and 95,000 in melbourne, australia. \newline 
liverpool fans often refer to themselves as kopites , a reference to the fans who once stood, and now sit, on the kop at anfield.
in 2008 a group of fans decided to form a splinter club, a.f.c. liverpool , to play matches for fans who had been priced out of watching premier league football. \newline
the song " you'll never walk alone ", originally from the rodgers and hammerstein musical carousel and later recorded by liverpool musicians gerry and the pacemakers , is the club's anthem and has been sung by the anfield crowd since the early 1960s.\newline
it has since gained popularity among fans of other clubs around the world.\newline
the song's title adorns the top of the shankly gates, which were unveiled on 2 august 1982 in memory of former manager bill shankly.\newline
the "you'll never walk alone" portion of the shankly gates is also reproduced on the club's crest.}\\
\multicolumn{2}{p{13cm}}{{\bf Explanation:} The year "Carousel" was made (1945) cannot be attributed to the selected passage.  The original interlocutor may have seen that detail elsewhere.}\\
    \bottomrule
    \end{tabular}
    \label{tab:examples:notais:ref}
\end{table}

\subsubsection{Limitations of Gold References}
\label{sec:goldref}
The last rows in Tables \ref{tab:qrecc}, \ref{tab:wow}, \ref{tab:cnndm} and  \ref{tab:totto} show annotation results on reference answers sampled from these datasets.  The results demonstrate that there is actually a limit on the AIS quality of the data itself in multiple tasks.  We include examples of non-AIS references in Table~\ref{tab:examples:notais:ref} to illustrate what some of these examples look like.  We hypothesize that this is because the originators of the data were not specifically instructed to be as faithful to the underlying documents as possible.  In the case of Wizard of Wikipedia \citep{dinan2019wizard}, the gold response is only AIS 16\% of the time.  But, this dataset was constructed for a different objective --- to contain \emph{both} informative and engaging responses. The MTurk workers who created the data were provided documents to enhance their conversations but could do so at their own discretion, often including their own thoughts and opinions in the conversation as well. This is also reflected in the CNN/DM AIS scores --- summaries in CNN/DM are only attributable to the documents in 54\% of the interpretable examples. 
Looking more closely, we speculate that this may be due to the post-hoc data creation process used to extract summaries from article highlights written by journalists.  We observed that the reference summaries in CNN/DM may sometimes refer to external pieces of information that may have accompanied the article (a picture, a headline, etc.) or sometimes make assumptions about what the intended audience of the article might already know that can affect either the interpretability or AIS scores (see Example 1 in Table~\ref{tab:examples:notais:ref} and Example 3 in Table~\ref{tab:examples:notint}). These results indicate that there is still a need for high-quality AIS data for training new NLG models.

\subsubsection{Examples}
In the Appendix, we separately list textual examples rated as uninterpretable (Table~\ref{tab:examples:notint}), interpretable but not AIS  (Table~\ref{tab:examples:notais}), or both interpretable and AIS (Table~\ref{tab:examples:ais}).  For the table-to-text task, we present examples in a more visual figure,  Figure~\ref{fig:ttt:ex}, for better legibility.  Common factors in marking text as ``uninterpretable'' include repetitive, degenerate language and ambiguous pronouns and ellipses.  Additionally, some outputs are marked as uninterpretable because they are hard to understand ``on their own''.  Whether or not a piece of text can be understood may also rely on things like commonsense and background knowledge that could vary depending on annotators' backgrounds (see Example 3 from Table~\ref{tab:examples:notint}). Ambiguous references can also affect both interpretability and the AIS scores.  In Example 2 of Table~\ref{tab:examples:notais}, the retrieved document did not provide enough information to completely verify the response since it never refers to Ann Veneman by her full name.  This is a seemingly minor detail, but annotators were often sensitive to this type of example since they could not verify whether the document was actually referring to the same entity as the model output.  Another type of non-AIS output that frequently appeared in the QReCC data were cases where a model outputted a seemingly informative statement that --- instead of being grounded to the document --- was actually grounded to a previous conversation turn, sometimes repeating itself verbatim.
Lastly, examples verify that AIS evaluations can be disentangled from other quality aspects, such as conversational relevance. This was challenging to instruct to annotators as it is instinctual to judge quality more holistically, and they were explicitly given instructions with multiple examples illustrating what types of quality aspects to ignore. In the resulting annotations, they would mark incoherent summaries or irrelevant conversational replies as AIS if they conveyed well supported information, appropriately disregarding other aspects of quality.

\section{Discussion}

Generative models have been advancing toward human-like competence in some aspects. Their real-world application in consumer-focused information products are becoming more attractive, for example, for summarizing original descriptions of events, or for deriving answers to pertinent questions about the world. Traditionally, this type of information transformation has been performed by specialized human experts (e.g., journalists, researchers), who are required to meet a variety of standards of accuracy and accountability, maintaining one or more sources for a proposition and performing fact-checking. The task could also be likened to the practice of law, where norms are examined for their subsumptive relationship to a set of circumstances, and where both close reading and a set of conventionalized tests aid this determination.

We formalize a specific sub-task of fact-checking, namely, verification against a known source, as a necessary but not sufficient step in ensuring the quality of generated text. We show that with the right training, careful instructions, and optimized user interfaces, we can delegate the judgment of attribution to underlying source(s) to crowd workers, but we also find limitations. Following the data collection we described, we found it necessary to set some standards in our instructions to raters. This includes setting expectations for named entities, for example, whether first and last names are needed to identify an individual and to link them between evidence and statement, or if a place name without qualification may be acceptable as long as there are no other well-known places of the same name. Similarly, as statements and evidence become more complex, raters inevitably draw inferences using individual world knowledge. This is unavoidable and is inherently noisy \cite{pavlick2019inference}; ground truth is ambiguous, just like journalists, researchers, or judges often legitimately disagree. Possible model outputs fall on a spectrum ranging from synthesized information to the mostly unassailable extractive generations \cite{ladhak2022faithful}. AIS does not set policy about where model output should fall: its users still need to decide where to draw the line.

AIS is limited to propositions that can be judged with the "according to" framework. AIS is not applicable to questions (without presuppositions) or imperatives (commands and requests). There are also scenarios where strict attribution contradicts other desirable output characteristics (e.g., chit--chat systems). We did not examine AIS on such data. How to evaluate hybrid systems that mix entertaining and informative communicative goals --- capturing the attribution of the informative portion but ignoring the rest --- is unclear, as is the question of whether systems with blurry boundaries between what is and is not subject to attribution should exist at all. 

We have purposefully limited the availability of context in our definition. Practical human--computer interactions may actually take place in context beyond the shared time $t$ that is used in the definition (Section~\ref{sec:definitions}), perhaps because the communication channel is richer than a text-based line of transmission, and because it may be further extended by multi-session interaction history. It is important that annotators remain aware of the notion of explicature, resolving explicit references and implicit topics available to the communicators. It is possible that the use of models that perform this task \cite{choi2021decontext} can improve the performance of raters. We are also aware that this task requires close reading, which is challenging to implement on crowdsourcing platforms where speed, efficiency, and cost are incentivized instead. Again, models may be useful in extracting explicit, elementary propositions from complex statements, making this task easier for raters. We will examine such approaches in future work.

\section{Conclusion}
\label{sec:conclusion}
In this paper, we define a new evaluation framework called \emph{Attributable to Identified Sources} which allows us to inspect whether information in generated text can be supported by source documents.  We provide formal definitions of AIS and descriptions of how it can be applied to three different NLG tasks (conversational QA, summarization, and table-to-text generation). We validate this evaluation framework quantitatively on human evaluation studies, in which annotators rated the AIS of model output as part of a two-stage annotation pipeline. The results of the human evaluation studies demonstrate that high-quality AIS ratings can be obtained empirically. The results shed light on some of the ongoing challenges in training NLG models; having solid AIS is the basis for addressing them.

\appendix
\appendixsection{Examples of Annotations}
\begin{table}[H]
    \centering
    \small
    \caption{Examples marked by annotators as \textbf{uninterpretable}.}
    \begin{tabularx}{\textwidth}{X}
    \toprule
    \textbf{Example 1. (Wizard of Wikipedia)} \\
    \textbf{Conversation History}\\
    \textsc{Apprentice}: hi, can you tell me about parenting? \\
    \textsc{Wizard}: its the process of promoting and supporting the physical, emotional, social, and intellectual development of your kid, its really important for kids \\
    \textsc{Apprentice}: thats fair enough, do you have kids? \\
    \textsc{Wizard}: no, but i am planing to have a big family one day, i really like the thought of raising a child \\
    \textbf{Query}\\ 
    yeah i dont have kid's but i'm really looking forward to it soon. when do you think is a good age to have kids? \\
    \textbf{Response} \textit{(T5 with Controls)}\\
    a good parent may be referred to as a "good parent" \\
    \textbf{Rating:} {Not Interpretable }\\
    \midrule
    \textbf{Example 2. (QReCC)} \\
    \textbf{Query}\\
    how did the wildhearts band members meet each other? \\
    \textbf{Response} \textit{(T5-Small Pretrained)}\\
    ginger \\
    \textbf{Rating:} {Not Interpretable} \\
    \midrule
    \textbf{Example 3. (CNN/DM)} \\
    \textbf{Summary} \textit{(Reference)}\\
    daniel andersson , helsinborg 's 42-year-old kit man , kept a clean sheet .\newline the emergency stopper played in season opener against kalmar . \newline henrik larsson 's first-choice goalkeepers were both out injured . \newline the former goalkeeper earned one cap for sweden back in 2001 .\\
    \textbf{Rating:} {Not Interpretable {\it (Note: This is actually a reference summary. Annotators selected this as unclear as it is difficult to understand on its own.  The original journalist writing this may have {\color{blue} assumed some amount of commonsense/background knowledge} about football/soccer for the intended audience as well as other information such as what games were being played that day.  The interpretability judgement could vary {\color{blue} depending on annotators' familiarity with the topic}.)}}\\
    \midrule
    \textbf{Example 4. (CNN/DM)} \\
    \textbf{Summary} \textit{(Pointer-Generator Network)}  \\
    ivan carlos , 22 , and eighteen-year-old brenda avilez , 18 , were sleeping in a trailer at the lone oak mobile home in palmetto at 2.30am wednesday . \newline the driver of the white chevrolet suv lost control of the vehicle and crashed through the fence at the lone oak mobile home in palmetto . \newline avilez was nine months pregnant at the time , the herald reports , the fetus did not survive . \newline crawford , 21 , was taken to a hospital with non-life-threatening injuries .\\
    \textbf{Rating:} {Not Interpretable {\it (Note: There's some {\color{blue} degenerate/repetitive} language (18, eighteen) as well as {\color{blue} unclear references that lack context} (``the white chevrolet suv''). )}}\\
    \bottomrule
    \end{tabularx}
    \label{tab:examples:notint}
\end{table}

\begin{table}[H]
    \centering
    \small
    \caption{Examples marked by annotators as \textbf{interpretable but not AIS}.}
    \begin{tabularx}{\textwidth}{X}
    \toprule
    \textbf{Example 1. (Wizard of Wikipedia)} \\
    \textbf{Conversation History}\\
    \textsc{Wizard:}  my pet horse is named beauty. she's domesticated :) \\ \textsc{Apprentice:} that must be awesome to have a pet horse! do you ride it? \\
    \textsc{Wizard:} yeah! we're not sure what breed she is, because we rescued her, but there are all sorts of different types of breeds and sizes of horses. \\
    \textbf{Query}\\
    how many different breeds are there? what color is your horse? \\
    \textbf{Response} \textit{(T5-Base)}\\
    i'm not sure, but there are a lot of domesticated horses, but some feral ones are wild.\\
    \textbf{Attributed Document}\\
    horses in the subspecies "caballus" are domesticated, although some domesticated populations live in the wild as feral horses. \\
    \textbf{Rating:} {Interpretable, not AIS}\\	
    \midrule
    \textbf{Example 2. (QReCC)} \\
    \textbf{Conversation History}\\
    \textsc{User:}are there any other interesting aspects about ann veneman?	\newline
    \textsc{System:} ann veneman has received several awards and distinctions throughout her career. \\
    \textbf{Query}\\
    what kind of awards has ann veneman won? \\
    \textbf{Response} \textit{(T5-Small)}\\ in 2009, ann veneman received the award of distinction from the university of california davis college of agricultural and environmental sciences.	\\
    \textbf{Attributed Document}\\
    In 2009, she received the award of distinction from the university of california davis college of agricultural and environmental sciences. veneman is an honorary member of rotary international (2008), received sesame workshop's leadership award for children (2006), and a humanitarian award from the united nations association of new york (2006). ... \\
    \textbf{Rating:} {Interpretable, Not AIS {\it (Note: The information is partially represented in the document but there's some {\color{blue} referential ambiguity} as the document only refers to ``she'' and ``veneman'' but it is unclear whether it's referring to ``Ann Veneman'' specifically.  3 of 5 annotators marked this as not AIS.)}}\\
    \midrule
    \textbf{Example 3. (QReCC)} \\
    \textbf{Conversation History} \\
    \textsc{User:} when was the verve's fiirst break up? \\
    \textsc{System:} the verve's turmoil continued into 1995's a northern soul. the album reached the uk top 20 in july, but ashcroft broke up the band three months later, just before history. \\
    \textsc{User:} was history by the verve a success? \\
    \textsc{System:} history was released 18 september 1995 as the third and final single from a northern soul, charting at \#24 in the uk singles chart. \\
    \textbf{Query} \\
    what other album was released by the verve besides a northen soul? \\
    \textbf{Response} \textit{(T5-Small without Evidence) }\\ 
    the verve's turmoil continued into 1995's a northern soul. \\
    \textbf{Attributed Document}\\ 
    commercial success and second break-up (1997–1999) in early 1997, ashcroft asked mccabe to return, stating: "i got to the point where nothing other than the verve would do for me". mccabe obliged and with the new line-up in place (tong remained on guitar alongside mccabe), the group went through a "spiritual" recording process to finish their third album urban hymns which was completed by early summer. "bitter sweet symphony" the first track on their album urban hymns , " bitter sweet symphony " is renowned for its signature swirling orchestral melody. problems playing this file? see media help ... \\
    \textbf{Rating:} {Interpretable, Not AIS {\it (Note: On a surface-level, the information seems plausibly informative but is {\color{blue} not supported in the provided document}.  This is actually only superficially {\color{blue} repeating a previous conversation turn}.  5 of 5 annotators marked this as not AIS.)}}\\
    \bottomrule
    \end{tabularx}
    \label{tab:examples:notais}
\end{table}

\begin{table}[H]
    \centering
    \small
    \caption{Examples marked by annotators as \textbf{interpretable and AIS}. }
    \begin{tabularx}{\textwidth}{X}
    \toprule
    \textbf{Example 1. (Wizard of Wikipedia)} \\
    \textbf{Conversation History}\\
    \textsc{Wizard:} what do you know about the chevrolet corvette? \\
    \textbf{Response} \textit{(T5-Base)} \\
    well, it's a sports car manufactured by chevrolet. \\
    \textbf{Attributed Document}\\
    the chevrolet corvette, known colloquially as the vette or chevy corvette, is a sports car manufactured by chevrolet. \\ \textbf{Rating:} {Interpretable, AIS}\\
    \midrule
    \textbf{Example 2. (QReCC)}\\
    \textbf{Conversation History}\\
    \textsc{User:} what is the british invasion in music? \\
    \textsc{System:} the british invasion was a cultural phenomenon of the mid-1960s, when rock and pop music acts from the united kingdom and other aspects of british culture, became popular in the united states and significant to the rising counterculture on both sides of the atlantic. \\
    \textbf{Query}\\
    what era did the british invasion in music happen? \\
    \textbf{Response} \textit{(T5-Base)} \\ 
    pop and rock groups such as the beatles, the rolling stones, the kinks, the dave clark five, herman's hermits, the zombies, and the animals were at the forefront of the british invasion.	 \\
    \textbf{Attributed Document}\\
    \dots the british invasion was a cultural phenomenon of the mid-1960s, when rock and pop music acts from the united kingdom and other aspects of british culture became popular in the united states and significant to the rising " counterculture " on both sides of the atlantic. pop and rock groups such as the beatles , the rolling stones , the kinks , the dave clark five , herman's hermits , the zombies , and the animals were at the forefront of the "invasion". \dots \\
    \textbf{Rating:} {Interpretable, AIS {\it (Note:  AIS is a focused metric on one dimension of quality. This response is AIS even though it fails in {\color{blue} other dialogue qualities}, i.e. relevance to previous query. )}}\\
    \midrule
    \textbf{Example 3. (CNN/DM)} \\
    \textbf{Summary}\\ 
    Gary Johnson took over as Cheltenham Town manager on March 30 . \\
    He asked his players to write 'I promise to do all I can to keep Cheltenham Town in the league' on an A3 sheet of paper .  \\
    'Some signed it and meant it, and some signed it and didn't mean it,' Johnson said . \\
    \textbf{Attributed Document}\\
    Cheltenham Town have two games to preserve their Football League status - and manager Gary Johnson has revealed one of the techniques he is using to try and bring the best out of his players. \\
    When Johnson took over as manager of the League Two club on March 30, he wrote 'I promise to do all I can to keep Cheltenham Town in the league' on an A3 sheet of paper and asked his players to put their signature on it. \\
    'They all signed it,' Johnson said to the BBC. 'Some signed it and meant it, and some signed it and didn't mean it. \\
    Cheltenham Town manager Gary Johnson got every payer to pledge to give his all when he took over . \\
    Cheltenham were beaten by Northampton in their last game and have two games left to try and stay up . \\
    'When you come to this stage of the season you need everyone to give everything for the cause,' Johnson added. \\
    'You also need team-mates you can rely on. The lads that are here need to know they can rely on the others - and if they can't rely on some then you have to move them on.' \\
    Cheltenham occupy 23rd in League Two and trail 22nd placed Hartlepool United and the safety places by a point. \\
    Their final two games are against second placed Shrewsbury and 13th placed Wimbledon.\\
    \textbf{Rating:} {Interpretable, AIS}\\
    \bottomrule
    \end{tabularx}
    \label{tab:examples:ais}
\end{table}

\begin{figure}[H]
    \centering
    \includegraphics[width=0.99\linewidth]{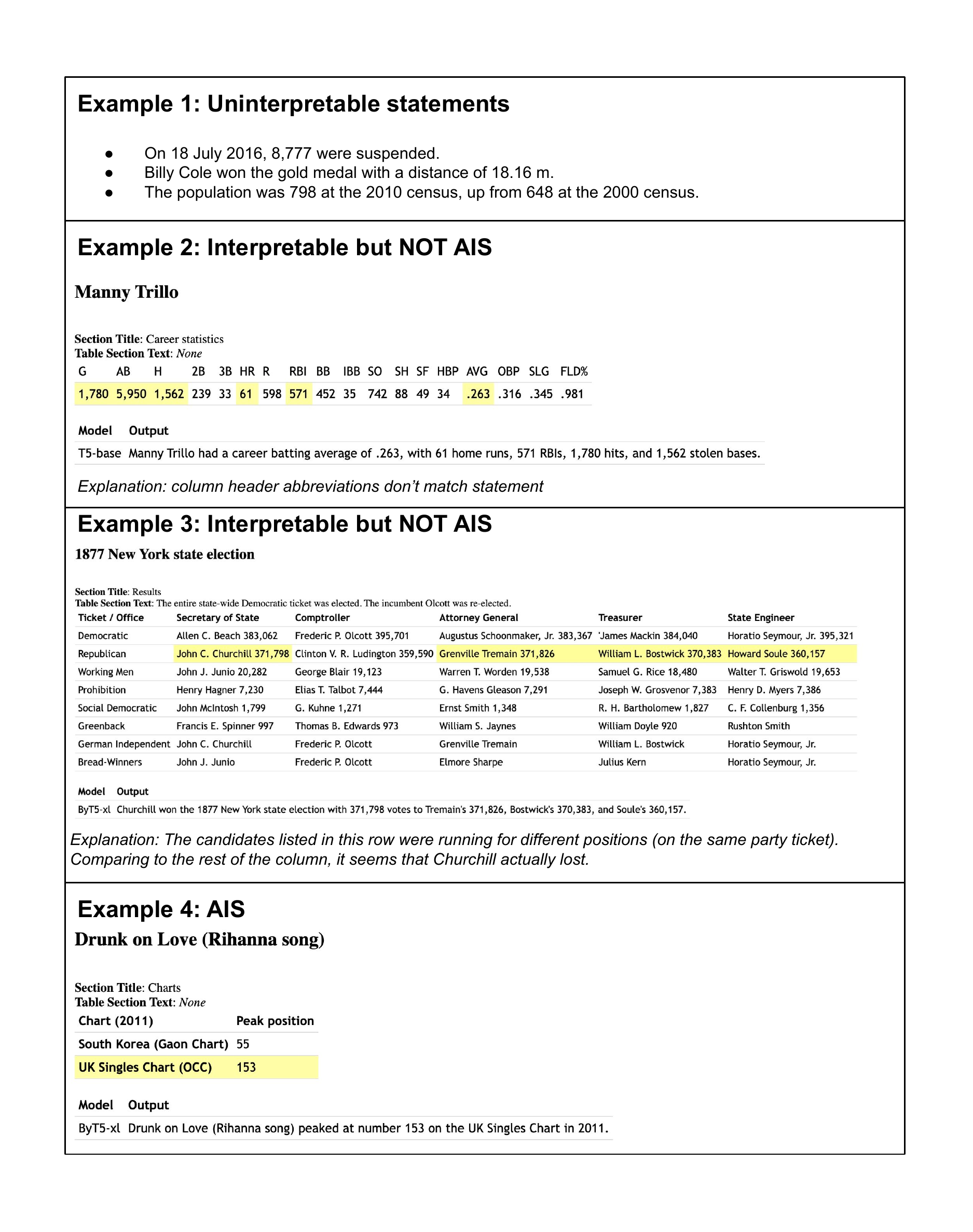}
    \caption{Examples from the table-to-text annotations.}
    \label{fig:ttt:ex}
\end{figure}

\definecolor{Instructions}{HTML}{ff8f80}
\onecolumn
\appendixsection{Evaluation Instructions for Conversational Question Answering}\label{app:instructions_qa}

\emph{The following is a verbatim representation of the instructions that were presented to paid crowd annotators for performing the task alongside the interface. The prompts in the rating interface include wording from the instructions; the rating interface also contains hyperlinks to example sections in the instructions for each question and rating.}

\section*{\textcolor{Instructions}{Overview}}
In this task you will evaluate the quality of a system-generated \textbf{response} to a \textbf{user query}. The system is trying to help the user learn about a particular topic by answering their questions. We want to rate the system response quality based on how well it represents the \textbf{original source}.

We will be using two categories to evaluate the quality of the summary: \textbf{Interpretability} and \textbf{Attribution}. You will evaluate these categories in succession. Some ratings will result in other categories being skipped. The task interface will guide you through the flow; you can also see the overall task flow in the diagram below.

\begin{figure}[htbp]
  \centering
  \includegraphics[width=3in]{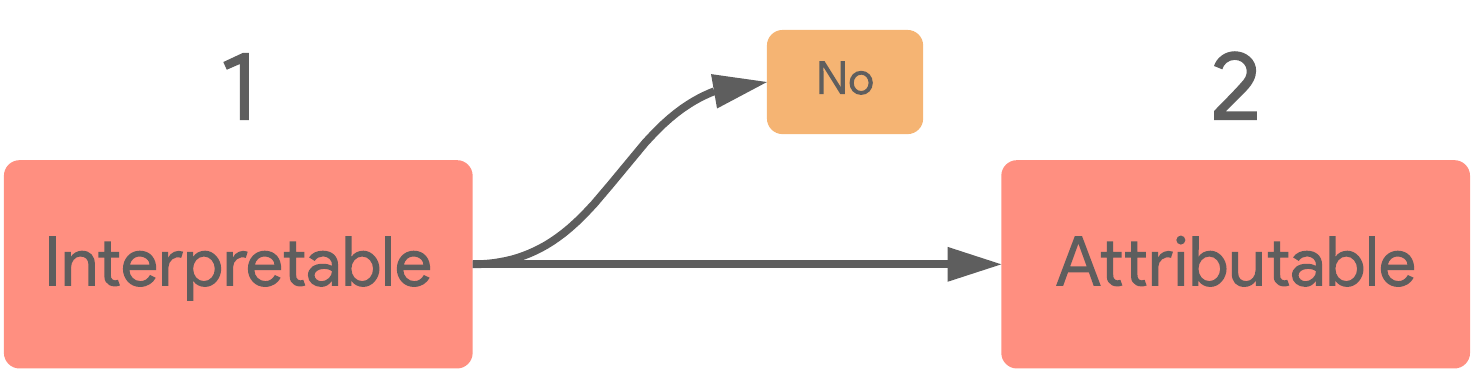}
\end{figure}

\textbf{Note}: The system-generated responses may appear very fluent and well-formed, but contain slight inaccuracies that are not easy to discern at first glance. Pay close attention to the text. Read it carefully as you would when proofreading.

The sections below describe each of the dimensions in detail. You can also \textbf{flag} ineligible tasks; the flagging criteria are described in  
\hyperref[sec:instructions_qa_flag]{this section}.

\section*{\textcolor{Instructions}{1. Interpretability}}\label{sec:instructions_qa_interpretability}

In this step you will evaluate whether the system response is \textbf{interpretable} by you. 

You will be shown an excerpt of a conversation between a user and an assistant-like computer system.  The last turn in the conversation will be a \textbf{user query} from the user followed by a \textbf{system response} attempting to respond to their query. Given the context of the user query, carefully read the system response and answer the following question:

\begin{itemize}
\item[\textit{(Q1)}] \textit{Is all of the information relayed by the system response \textbf{interpretable} to you?}
\end{itemize}

This is asking whether you can understand the response. If there is some part of the response that is unclear or hard to interpret, select “No”. If prompted by the interface, enter a succinctly detailed justification of your rating. 

\paragraph{Definition:} An \textbf{uninterpretable} response has diminished intelligibility due to:

\begin{itemize}
    \item Vague or ambiguous meaning, e.g., unclear pronouns usage.
    \item Malformed phrases and sentences that are difficult to understand.
\end{itemize}

If the system response is interpretable by you, you will proceed to the next category.

Examples of interpretability ratings and justifications are in \hyperref[sec:instructions_qa_interpretabilty_examples]{this section}.

\section*{\textcolor{Instructions}{2. Attribution}}

In this step, you will evaluate how well a system-generated response is \textbf{attributable} to the source document. Note that the source document is a new element that will appear in the task only when you reach this question.
 
\medskip
\footnotesize{\color{darkgray}\textbf{Note}: We refer to “\textbf{attributable to source document}” interchangeably as “\textbf{attribution}” and “\textbf{supported by the source document}”. By which we mean, \textbf{all} of the information in the system response can be verified from the source document.}

\medskip
\normalsize
You will be shown an excerpt of a conversation between a user and an assistant-like computer system.  The last turn in the conversation will be a \textbf{user query} from the user followed by a \textbf{system response} attempting to reply to their query. You will also be shown a document that was cited by the system as its source in attempting to answer the question (\textbf{source document}). You will use all three (user query, system response, source document) to answer the following question:

\begin{itemize}
\item[\textit{(Q2)}] \textit{Is all of the information provided by the system response \textbf{fully supported} by the source document?}
\end{itemize}

This is asking whether all of the information in the system response can be attributed to the information present in the source document. If prompted by the interface, enter a succinctly detailed justification of your rating. 

\paragraph{{Definition:}} Attribution of system-generated response in relation to the source document can be established by considering the following:

\begin{enumerate}
    \item[A.] What is the \textbf{information provided by the system response?}
    \item[B.] Is this information an \textbf{accurate representation of information in the source document?}
\end{enumerate}

\subsubsection*{\textcolor{Instructions}{A. Definition of “the Information Provided by the System Response”}}

Two points are key in determining the information provided by the system response:

\begin{enumerate}
    \item The context of the system response — that is, the query and previous conversation turns — is often critical in determining the “information provided by the system response”.
    \item The source document should be completely ignored when determining “the information provided by the system response.” (i.e., it should not be used as additional context).
\end{enumerate}

Consider the following example:

\medskip
\small
    \begin{tabular}{lp{11.4cm}}
       \textbf{User query}\\ 
       \multicolumn{2}{p{11.4cm}}{
        who plays doug and julie in days of our lives
        \smallskip
        } \\
        \textbf{System response}\\ \multicolumn{2}{p{11.4cm}}{\textcolor{teal}{
        In the American daytime drama Days of Our Lives, Doug Williams and Julie Williams are portrayed by Bill Hayes and Susan Seaforth Hayes}	
        } \\
    \end{tabular}

\medskip
\normalsize
In the above example, the meaning of the system response is clear even without seeing the query. But consider another example:

\medskip
\small
    \begin{tabular}{lp{11.4cm}}
        \textbf{User query}\\ \multicolumn{2}{p{11.4cm}}{
        who plays doug and julie in days of our lives
        \smallskip
        } \\
        \textbf{System response}\\ \multicolumn{2}{p{11.4cm}}{\textcolor{teal}{
        he is played by Bill Hayes}	
        } \\
    \end{tabular}

\medskip
\normalsize
In this case the pronoun “he” depends on the context (i.e., the query): but it is clear that the intended meaning of the system response can be paraphrased as something along the lines of “Doug Williams in days of our lives is played by Bill Hayes”. In this case this paraphrase is the “information provided by the system response”. 

Pronouns such as he/she/it/they etc. are one case where context is needed to figure out the intended meaning of the system response. Other examples are the following (given with paraphrases of the information that is provided by the system response):

\medskip
\small
    \begin{tabular}{lp{14cm}}
        \textbf{User query 1}\\ \multicolumn{2}{p{15cm}}{
        which network is days of our lives on
        \smallskip
        } \\
        \textbf{System response 1}\\ \multicolumn{2}{p{15cm}}{\textcolor{teal}{
        the show is on NBC}	
        \smallskip
        } \\
        \textbf{Paraphrase of information provided by system response}\\ \multicolumn{2}{p{15cm}}{
        days of our lives is on NBC
        } \\
        \\
        \textbf{User query 2}\\ \multicolumn{2}{p{15cm}}{
        which network is days of our lives on
        \smallskip
        } \\
        \textbf{System response 2}\\ \multicolumn{2}{p{15cm}}{\textcolor{teal}{
        NBC}
        \smallskip
        } \\
        \textbf{Paraphrase of information provided by system response}\\ \multicolumn{2}{p{15cm}}{
        days of our lives is on NBC
        } \\
        \\
        \textbf{User query 3}\\ \multicolumn{2}{p{15cm}}{
        how many seasons are there of days of our lives
        \smallskip
        } \\
        \textbf{System response 3}\\ \multicolumn{2}{p{15cm}}{\textcolor{teal}{
        there are 56 seasons}
        \smallskip
        } \\
        \textbf{Paraphrase of information provided by system response}\\ \multicolumn{2}{p{15cm}}{
        there are 56 seasons of days of our lives
        } \\
        \\
    \end{tabular}

\medskip
\normalsize
In system response 1, the phrase “the show” needs context for interpretation, but it is clear from the context of the query that it refers to “days of our lives”. In system response 2, the system gives a direct answer to the query, simply “NBC”, but it is clear given the query that the information provided by the system is “days of our lives is on NBC”. In query 3, the phrase “56 seasons” needs context for interpretation, but given the query it is clear that the response is referring to “56 seasons of days of our lives”.

In general, use your best judgment to determine the information provided by the system response. If you are unsure what the intended meaning is of the system response, make sure that you marked the example as “\textbf{No, the response is unclear.}” as part of the \hyperref[sec:instructions_qa_interpretability]{Interpretability} stage. As one example, take the following:

\medskip
\small
    \begin{tabular}{lp{11.4cm}}
        \textbf{User query}\\ \multicolumn{2}{p{11.4cm}}{
        how many NBA championships did Michael Jordan win?
        \smallskip
        } \\
        \textbf{System response}\\ \multicolumn{2}{p{11.4cm}}{\textcolor{teal}{
         it is the best team in the NBA}
         \smallskip
        } \\
        \textbf{Paraphrase of information provided by system response}\\ \multicolumn{2}{p{11.4cm}}{
        \emph{meaning is unclear}
        } \\
    \end{tabular}

\medskip
\normalsize
In this case it is not clear what “it” is referring to, and the meaning should be marked as being unclear. Again, use your best judgment in determining whether or not the meaning of the system response is clear.

\subsubsection*{\textcolor{Instructions}{B. Definition of “An Accurate representation of Information in the Source Document”}}

Again, you should use your best judgment in determining whether all of the information provided by the system response is “an accurate representation of information in the source document”. We give the following guidance:

\begin{itemize}
    \item In determining this question, ask yourself whether it is accurate to say ``the document says\dots'' or ``according to the document\dots'' with the system response following this phrase. For example, is it accurate to say ``according to the document below, In the American daytime drama Days of Our Lives, Doug Williams and Julie Williams are portrayed by Bill Hayes and Susan Seaforth Hayes'' in the example given above?
    \item Be sure to check \textbf{all} of the information in the response. If only some of the information is supported in the document, but other parts of the information are missing from the document or not an accurate representation, then please mark ``\textbf{No, not fully attributable}.''
    \item The concept of ``accurate representation'' should be close to a journalist’s conception of this phrase. For example take this excerpt from \href{https://www.npr.org/about-npr/688139552/accuracy}{this page} on Accuracy in the NPR Ethics Handbook: ``When quoting or paraphrasing anyone\dots consider whether the source would agree with the interpretation\dots'' In other words, if you had written the source document, consider whether you would view the system response as an accurate representation of information in that source document.
\end{itemize}

\subsubsection*{\textcolor{Instructions}{Some Final Important Notes}}

\paragraph{Source quality.} When making your judgments in this template, \textbf{do not take into account whether the underlying source document is correct or trustworthy}. This is clearly important, but will be evaluated in a separate task. The “attribution” category is used only to judge whether the information provided by the system response is an accurate representation of the underlying source document.

\paragraph{Context and relevance.} Additionally, when rating attribution, \textbf{do not take into account the degree of relevance of the system response to the user query}. Partially and fully relevant responses can be equally assessed for attribution, regardless of how much information contained in them is relevant to the query. In both cases you should judge whether or not the system responses are an accurate representation of information in the source document even if it doesn’t perfectly address the question.

Examples of attribution ratings and justifications are in \hyperref[sec:instructions_qa_attribution_examples]{this section}.

\section*{\textcolor{Instructions}{Scoring and Examples}}

\subsection*{\textcolor{Instructions}{Interpretability}}\label{sec:instructions_qa_interpretabilty_examples}

\paragraph{No, the response is unclear:} The response is unclear and/or difficult to understand.

\medskip
\small
    \begin{tabular}{lp{11.4cm}}
        \textbf{User query}\\ \multicolumn{2}{p{11.4cm}}{
        when was the last time it snowed in england on christmas day
        } \\
    \end{tabular}

    \smallskip
    \begin{tabular}{lp{11.4cm}} 
        \textbf{System response}\\ \multicolumn{2}{p{11.4cm}}{\textcolor{teal}{
         it was in 2009, but it was white}	
         \smallskip
        } \\
    \end{tabular}

    \smallskip
    \begin{tabular}{lp{11.4cm}}
        \textbf{Justification}\\ \multicolumn{2}{p{11.4cm}}{
        \emph{The answer contains the pronoun “it” with an unclear meaning; it also sets up a contrasting statement with “but”, which is hard to interpret.}
        } \\
    \end{tabular}

    \smallskip
    \begin{tabular}{lp{11.4cm}}
        \textbf{User query}\\ \multicolumn{2}{p{11.4cm}}{
        who sings in walk hard the dewey cox story
        \smallskip
        } \\
    \end{tabular}

    \smallskip
    \begin{tabular}{lp{11.4cm}}
        \textbf{System response}\\ \multicolumn{2}{p{11.4cm}}{\textcolor{teal}{
         he and john c. Reilly sang all the songs and played guitar on most of them.}
         \smallskip
        } \\
    \end{tabular}

    \smallskip
    \begin{tabular}{lp{11.4cm}}
        \textbf{Justification}\\ \multicolumn{2}{p{11.4cm}}{
        \emph{The answer contains the pronoun “he” with an unclear meaning.}
        } \\
    \end{tabular}

    \smallskip
    \begin{tabular}{lp{11.4cm}}
        \textbf{User query}\\ \multicolumn{2}{p{11.4cm}}{
         what event marks the beginning of the great depression
         \smallskip
        } \\
        \textbf{System response}\\ \multicolumn{2}{p{11.4cm}}{\textcolor{teal}{
         after a major fall in stock prices that began around september 4, 1929, and became worldwide news with the stock market crash of October 29, 1929}
         \smallskip
        } \\
        \textbf{Justification}\\ \multicolumn{2}{p{11.4cm}}{
        \emph{The answer is malformed and difficult to interpret: the sentence appears to be missing the subject somewhere near “and became”.}
        } \\
    \end{tabular}

\medskip    
\normalsize    
\hrule
\subsection*{\textcolor{Instructions}{Attribution}}\label{sec:instructions_qa_attribution_examples}

\paragraph{Yes, fully attributable:} All of the information in the system response is supported by the document. 

\medskip
\small
    \begin{tabular}{lp{11.4cm}}
        \textbf{User query}\\ \multicolumn{2}{p{11.4cm}}{
        when did the subway open in new york
        \smallskip
        } \\
        \textbf{System response}\\ \multicolumn{2}{p{11.4cm}}{\textcolor{teal}{
         the subway was first opened in 1904}
         \smallskip
        } \\
        \textbf{Source document}\\ \multicolumn{2}{p{11.4cm}}{
        The New York City Subway is a rapid transit system owned by the City of New York and leased to the New York City Transit Authority, a subsidiary agency of the state-run Metropolitan Transportation Authority (MTA). Opened in 1904, the New York City Subway is one of the world's oldest public transit systems, one of the world's most used metro systems, and the metro system with the most stations. It offers service 24 hours per day on every day of the year.
        \smallskip
        } \\
        \textbf{Justification}\\ \multicolumn{2}{p{11.4cm}}{
        \emph{The information in the response is accurately representing the document.}
        } \\
    \end{tabular}
\medskip    
\normalsize
\paragraph{No, not fully attributable:} The response contains any amount of information that is not supported by the document (including responses that are only partially or not at all supported).

\medskip
\small
    \begin{tabular}{lp{11.4cm}}
        \textbf{User query}\\ \multicolumn{2}{p{11.4cm}}{
        when did the subway open in new york
        \smallskip
        } \\
        \textbf{System response}\\ \multicolumn{2}{p{11.4cm}}{\textcolor{teal}{
          the subway was first opened in late 1904.}
          \smallskip
        } \\
        \textbf{Source document}\\ \multicolumn{2}{p{11.4cm}}{
        The New York City Subway is a rapid transit system owned by the City of New York and leased to the New York City Transit Authority, a subsidiary agency of the state-run Metropolitan Transportation Authority (MTA). Opened in 1904, the New York City Subway is one of the world's oldest public transit systems, one of the world's most used metro systems, and the metro system with the most stations. It offers service 24 hours per day on every day of the year.
        \smallskip
        } \\
        \textbf{Justification}\\ \multicolumn{2}{p{11.4cm}}{
        \emph{The document confirms that it was opened in 1904 but it doesn’t specify if it was “late 1904”.}
        } \\
    \end{tabular}

    \smallskip
    \begin{tabular}{lp{11.4cm}}
        \textbf{User query}\\ \multicolumn{2}{p{11.4cm}}{
        who has the most all star mvp awards
        \smallskip
        } \\
        \textbf{System response}\\ \multicolumn{2}{p{11.4cm}}{\textcolor{teal}{
        bob pettit, kobe bryant, oscar robertson, michael jordan, shaquille o'neal, and lebron james have all won the award four times.}
        \smallskip
        } \\
        \textbf{Source document}\\ \multicolumn{2}{p{11.4cm}}{
         Bob Pettit and Kobe Bryant are the only two players to win the All-Star Game MVP four times. Oscar Robertson, Michael Jordan, Shaquille O'Neal, and LeBron James have each won the award three times.
         \smallskip
        } \\
        \textbf{Justification}\\ \multicolumn{2}{p{11.4cm}}{
        \emph{The cited document only supports Bob Petit and Kobe Bryant. The others listed only won the award \textbf{three} times according to the document.}
        } \\
    \end{tabular}

    \smallskip
    \begin{tabular}{lp{11.4cm}}
        \textbf{User query}\\ \multicolumn{2}{p{11.4cm}}{
        who expanded the palace of versailles to its present size
        \smallskip
        } \\
        \textbf{System response}\\ \multicolumn{2}{p{11.4cm}}{\textcolor{teal}{
        louis xiv, his successor, was a very good person.}
        \smallskip
        } \\
        \textbf{Source document}\\ \multicolumn{2}{p{11.4cm}}{
         Louis XIII's successor, Louis XIV, had a great interest in Versailles. He settled on the royal hunting lodge at Versailles, and over the following decades had it expanded into one of the largest palaces in the world. Beginning in 1661, the architect Louis Le Vau, landscape architect André Le Nôtre, and painter-decorator Charles Lebrun began a detailed renovation and expansion of the château. This was done to fulfill Louis XIV's desire to establish a new centre for the royal court. Following the Treaties of Nijmegen in 1678, he began to gradually move the court to Versailles. The court was officially established there on 6 May 1682.
         \smallskip
        } \\
        \textbf{Justification}\\ \multicolumn{2}{p{11.4cm}}{
        \emph{The answer is correct but the last part (he was a “very good person”) is an opinion and not supported by the document.}
        } \\
    \end{tabular}

\medskip    
\normalsize    
\hrule
\subsection*{\textcolor{Instructions}{Flag}}\label{sec:instructions_qa_flag}

There is a flag button in the bottom left corner of the task interface. Once flagged, you can proceed onto the next task. Use it report tasks that are ineligible for reasons such as:

\paragraph{1. Missing components in the task.} Some tasks may have missing user queries, responses or source text. They should be flagged.

\paragraph{2. Malformed text.} Some text may be severely malformed with unintelligible artifacts (e.g. html code, unformatted tables, etc.). If any component of the task contains malformed text, the task should be flagged. The example below shows various types of malformed text.

\medskip
\small
    \begin{tabular}{lp{11.4cm}}
        \textbf{User query}\\ \multicolumn{2}{p{11.4cm}}{
        where did remember the titans camp take place
        \smallskip
        } \\
        \textbf{System response}\\ \multicolumn{2}{p{11.4cm}}{\textcolor{teal}{
        i think it took place in a gymnasium in san diego, where the black students auditioned to play for the \hl{\$team\_winner}  until boone arrives, but the meeting turns into a fiasco when yoast and white students interrupt.}
        \smallskip
        } \\
        \textbf{Source document}\\
        \multicolumn{2}{p{11.4cm}}{
        The black students have a meeting in the gymnasium in auditioning to play for the team until \hl{\&amp;\&amp;\&amp;\&amp;}Boone\hl{\&amp;\&amp;\&amp;\&amp;\&amp;} arrives, but the meeting turns into a fiasco when Yoast and white students interrupt. On August 15, 1971, the players gather and journey to Gettysburg College, where their training camp takes place.
        } \\
    \end{tabular}
\medskip    
\normalsize    
\paragraph{3. Source document is underspecified.} Some source documents may not have sufficient information to determine whether it does/doesn’t support the response. They may be too short and/or lack critical information that would be necessary to rate the response.

\medskip
\small
    \begin{tabular}{lp{11.4cm}}
        \textbf{User query}\\ \multicolumn{2}{p{11.4cm}}{
        where is the hero golf challenge being played
        \smallskip
        } \\
        \textbf{System response}\\ \multicolumn{2}{p{11.4cm}}{\textcolor{teal}{
        the event will be held at the Albany development in the Bahamas}
        \smallskip
        } \\
        \textbf{Source document}\\ \multicolumn{2}{p{11.4cm}}{
        Beginning in 2015, \hl{the event} will be played at the Albany development in the Bahamas, for at least three years.
        \smallskip
        } \\
        \textbf{Justification}\\ \multicolumn{2}{p{11.4cm}}{
        \emph{The source document does not specify what event it is referring to, and so it’s impossible to determine whether this is referring to the same “hero golf challenge” or not . Without the explicit mention of “hero gold challenge”, the task should be flagged.}
        } \\
    \end{tabular}

    \smallskip
    \begin{tabular}{lp{11.4cm}}
        \textbf{User query}\\ \multicolumn{2}{p{11.4cm}}{
        who is known as the super fast boy in the series the icredible
        } \\
        \textbf{System response}\\ \multicolumn{2}{p{11.4cm}}{\textcolor{teal}{
        spencer fox as dashiell robert parr}
        \smallskip
        } \\
        \textbf{Source document}\\ \multicolumn{2}{p{11.4cm}}{
        Spencer Fox as Dashiell Robert Parr / Dash, the Parrs' second child, who possesses super-speed
        \smallskip
        } \\
        \textbf{Justification}\\ \multicolumn{2}{p{11.4cm}}{
        \emph{It’s unclear whether the source document is referring to the same character as the one in the query because it does not specify the series (“the incredible”) or the name of the character (“the super fast boy”).}
        } \\
    \end{tabular}
\medskip
\normalsize

\paragraph{4. Source document is difficult to understand because it requires expert-level knowledge.} Some documents may include scientific formulas, obscure terminology, etc. If you can still understand enough of the document to rate the attribution, please do so. But, on the other hand, if properly evaluating the response requires expertise in a particular area, please flag it.

\medskip
\small
    \begin{tabular}{lp{11.4cm}}
        \textbf{User query}\\ \multicolumn{2}{p{11.4cm}}{
        what is the rate limiting enzyme of kreb's cycle
        \smallskip
        } \\
        \textbf{System response}\\ \multicolumn{2}{p{11.4cm}}{\textcolor{teal}{
        kreb is the enzyme that uses fad as a prosthetic group}	
        \smallskip
        } \\
        \textbf{Source document}\\
        \multicolumn{2}{p{11.4cm}}{
        Substrates Products Enzyme Reaction type Comment 0 / 10 Oxaloacetate + Acetyl CoA + H2O Citrate + CoA-SH Citrate synthase Aldol condensation irreversible, extends the 4C oxaloacetate to a 6C molecule 1 Citrate cis-Aconitate + H2O Aconitase …
        \smallskip
        } \\
        \textbf{Justification}\\ \multicolumn{2}{p{11.4cm}}{
        \emph{In order to be able to evaluate whether this source document supports the response, it requires a deeper understanding of scientific equations and terminology contained in the document.  Because this example requires scientific expertise to evaluate it properly, it should be flagged.}
        } \\
    \end{tabular}
\medskip    
\normalsize  

\definecolor{Instructions}{HTML}{ff8f80}
\appendixsection{Evaluation Instructions for Summarization}\label{sec:instructions_summarization}

\emph{The following is a verbatim representation of the instructions that were presented to paid crowd annotators for performing the task alongside the interface. The prompts in the rating interface include wording from the instructions; the rating interface also contains hyperlinks to example sections in the instructions for each question and rating. The summarization instructions were developed after the conversational QA instructions had been established and the annotators had been trained on the conversation QA task.}

\section*{\textcolor{Instructions}{Overview}}
In this task you will evaluate the quality of a system-generated \textbf{summary}. The system’s goal is to summarize the \textbf{source news article}, while remaining truthful to it. We want to rate the quality of the summary based on how well it represents the original source.

We will be using two categories to evaluate the quality of the summary: \textbf{Interpretability} and \textbf{Attribution}. You will evaluate these categories in succession. Some ratings will result in other categories being skipped. The task interface will guide you through the flow; you can also see the overall task flow in the diagram below.

\begin{figure}[htbp]
  \centering
  \includegraphics[width=3in]{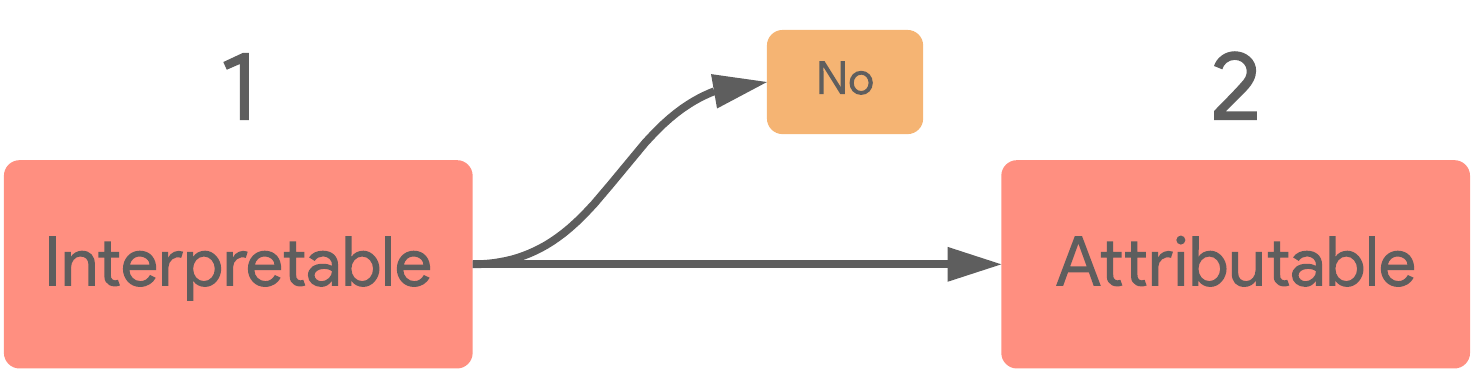}
\end{figure}

\textbf{Note}: The system-generated summaries may appear very fluent and well-formed, but contain slight inaccuracies that are not easy to discern at first glance. Pay close attention to the text. Read it as carefully as you would when proofreading.

The sections below describe each of the dimensions in detail. You can also \textbf{flag} ineligible tasks; the flagging criteria are described in  
\hyperref[sec:instructions_summ_flag]{this section}.

\section*{\textcolor{Instructions}{1. Interpretability}}\label{sec:instructions_summ_interpretability}

In this step you will evaluate whether the system summary is \textbf{interpretable} by you. 

You will be shown a system-generated \textbf{summary} of a news article. Note that the news article from which the summary is derived is hidden at this step, because we need to evaluate whether the summary is \textit{interpretable on its own}. Carefully read the summary and answer the following question:

\begin{itemize}
\item[\textit{(Q1)}] \textit{Is all of the information relayed by the system summary \textbf{interpretable} to you?}
\end{itemize}

This is asking whether you can understand the summary on its own. If there is any part of the summary that is unclear or hard to interpret, select “No”. If prompted by the interface, enter a succinctly detailed justification of your rating. 

\paragraph{Definition:} An \textbf{uninterpretable} summary has diminished intelligibility due to:

\begin{itemize}
    \item Vague or ambiguous meaning, e.g., unclear noun references or pronouns usage.
    \item Malformed phrases and sentences that are difficult to understand.
\end{itemize}

If the summary is interpretable by you, you will proceed to the next category.

In the section below, we show in more detail the kind of reasoning that should be used for establishing interpretability of summaries. More examples of interpretability ratings and justifications are in \hyperref[sec:instructions_summ_interpretabilty_examples]{this section} of the appendix.

\subsubsection*{\textcolor{Instructions}{Interpreting the information provided in the system summary}}

Consider the following example:

\medskip
\small
    \begin{tabular}{lp{11.4cm}}
        \textbf{Summary}\\ \multicolumn{2}{p{11.4cm}}{\textcolor{teal}{
         seismologists put the magnitude at 7.9 for an earthquake that hit kathmandu today, which would actually make it about 40\% larger than the 7.8 currently being reported .}	
        } \\
    \end{tabular}

\medskip
\normalsize
In the above example, the meaning of the summary is clear even without seeing the original news article.  It is clear what the summary is reporting on and it stands on its own, that is, this summary is \textbf{interpretable}. It should be marked as “\textbf{Yes, I understand it.}” But consider another example:

\medskip
\small
    \begin{tabular}{lp{11.4cm}}
        \textbf{Summary}\\ \multicolumn{2}{p{11.4cm}}{\textcolor{teal}{
         seismologists put the magnitude at 7.9 , which would actually make it about 40\% larger than the 7.8 currently being reported .}	
        } \\
    \end{tabular}

\medskip
\normalsize
In this case the meaning of the phrase “the magnitude” obviously depends on some context, but that context is missing in the summary. Without additional information that clarifies that the magnitude refers to an earthquake that occurred in a specific location (Kathmandu), the summary is \textbf{difficult to interpret} and it does not stand on its own. It should be marked as “\textbf{No, the summary is unclear.}”

Noun phrases that require clarifications of this kind are one case where interpretability can be diminished. Other examples include nouns and pronouns without a (clear) reference and malformed phrases and sentences:

\medskip
\small
    \begin{tabular}{lp{11.4cm}}
        \textbf{Summary 1}\\ \multicolumn{2}{p{11.4cm}}{\textcolor{teal}{
        \hl{the project} is hoped to open at the former arts centre , la llotja . friend and producer jaume roures of mediapro is leading the tribute . \hl{the museum} follows allen 's vicky cristina barcelona, set in \hl{the city}}
        } \\
        \\
        \textbf{Summary 2}\\ \multicolumn{2}{p{11.4cm}}{\textcolor{teal}{
        new england patriots tight end aaron hernandez has pleaded not guilty to murder and two weapons charges . he 's accused of orchestrating the shooting death of odin lloyd . \hl{it} 's scheduled to begin in may , \hl{but not legally required to get a conviction .}}	
        } \\
        \\
        \textbf{Summary 3}\\ \multicolumn{2}{p{11.4cm}}{\textcolor{teal}{
        john stamos \hl{announced} monday night on `` jimmy kimmel live '' . the show will feature candace cameron bure , who played eldest daughter d.j . tanner in the original series , which aired from 1987 to 1995 , \hl{will both return} for the new series .}	
        } \\
        \\
    \end{tabular}

\medskip
\normalsize
In summary 1, the phrase “the project” needs context for interpretation (“\textit{what project is being reported on?}”). Likewise, “the museum” and “the city” are unclear (“\textit{what museum is this?}”, “\textit{what city is this taking place in?}”). In summary 2, the system provides a clear reference for the pronoun “he” (“hernandez”), but a reference for the pronoun “it” is missing (“\textit{what is scheduled to begin in may?}”). Also it is not clear what “not legally required to get a conviction” is referring to. In summary 3, the first sentence is malformed because it is missing the announcement (“\textit{what did john stamos announce?}”). The second sentence is difficult to understand (“\textit{who does ‘both’ refer to?}”, “\textit{who will return to the new series?}”).

In general, use your best judgment to determine the information provided by the summary. If you are unsure what the intended meaning of the summary is, err on the side of marking it with “\textbf{No, the summary is unclear.}”

\section*{\textcolor{Instructions}{2. Attribution}}

In this step, you will evaluate how well a system-generated summary is \textbf{attributable} to the source news article. Note that the source news article is a new element that will appear in the task only when you reach this question.
 
\medskip
\footnotesize{\color{darkgray}\textbf{Note}:  We refer to “\textbf{attributable to the source news article}” interchangeably as “\textbf{attribution}” and “\textbf{supported by the source news article}”. By which we mean, \textbf{all} of the information in the system-generated summary can be verified from the source news article.}

\medskip
\normalsize
You will be shown a system-generated \textbf{summary} of a news article. You will also be shown the news article that was used by the system to generate this summary (\textbf{source news article}). You will use both of these to answer the following question:

\begin{itemize}
\item[\textit{(Q2)}] \textit{Is all of the information provided by the system summary \textbf{fully supported} by the source document?}
\end{itemize}

This is asking whether all of the information in the system summary can be attributed to the information present in the source news article. If prompted by the interface, enter a succinctly detailed justification of your rating. 

\paragraph{{Definition:}} a \textbf{fully supported} (or \textbf{attributable}) system-generated summary contains an accurate representation of information in the source news article. No information in the summary is unattested when compared against the source news article.

In the section below, we show in more detail the kind of reasoning that should be used for establishing attribution of summaries. More examples of attribution ratings and justifications are in \hyperref[sec:instructions_summ_attribution_examples]{this section} of the appendix.

\subsubsection*{\textcolor{Instructions}{Assessing the accuracy of the information in the summary against the original news article}}

Again, you should use your best judgment in determining whether all of the information provided by the system summary is “an accurate representation of information in the source news article”. We give the following guidance:

\begin{itemize}
    \item In determining this question, ask yourself whether it is accurate to say “the provided news article says\dots” or “according to the news article\dots” with the system summary following this phrase.
    \item Be sure to check \textbf{all} of the information in the summary. If only some of the information is supported in the news article, but other parts of the information are missing from the news article or not an accurate representation, then please mark “\textbf{No, not fully attributable}.”
    \item The concept of “accurate representation” should be close to a journalist’s conception of this phrase. For example take this excerpt from \href{https://www.npr.org/about-npr/688139552/accuracy}{this page} on Accuracy in the NPR Ethics Handbook: “When quoting or paraphrasing anyone\dots consider whether the source would agree with the interpretation\dots” In other words, if you had written the source document, consider whether you would view the summary as an accurate representation of information in that source document.
\end{itemize}

\subsubsection*{\textcolor{Instructions}{Some Final Important Notes}}

\paragraph{Source quality.} When making your judgments in this template, \textbf{do not take into account whether the underlying source news article is correct or trustworthy}. This is clearly important, but will be evaluated in a separate task. The “attribution” category is used only to judge whether the information provided by the system summary is an accurate representation of the underlying source news article.

Examples of attribution ratings and justifications are in \hyperref[sec:instructions_qa_attribution_examples]{this section}.

\section*{\textcolor{Instructions}{Scoring and Examples}}

\subsection*{\textcolor{Instructions}{Interpretability}}\label{sec:instructions_summ_interpretabilty_examples}

\paragraph{No, the summary is unclear:} The summary is unclear and/or difficult to understand. 

\medskip
\small
    \begin{tabular}{lp{11.4cm}}
        \textbf{Summary}\\ \multicolumn{2}{p{11.4cm}}{\textcolor{teal}{
        The bill would prevent adolescents from smoking, buying or possessing both traditional and electronic cigarettes . The bill includes a \$10 fine for first-time offenders. Subsequent violations would lead to a \$50 fine or mandatory community service . Dozens of local governments have similar bans, including Hawaii County and New York City .}	
        \smallskip
        } \\
    \end{tabular}

    \smallskip
    \begin{tabular}{lp{11.4cm}}
        \textbf{Justification}\\ \multicolumn{2}{p{11.4cm}}{
        \emph{The summary revolves around the noun phrase “the bill” that doesn’t have a clear reference (what bill is it referring to?); it makes the summary difficult to understand fully.}
        } \\
    \end{tabular}

    \medskip
    \begin{tabular}{lp{11.4cm}}
        \textbf{Summary}\\ \multicolumn{2}{p{11.4cm}}{\textcolor{teal}{
        The former England rugby star tweeted ‘Holy s***, I’m lost for words and emotions. All I can say is yes the dude!!!!!’ After he bought the horse Zara called him an idiot, but she was there with him, cheering the horse home . Monbeg Dude was given the all-clear to run in the Grand National at Aintree after a poor showing at the Cheltenham Festival .}	
        \smallskip
        } \\
    \end{tabular}

    \smallskip
    \begin{tabular}{lp{11.4cm}}
        \textbf{Justification}\\ \multicolumn{2}{p{11.4cm}}{
        \emph{The answer contains the noun phrase “the former England rugby star” that lacks a clear reference (who is it?). Furthermore, the usage of pronouns “he”, “him” and “she” is too vague (who are they referring to in the summary?).}
        } \\
    \end{tabular}

    \medskip
    \begin{tabular}{lp{11.4cm}}
        \textbf{Summary}\\ \multicolumn{2}{p{11.4cm}}{\textcolor{teal}{
        john stamos announced monday night on `` jimmy kimmel live '' . the show will feature candace cameron bure , who played eldest daughter d.j . tanner in the original series , which aired from 1987 to 1995 , will both return for the new series .}
        \smallskip
        } \\
    \end{tabular}

    \smallskip
    \begin{tabular}{lp{11.4cm}}
        \textbf{Justification}\\  \multicolumn{2}{p{11.4cm}}{
        \emph{The answer is malformed and difficult to interpret. The first sentence is missing the object of “announced” (what did john stamos announce?). Additionally, “will both return for the new series” appears to be poorly connected to the previous context.}
        } \\
        \\
    \end{tabular}

\medskip    
\normalsize    
\hrule
\subsection*{\textcolor{Instructions}{Attribution}}\label{sec:instructions_summ_attribution_examples}

\paragraph{Yes, fully attributable:} All of the information in the system summary is supported by the document.

\medskip
\small
    \begin{tabular}{lp{11.4cm}}
        \textbf{Summary}\\ \multicolumn{2}{p{11.4cm}}{\textcolor{teal}{
         Convicted murderer Nikko Jenkins, 28, tried to carve '666' into his forehead . But in a phenomenal case of idiocy, he used a mirror - so the numbers came out backwards . The symbol is described in the biblical book of Revelation as 'the sign of the beast', and has since been popularized by the horror movie The Omen . Jenkins was jailed exactly one year ago for shooting dead four people in 10 days after being released from prison .}	
        } \\
    \end{tabular}

    \smallskip
    \begin{tabular}{lp{11.4cm}}
        \textbf{Original news article}\\ \multicolumn{2}{p{11.4cm}}{
        It was meant to be the ultimate symbol of menace: carving '666' into his forehead.
        } \\
    \end{tabular}

    \smallskip
    \begin{tabular}{lp{11.4cm}}
        \multicolumn{2}{p{11.4cm}}{
        But in a phenomenal case of idiocy, convicted murderer Nikko Jenkins used a mirror - so the numbers came out backwards.
        } \\
    \end{tabular}

    \smallskip
    \begin{tabular}{lp{11.4cm}}
        \multicolumn{2}{p{11.4cm}}{
        The symbol is described in the biblical book of Revelation as 'the sign of the beast', and has since been popularized by the horror movie The Omen. 
        } \\
    \end{tabular}

    \smallskip
    \begin{tabular}{lp{11.4cm}}
        \multicolumn{2}{p{11.4cm}}{
        However, with a series of upside-down 9s, Jenkins has fashioned himself an entirely unique - and irreversible - engraving.
        } \\
    \end{tabular}

    \smallskip
    \begin{tabular}{lp{11.4cm}}
        \multicolumn{2}{p{11.4cm}}{
        Botched: Nikko Jenkins (pictured in 2014) recently tried to carve '666' into his forehead but did it backwards . 
        } \\
    \end{tabular}

    \smallskip
    \begin{tabular}{lp{11.4cm}}
        \multicolumn{2}{p{11.4cm}}{
        According to Omaha.com, Jenkins told his attorney about the incident in a phone call from his cell in Omaha, Nebraska.
        } \\
    \end{tabular}

    \smallskip
    \begin{tabular}{lp{11.4cm}}
        \multicolumn{2}{p{11.4cm}}{
        It comes amid the 28-year-old's ongoing appeal that he is mentally unstable and therefore ineligible to face the death penalty.
        } \\
    \end{tabular}

    \smallskip
    \begin{tabular}{lp{11.4cm}}
        \multicolumn{2}{p{11.4cm}}{
        Jenkins was jailed exactly one year ago for shooting dead four people in 10 days after being released from prison.
        } \\
    \end{tabular}

    \smallskip
    \begin{tabular}{lp{11.4cm}}
        \multicolumn{2}{p{11.4cm}}{
        During his murder trial in Douglas County, Jenkins was assessed by a doctor who concluded that he was 'a psychopath' and 'one of the most dangerous people' he had ever encountered. 
        } \\
    \end{tabular}

    \smallskip
    \begin{tabular}{lp{11.4cm}}
        \multicolumn{2}{p{11.4cm}}{
        Psychopath': The 28-year-old, who a doctor described as 'one of the most dangerous people' he had ever encountered, may use the botched case of self-mutilation as evidence he is mentally unstable .
        } \\
    \end{tabular}

    \smallskip
    \begin{tabular}{lp{11.4cm}}
        \multicolumn{2}{p{11.4cm}}{
        Jenkins pleaded not guilty, then guilty, then ineligible for trial on the grounds of insanity. However, a judge dismissed the appeals and he was sentenced to life.
        } \\
    \end{tabular}

    \smallskip
    \begin{tabular}{lp{11.4cm}}
        \multicolumn{2}{p{11.4cm}}{
        The decision of whether he would be sentenced to death was delayed after Jenkins revealed he had carved a swastika into his skin.
        } \\
    \end{tabular}

    \smallskip
    \begin{tabular}{lp{11.4cm}}
        \multicolumn{2}{p{11.4cm}}{
        Following months of delays, he will face a panel in July to decide his fate.
        } \\
    \end{tabular}

    \smallskip
    \begin{tabular}{lp{11.4cm}}
        \multicolumn{2}{p{11.4cm}}{
        It is believed Jenkins may use his latest botched case of self-mutilation as further evidence that he is mentally unstable.
        } \\
    \end{tabular}

    \smallskip
    \begin{tabular}{lp{11.4cm}}
        \textbf{Justification}\\ \multicolumn{2}{p{11.4cm}}{
        \emph{The information in the summary accurately represents the information in the source news article.}
        } \\
    \end{tabular}
\medskip    
\normalsize
\paragraph{No, not fully attributable:} The summary contains any amount of information that is not supported by the source new article (including summaries that are only partially or not at all supported).

\medskip
\small
    \begin{tabular}{lp{11.4cm}}
        \textbf{Summary}\\ \multicolumn{2}{p{11.4cm}}{\textcolor{teal}{
         saracens lost 13-9 to clermont at stade geoffroy-guichard on saturday . the sarries pack contained five english-qualified forwards . saracens ' millionaire chairman nigel wray wants the salary cap scrapped .}	
        } \\
    \end{tabular}

    \smallskip
    \begin{tabular}{lp{11.4cm}}
        \textbf{Original news article}\\ \multicolumn{2}{p{11.4cm}}{
        saracens director of rugby mark mccall lauded his young guns after their latest european heartache before declaring he has no intention of overspending in a competitive post-world cup transfer market .
        } \\
    \end{tabular}

    \smallskip
    \begin{tabular}{lp{11.4cm}}
        \multicolumn{2}{p{11.4cm}}{
        mccall watched his side , which contained five english-qualified forwards in the starting pack , battle in vain before losing 13-9 to the clermont on saturday .
        } \\
    \end{tabular}

    \smallskip
    \begin{tabular}{lp{11.4cm}}
        \multicolumn{2}{p{11.4cm}}{
        saracens ' millionaire chairman nigel wray spent much of last week repeating his belief the cap should be scrapped in order for saracens to compete at europe 's top table , raising expectations they could be set to land a ` marquee ' player from outside the league whose wages would sit outside next season 's \# 5.5 m cap. 
        } \\
    \end{tabular}

    \smallskip
    \begin{tabular}{lp{11.4cm}}
        \multicolumn{2}{p{11.4cm}}{
        However, with a series of upside-down 9s, Jenkins has fashioned himself an entirely unique - and irreversible - engraving.
        } \\
    \end{tabular}

    \smallskip
    \begin{tabular}{lp{11.4cm}}
        \multicolumn{2}{p{11.4cm}}{
        maro itoje ( second left ) was one of five england-qualified forwards in the saracens pack that faced clermont mako vunipola tries to fend off clermont lock jamie cudmore during a ferocious contest saracens director of rugby mark mccall saw his side come agonisingly close to reaching the final but mccall said : ` we know where we 'd like to improve our side and we 're prepared to wait for the right person . we do n't want to jump in and get " a name " just because he 's available post-world cup .
        } \\
    \end{tabular}

    \smallskip
    \begin{tabular}{lp{11.4cm}}
        \multicolumn{2}{p{11.4cm}}{
        ` the fact our pack is as young as it is is incredibly exciting for us . they could be the mainstay of the club for the next four to five seasons . '
        } \\
    \end{tabular}

    \smallskip
    \begin{tabular}{lp{11.4cm}}
        \multicolumn{2}{p{11.4cm}}{
        billy vunipola ( left ) , jim hamilton and itoje leave the field following their 13-9 loss against clermont
        } \\
    \end{tabular}

    \smallskip
    \begin{tabular}{lp{11.4cm}}
        \textbf{Justification}\\  \multicolumn{2}{p{11.4cm}}{
        \emph{The summary includes unattributed information: “at stade geoffroy-guichard”, and the reference to the opposing team as the “the sarries”, which is not supported in the original new article. }
        } \\
        \\
    \end{tabular}

\medskip    
\normalsize    
\hrule
\subsection*{\textcolor{Instructions}{Flag}}\label{sec:instructions_summ_flag}

There is a flag button in the bottom left corner of the task interface. Once flagged, you can proceed onto the next task. Use it to report tasks that are ineligible for reasons such as:

\paragraph{1. Missing components in the task.} Some tasks may have missing summaries or news articles. They should be flagged.

\paragraph{2. Malformed text.} Some text may be severely malformed with unintelligible artifacts (e.g. html code, unformatted tables, etc.). If any component of the task contains malformed text that makes it difficult for you to accomplish the task, the task should be flagged.

\paragraph{3. Source document is difficult to understand because it requires expert-level knowledge.} Some documents may include scientific formulas, obscure terminology, etc. If you can still understand enough of the document to rate the attributability, please do so. But, on the other hand, if properly evaluating the summary requires expertise in a particular area, please flag it.
\definecolor{Instructions}{HTML}{ff8f80}
\appendixsection{Evaluation Instructions for Table-to-Text}\label{sec:instructions_t2t}

\emph{The following is a verbatim representation of the instructions that were presented to paid crowd annotators for performing the task alongside the interface. The prompts in the rating interface include wording from the instructions; the rating interface also contains hyperlinks to example sections in the instructions for each question and rating. The table-to-text instructions were developed after the conversational QA and summarization instructions had been established and the annotators had been trained on the conversation QA and summarization tasks.}

\section*{\textcolor{Instructions}{Overview}}

In this task you will evaluate the quality of a system-generated \textbf{caption} for highlighted parts of a table. The system is trying to convert the information in the table into a natural language description (what we are referring to as the “system-generated caption”). We want to rate the quality of the system-generated caption based on how well it represents information from the \textbf{source table}.

We will be using two categories to evaluate the quality of the caption: \textbf{Interpretability} and \textbf{Attribution}. You will evaluate these categories in succession. Some ratings will result in other categories being skipped. The task interface will guide you through the flow; you can also see the overall task flow in the diagram below. 

\begin{figure}[htbp]
  \centering
  \includegraphics[width=3in]{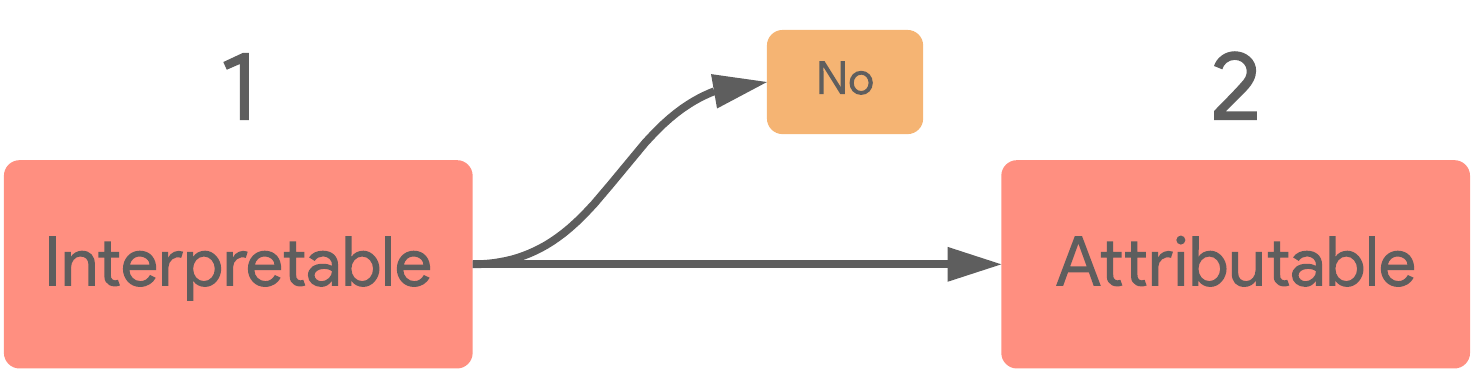}
\end{figure}

\textbf{Note}: The system-generated captions may appear very fluent and well-formed, but contain slight inaccuracies that are not easy to discern at first glance. Pay close attention to the text. Read it carefully as you would when proofreading.

The sections below describe each of the dimensions in detail. You can also \textbf{flag} ineligible tasks; the flagging criteria are described in  
\hyperref[sec:instructions_t2t_flag]{this section}.

\section*{\textcolor{Instructions}{1. Interpretability}}\label{sec:instructions_t2t_interpretability}

In this step you will evaluate whether the system caption is \textbf{interpretable} by you. 

You will be shown a system-generated \textbf{caption} of a table. Note that the table from which the caption is derived is hidden at this step, because we need to evaluate whether the caption is \textit{interpretable on its own}. Carefully read the caption and answer the following question:

\begin{itemize}
\item[\textit{(Q1)}] \textit{Is all of the information relayed by the system caption \textbf{interpretable} to you?}
\end{itemize}

This is asking whether you can understand the caption. If there is some part of the caption that is unclear or hard to interpret, select “No”. If prompted by the interface, enter a succinctly detailed justification of your rating.

\paragraph{Definition:} An \textbf{uninterpretable} caption has diminished intelligibility due to:

\begin{itemize}
    \item Vague or ambiguous meaning, e.g., unclear noun references or insufficient context.
    \item Malformed phrases and sentences that are difficult to understand.
\end{itemize}

If the system caption is interpretable by you, you will proceed to the next category.

In the section below, we show in more detail the kind of reasoning that should be used for establishing interpretability of captions. More examples of interpretability ratings and justifications are in \hyperref[sec:instructions_t2t_interpretabilty_examples]{this section} of the appendix.

\subsubsection*{\textcolor{Instructions}{Interpreting the information provided in the system caption}}

Consider the following example:

\medskip
\small
    \begin{tabular}{lp{11.4cm}}
        \textbf{Caption}\\ \multicolumn{2}{p{11.4cm}}{\textcolor{teal}{
         Mikhnevich and Avdeyeva both finished with 19.66 meters in the 2009 World Championships in Athletics – Women's shot put.}	
        } \\
    \end{tabular}

\medskip
\normalsize
In the above example, the meaning of the caption is clear even without seeing the source table. It is clear what the caption is reporting on and it stands on its own; that is, this caption is \textbf{interpretable}. It should be marked as “\textbf{Yes, I understand it.}” But consider another example:

\medskip
\small
    \begin{tabular}{lp{11.4cm}}
        \textbf{Caption}\\ \multicolumn{2}{p{11.4cm}}{\textcolor{teal}{
         Mikhnevich and Avdeyeva finished with 19.66 metres.}	
        } \\
    \end{tabular}

\medskip
\normalsize
In this case the meaning of the caption obviously depends on some context (“\textit{what did they finish?}”), but that context is missing in the caption. Without additional information that clarifies that this is a result of a sports competition, the caption is \textbf{difficult to interpret} and it does not stand on its own. It should be marked as “\textbf{No, the caption is unclear.}”

Captions that require clarifications of this kind are one case where interpretability can be diminished. Other examples include nouns without a (clear) reference and malformed phrases and sentences:

\medskip
\small
    \begin{tabular}{lp{11.4cm}}
        \textbf{Caption 1}\\ \multicolumn{2}{p{11.4cm}}{\textcolor{teal}{
        There are 87,814 Albanians, 624 Serbs and 361 Roma.}
        } \\
        \\
        \textbf{Caption 2}\\ \multicolumn{2}{p{11.4cm}}{\textcolor{teal}{
        J. Thomas Heflin (D) was \hl{a member} until November 1, 1920 and William B. Bowling (D) succeeded him from December 14, 1920.}	
        } \\
        \\
        \textbf{Caption 3}\\ \multicolumn{2}{p{11.4cm}}{\textcolor{teal}{
        George A. Gillett was \hl{a New Zealand dual-code international rugby}.}	
        } \\
        \\
    \end{tabular}

\medskip
\normalsize
In caption 1, the numbers are missing the context of what they are being reported on (“\textit{what location or event do these numbers represent?}”). In caption 2, the phrase “a member” lacks a specifying reference (“\textit{what was Thomas Heflin a member of?}”). In caption 3, the sentence is difficult to understand as it appears to be missing a noun (“\textit{was George A. Gilet a rugby \textbf{player}?}”).

In general, use your best judgment to determine the information provided by the caption. If you are unsure what the intended meaning of the caption is, err on the side of marking it with “\textbf{No, the caption is unclear.}”

\section*{\textcolor{Instructions}{2. Attribution}}

In this step, you will evaluate how well a system-generated caption is \textbf{attributable} to the source table. Note that the source table is a new element that will appear in the task only when you reach this question.
 
\medskip
\footnotesize{\color{darkgray}\textbf{Note}: We refer to “\textbf{attributable to source table}” interchangeably as “\textbf{attribution}” and “\textbf{supported by the source table}”. By which we mean, \textbf{all} of the information in the system caption can be verified from the source table.}

\medskip
\normalsize
You will be shown a system-generated \textbf{caption}. You will also be shown a table and its associated descriptions: title, section title and table section text. These elements provide additional context for understanding the information in the table. Finally, some cells in the table will be highlighted as helpful hints for which parts of the table are the focus of the caption. The table, descriptions and highlighted cells (\textbf{source table}) were used by the system to create the caption. You will use all of these elements to answer the following question:

\begin{itemize}
\item[\textit{(Q2)}] \textit{Is all of the information provided by the system caption \textbf{fully supported }by the source table?}
\end{itemize}

This is asking whether all of the information in the system caption can be attributed to the information present in the source table. If prompted by the interface, enter a succinctly detailed justification of your rating. 

\paragraph{{Definition:}} a \textbf{fully supported} (or \textbf{attributable}) system-generated caption contains an accurate representation of information in the source table. No information in the caption is unattested when compared against the source table and its associated descriptions (title, section title and table section text).

In the section below, we show in more detail the kind of reasoning that should be used for establishing attribution of captions. More examples of attribution ratings and justifications are in \hyperref[sec:instructions_t2t_attribution_examples]{this section} of the appendix. 

\subsubsection*{\textcolor{Instructions}{Assessing the accuracy of the information in the caption against the source table}}

Again, you should use your best judgment in determining whether all of the information provided by the system caption is “an accurate representation of information in the source table”. We give the following guidance:

\begin{itemize}
    \item In determining this question, ask yourself whether it is accurate to say “the provided table says…” or “according to the table…” with the system caption following this phrase.
    \item Be sure to check \textbf{all} of the information in the caption. If only some of the information is supported in the table, but other parts of the information are missing from the table or not an accurate representation, then please mark “\textbf{No, not fully attributable.}”
    \item The concept of “accurate representation” should be close to a journalist’s conception of this phrase. For example take this excerpt from \href{https://www.npr.org/about-npr/688139552/accuracy}{this page} on Accuracy in the NPR Ethics Handbook: “When quoting or paraphrasing anyone\dots consider whether the source would agree with the interpretation\dots” In other words, if you had written the source document, consider whether you would view the caption as an accurate representation of information in that source document.
\end{itemize}

\subsubsection*{\textcolor{Instructions}{Some Final Important Notes}}

\paragraph{Source quality.} When making your judgments in this template,\textbf{ do not take into account whether the underlying source table is correct or trustworthy}. This is clearly important, but will be evaluated in a separate task. The “attribution” category is used only to judge whether the information provided by the system caption is an accurate representation of the underlying source table.

\paragraph{Highlighted cells.} Some of the cells in the table are highlighted. The highlighted cells are intended to be the focus of the caption, and can be used as a helpful hint of where to look in the table for information in the caption — though some captions may also refer to information from elsewhere in the table. \textbf{If the caption does not capture the information in the highlighted cells, but otherwise accurately represents the information elsewhere in the table and its description, please still mark it “Yes, fully attributable.”}

Examples of attribution ratings and justifications are in \hyperref[sec:instructions_t2t_attribution_examples]{this section}.

\section*{\textcolor{Instructions}{Scoring and Examples}}

\subsection*{\textcolor{Instructions}{Interpretability}}\label{sec:instructions_t2t_interpretabilty_examples}

\paragraph{No, the caption is unclear:} The caption is unclear and/or difficult to understand. 

\medskip
\small
    \begin{tabular}{lp{11.4cm}}
        \textbf{Caption}\\ \multicolumn{2}{p{11.4cm}}{\textcolor{teal}{
        Bradman scored 299.}	
        \smallskip
        } \\
    \end{tabular}

    \smallskip
    \begin{tabular}{lp{11.4cm}}
        \textbf{Justification}\\ \multicolumn{2}{p{11.4cm}}{
        \emph{The caption lacks sufficient context (“what did Bradman score in?”).}
        } \\
    \end{tabular}

    \medskip
    \begin{tabular}{lp{11.4cm}}
        \textbf{Caption}\\ \multicolumn{2}{p{11.4cm}}{\textcolor{teal}{
        Amlogic Quad-core Cortex-A53, Mali-450 MP5 OpenGL ES 2.0, and H.264.}	
        \smallskip
        } \\
    \end{tabular}

    \smallskip
    \begin{tabular}{lp{11.4cm}}
        \textbf{Justification}\\ \multicolumn{2}{p{11.4cm}}{
        \emph{The caption is malformed and difficult to interpret: the sentence appears to be missing a verb.}
        } \\
    \end{tabular}

    \medskip
    \begin{tabular}{lp{11.4cm}}
        \textbf{Caption}\\ \multicolumn{2}{p{11.4cm}}{\textcolor{teal}{
        The number one album of the year was Watch the Throne by Jay-Z and Kanye West, which sold 436,000 copies, and Lupe Fiasco's Lasers, which sold 204,000 copies.}
        \smallskip
        } \\
    \end{tabular}

    \smallskip
    \begin{tabular}{lp{11.4cm}}
        \textbf{Justification}\\  \multicolumn{2}{p{11.4cm}}{
        \emph{The caption is malformed because it has two albums listed as “number one album of the year”. Additionally, “the year” lacks a reference (“what year was it?”).}
        } \\
        \\
    \end{tabular}

\medskip    
\normalsize    
\hrule
\subsection*{\textcolor{Instructions}{Attribution}}\label{sec:instructions_t2t_attribution_examples}

\paragraph{Yes, fully attributable:} All the information in the caption is supported by the table and its description.

\medskip
\small
    \begin{tabular}{lp{11.4cm}}
        \textbf{Caption}\\ \multicolumn{2}{p{11.4cm}}{\textcolor{teal}{
        The first-week sales of the album The Watch the Throne by Jay-Z and Kanye West sold 436,000 copies, while Lupe Fiasco's Lasers sold 204,000 copies in the first week.}	
        \smallskip
        } \\
    \end{tabular}
    
    \begin{tabular}{lp{11.4cm}}
        \textbf{Source table}
        \smallskip
    \end{tabular}
    
    \includegraphics[width=.9\textwidth]{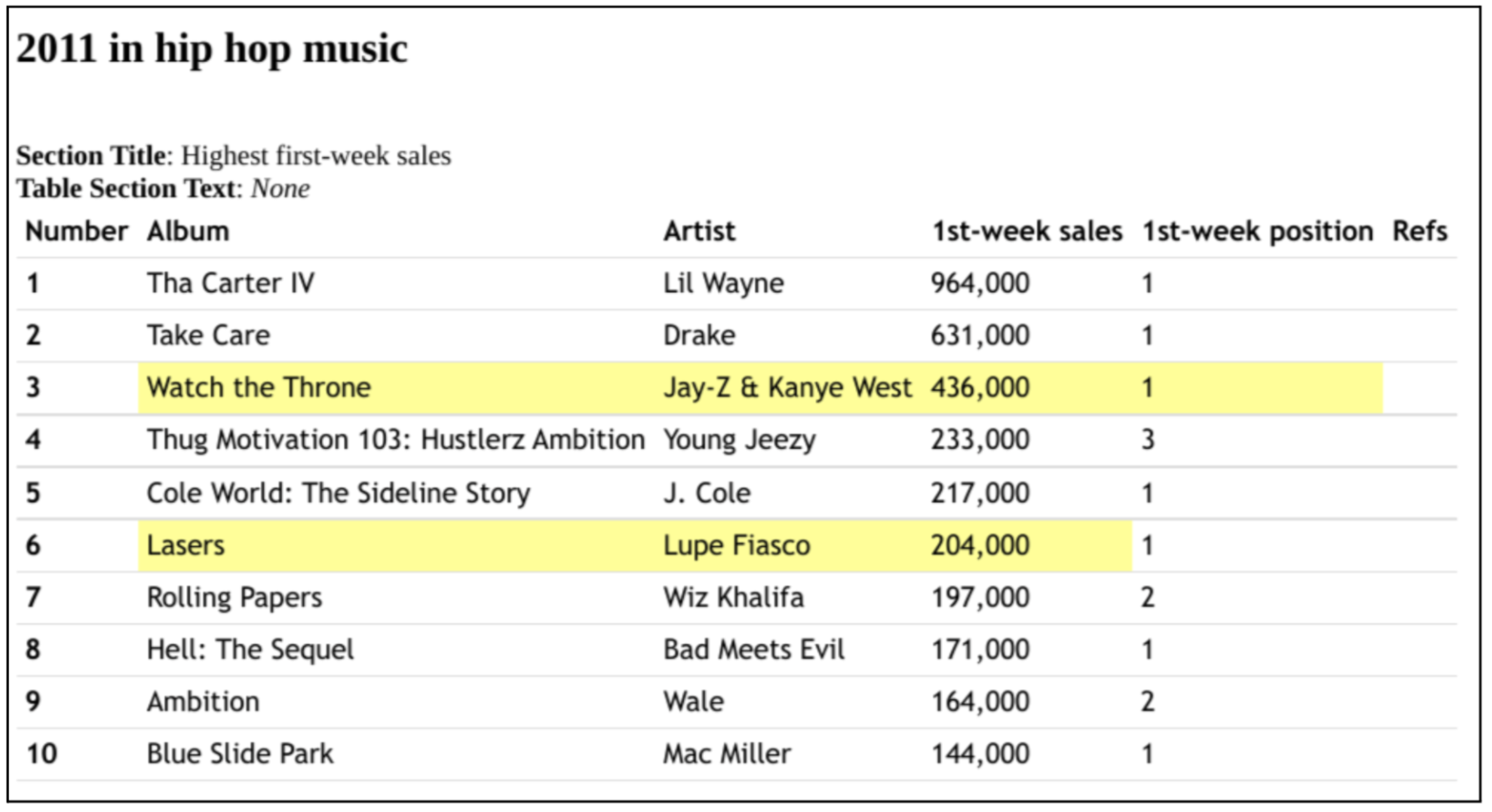}
    \smallskip

    \smallskip
    \begin{tabular}{lp{11.4cm}}
        \textbf{Justification}\\ \multicolumn{2}{p{11.4cm}}{
        \emph{The information in the caption accurately represents the source table.}
        } \\
    \end{tabular}

\medskip
\small
    \begin{tabular}{lp{11.4cm}}
        \textbf{Caption}\\ \multicolumn{2}{p{11.4cm}}{\textcolor{teal}{
        Albanians are the largest ethnic group in Gjilan, followed by Serbs with 624 and Roma with 361 persons.}	
        \smallskip
        } \\
    \end{tabular}
    
    \begin{tabular}{lp{11.4cm}}
        \textbf{Source table}
        \smallskip
    \end{tabular}
    
    \includegraphics[width=.9\textwidth]{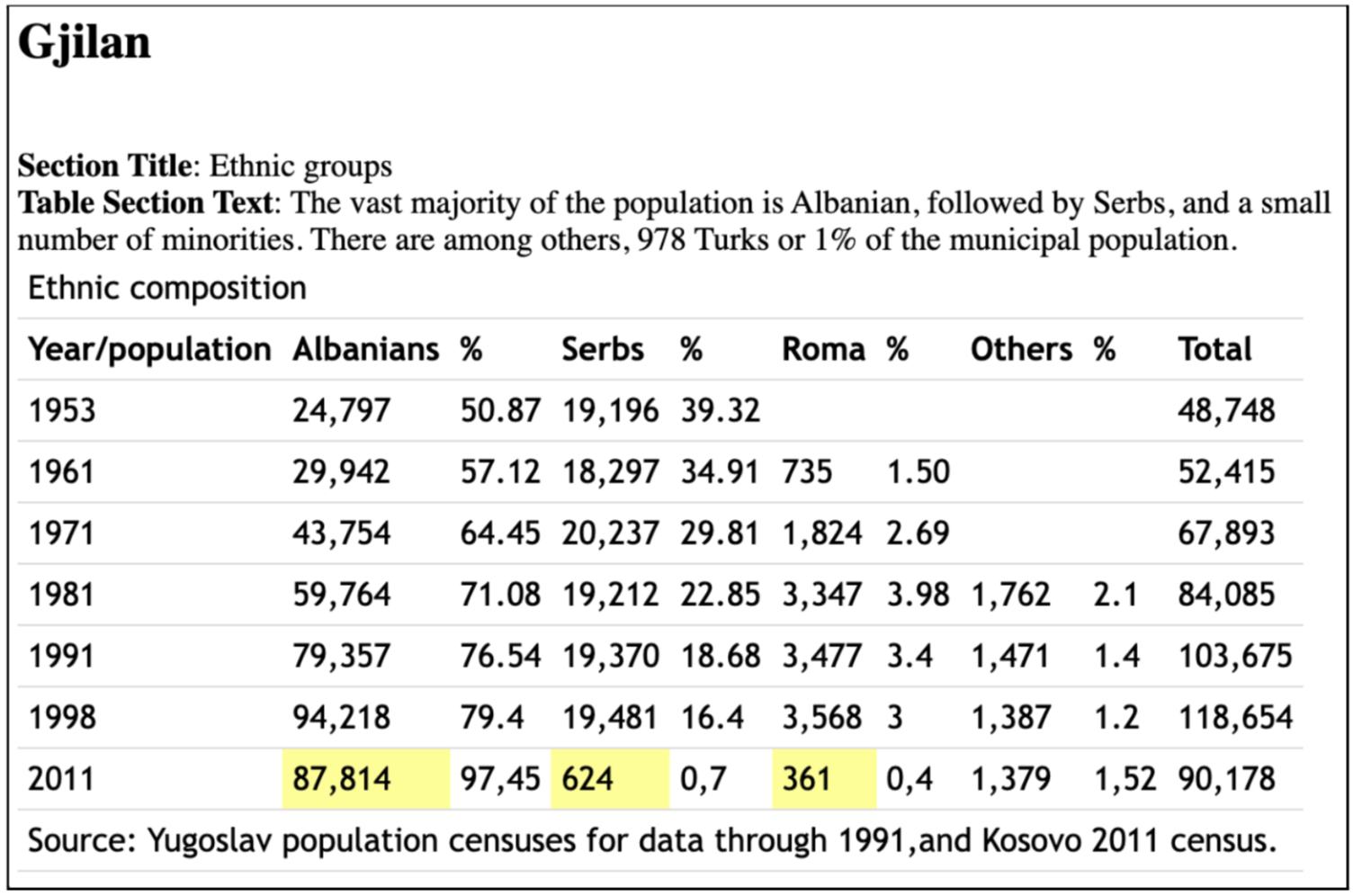}
    \smallskip

    \smallskip
    \begin{tabular}{lp{11.4cm}}
        \textbf{Justification}\\ \multicolumn{2}{p{11.4cm}}{
        \emph{The caption accurately reflects information from the source table. The fact that the Albanian ethnic group is the largest can be easily reasoned from the information in the table.}
        } \\
    \end{tabular}
    
\medskip
\normalsize
\paragraph{No, not fully attributable:} The caption cannot be fully attributed to the source table and its description (including captions that are only PARTIALLY or NOT AT ALL supported).

\medskip
\small
    \begin{tabular}{lp{11.4cm}}
        \textbf{Caption}\\ \multicolumn{2}{p{11.4cm}}{\textcolor{teal}{
        The first-week sales of the album Watch the Throne by Jay-Z \& Kanye West and Lupe Fiasco were 204,000 and the first-week sales of the album Lasers by Lupe Fiasco were 436,000.}	
        \smallskip
        } \\
    \end{tabular}
    
    \begin{tabular}{lp{11.4cm}}
        \textbf{Source table}
        \smallskip
    \end{tabular}
    
    \includegraphics[width=.9\textwidth]{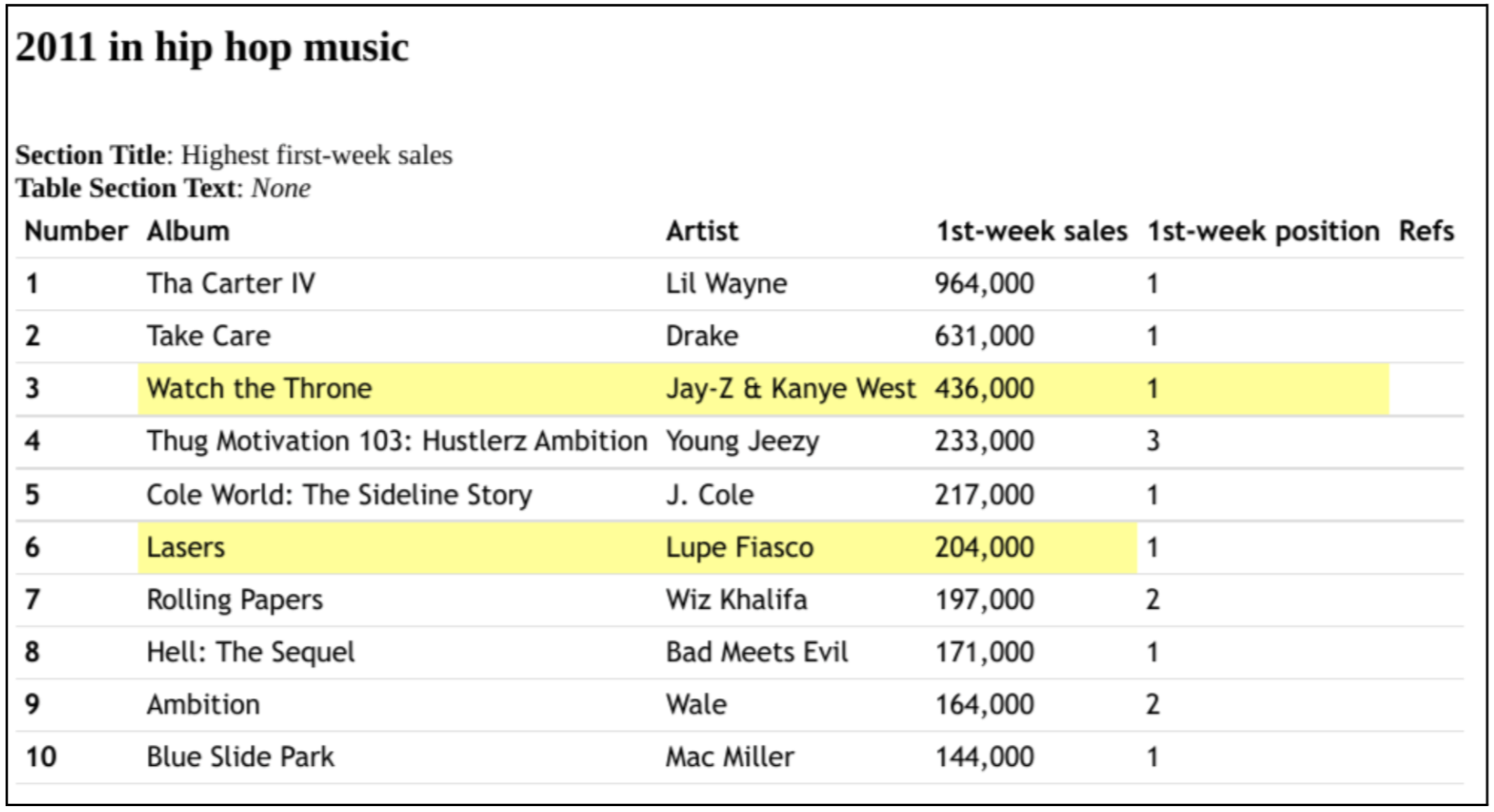}
    \smallskip

    \smallskip
    \begin{tabular}{lp{11.4cm}}
        \textbf{Justification}\\ \multicolumn{2}{p{11.4cm}}{
        \emph{The artists on the album Watch the Throne are only Jay-Z and Kanye West, excluding Lupe Fiasco. The first week’s sales of Watch the Throne were 436,000, not 204,000. The first week’s sales of Lasers were 204,000, not 436,000.}
        } \\
    \end{tabular}

\medskip
\small
    \begin{tabular}{lp{11.4cm}}
        \textbf{Caption}\\ \multicolumn{2}{p{11.4cm}}{\textcolor{teal}{
        Juan Mora Fernández was the Head of State of Costa Rica, winning 11 seats in the San José, 8 in Cartago, 8 in Heredia, 3 in Escazú , 2 in Ujarrás, 1 in Térraba and 1 in Bagaces.}	
        \smallskip
        } \\
    \end{tabular}
    
    \begin{tabular}{lp{11.4cm}}
        \textbf{Source table}
        \smallskip
    \end{tabular}
    
    \includegraphics[width=.9\textwidth]{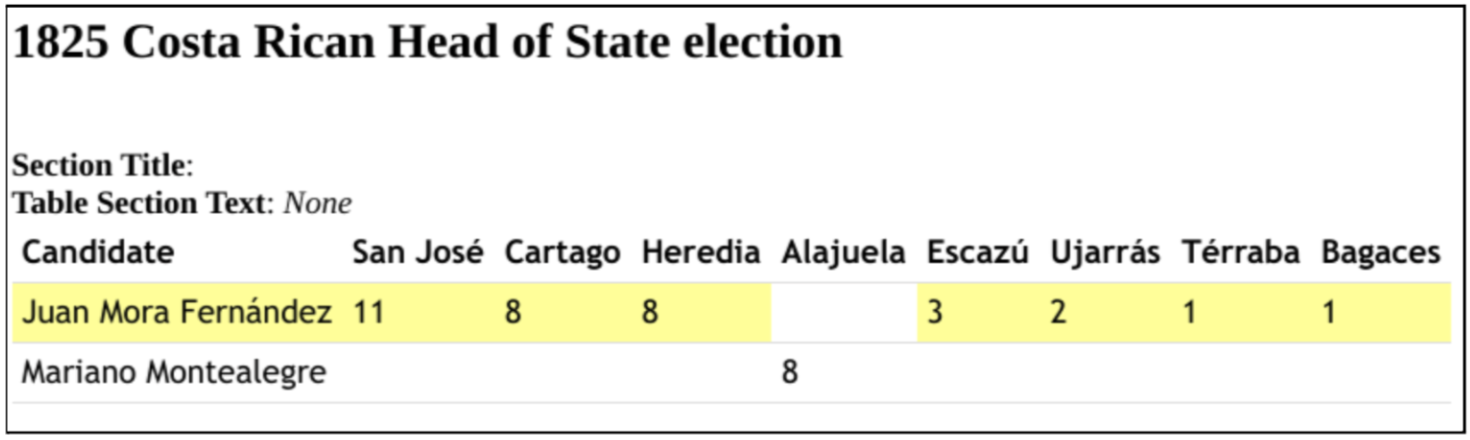}
    \smallskip

    \smallskip
    \begin{tabular}{lp{11.4cm}}
        \textbf{Justification}\\ \multicolumn{2}{p{11.4cm}}{
        \emph{The table does not have a clear identification of the reported numbers, while the caption identifies them as “seats”, which is not attributable anywhere in the table or its descriptions. }
        } \\
    \end{tabular}

\medskip    
\normalsize    
\hrule
\subsection*{\textcolor{Instructions}{Flag}}\label{sec:instructions_t2t_flag}

There is a flag button in the bottom left corner of the task interface. Once flagged, you can proceed onto the next task. Use it to report tasks that are ineligible for reasons such as:

\paragraph{1. Missing components in the task.} Some tasks may have missing summaries or news articles. They should be flagged.

Note that table title, section title, or table section text could be empty or designated with “None”. These are acceptable and should not be flagged. See an example of an acceptable table description below:

\bigskip
\includegraphics[scale=0.15]{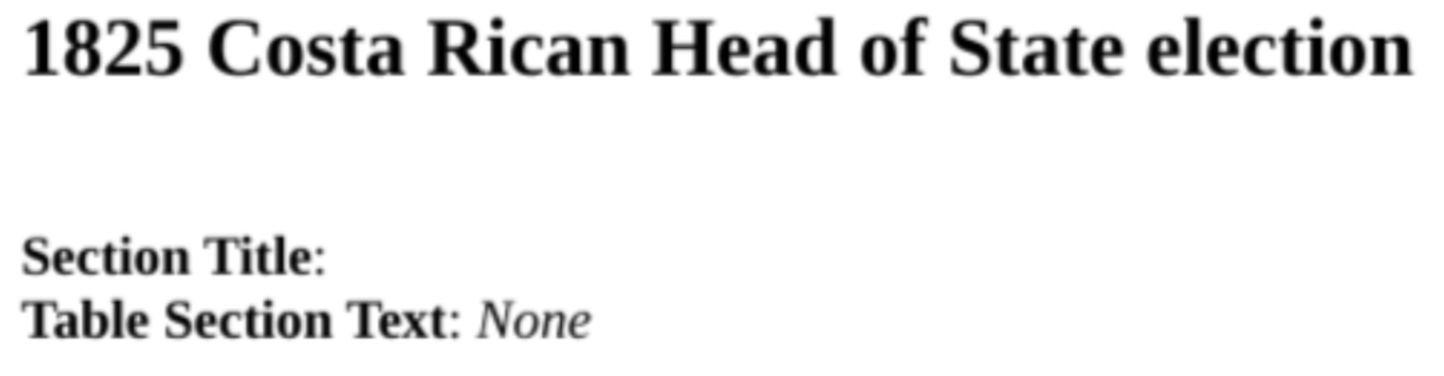}
\bigskip

\paragraph{2. Malformed text.} Some text may be severely malformed with unintelligible artifacts (e.g. html code, unformatted tables, etc.). If any component of the task contains malformed text, the task should be flagged.

\medskip
\small
    \begin{tabular}{lp{11.4cm}}
        \textbf{Caption}\\ \multicolumn{2}{p{11.4cm}}{\textcolor{teal}{
        The lowest temperature recorded in Porto Alegre was ?? 0.3$^{\circ}$C (31.5$^{\circ}$F).}
        } \\
        \\
    \end{tabular}

\medskip
\normalsize

\paragraph{3. Source table is difficult to understand because it requires expert-level knowledge.} Some tables may include scientific formulas, obscure terminology, etc. If you can still understand enough of the table to rate its attributability, please do so. But if properly evaluating the response requires expertise in a particular area, please flag it.

\medskip
\small
    \begin{tabular}{lp{11.4cm}}
        \textbf{Caption}\\ \multicolumn{2}{p{11.4cm}}{\textcolor{teal}{
        The longest-lived isotope is 18mF with a half-life of 162 ns.}	
        \smallskip
        } \\
    \end{tabular}
    
    \begin{tabular}{lp{11.4cm}}
        \textbf{Source table}
        \smallskip
    \end{tabular}
    
    \includegraphics[width=.9\textwidth]{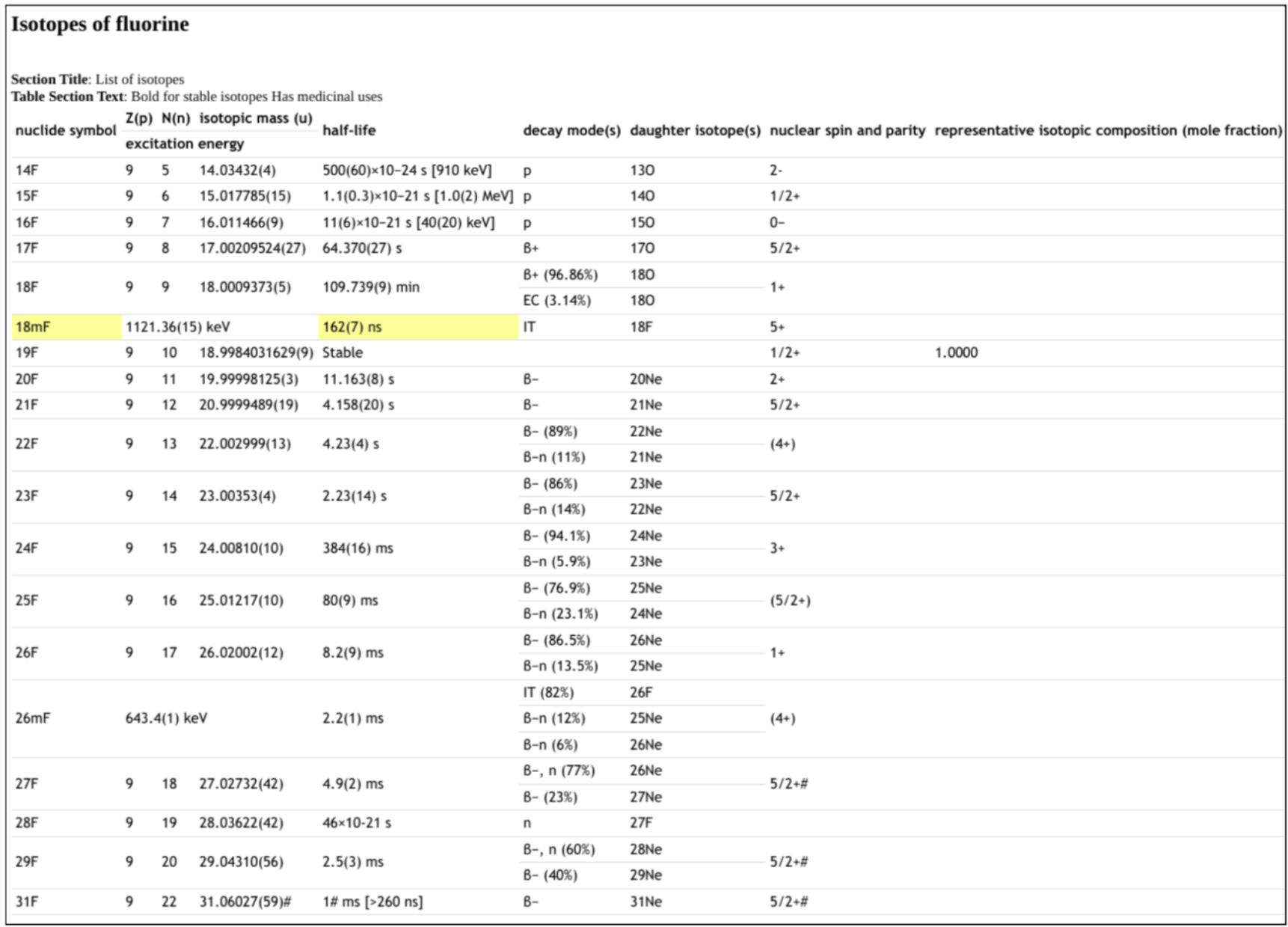}
    \smallskip

    \smallskip
    \begin{tabular}{lp{11.4cm}}
        \textbf{Justification}\\ \multicolumn{2}{p{11.4cm}}{
        \emph{In order to be able to evaluate whether this table supports the caption, it requires a deeper understanding of scientific equations and terminology contained in the table.  Because this example requires scientific expertise to evaluate it properly, it should be flagged.}
        } \\
    \end{tabular} \\
\normalsize  \\
\appendixsection{Annotator Interface for Conversational QA Tasks}
\begin{figure}[H]
\centering
\includegraphics[width=0.8\linewidth]{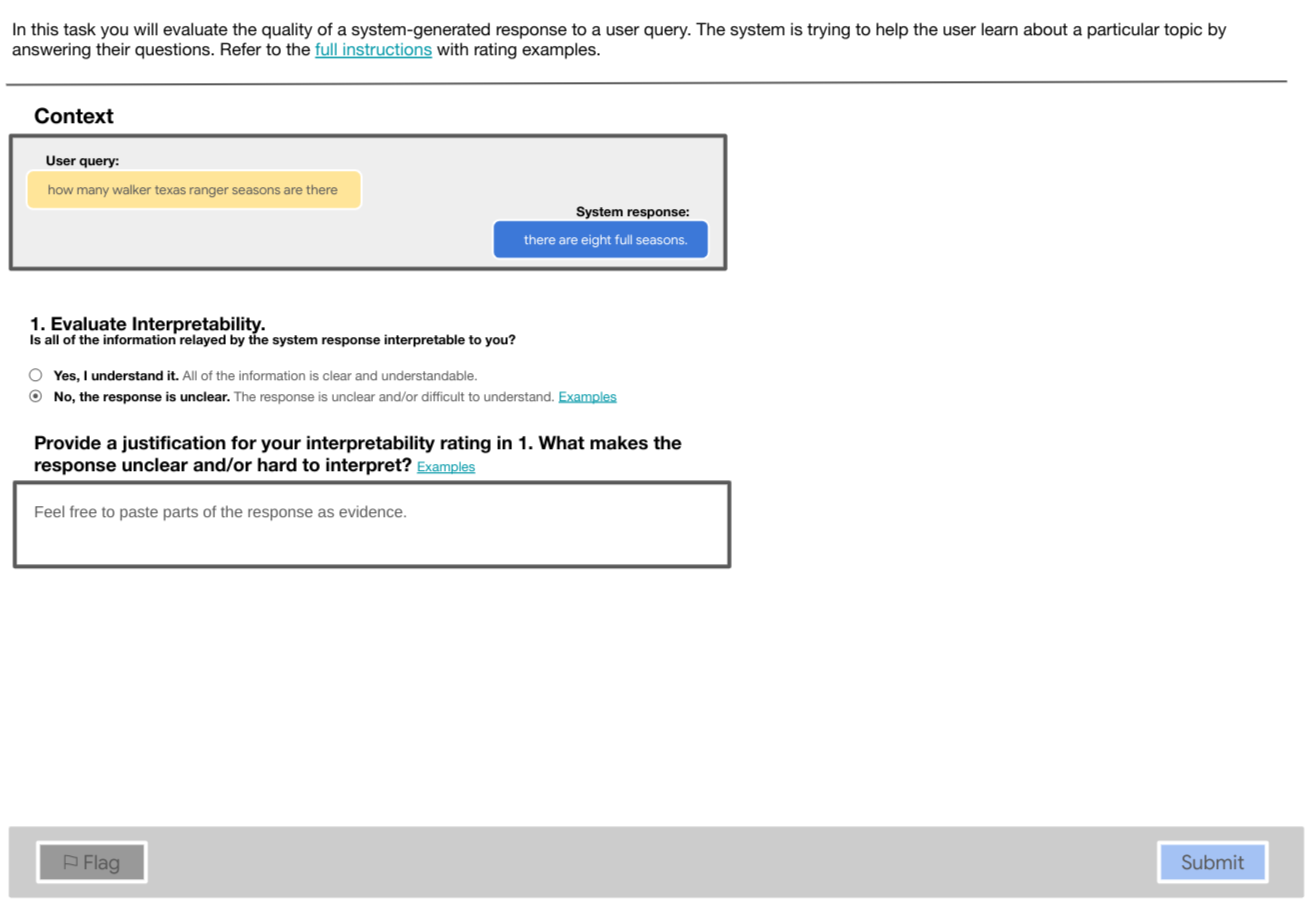} 
\caption{Interpretability stage. The source document is hidden. If the task is rated as not interpretable, the attribution stage is skipped and the annotator proceeds to the next task in the queue. During training and pilot, the justification element is shown only if the task is rated as not interpretable.}
\label{fig:AIS_QA_UI_1}
\end{figure}
\begin{figure}[H]
\centering
\includegraphics[width=0.8\linewidth]{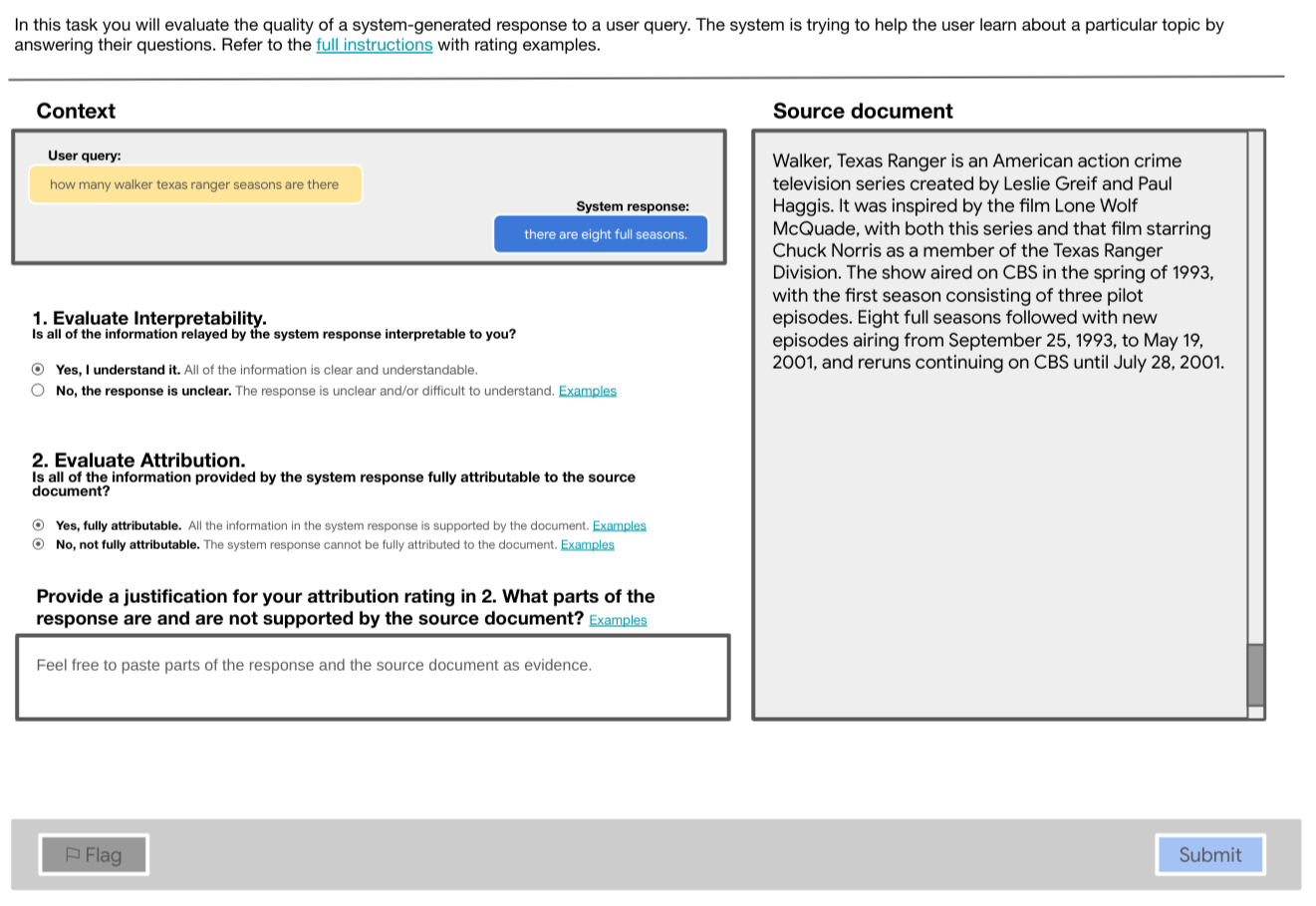}
\caption{Attribution stage. The source document is shown. During training and pilot, the justification element is required for all ratings.}
\label{fig:AIS_QA_UI_2}
\end{figure}

\appendixsection{Annotator Interface for Summarization Tasks}
\begin{figure}[H]
\centering
\includegraphics[width=0.8\linewidth]{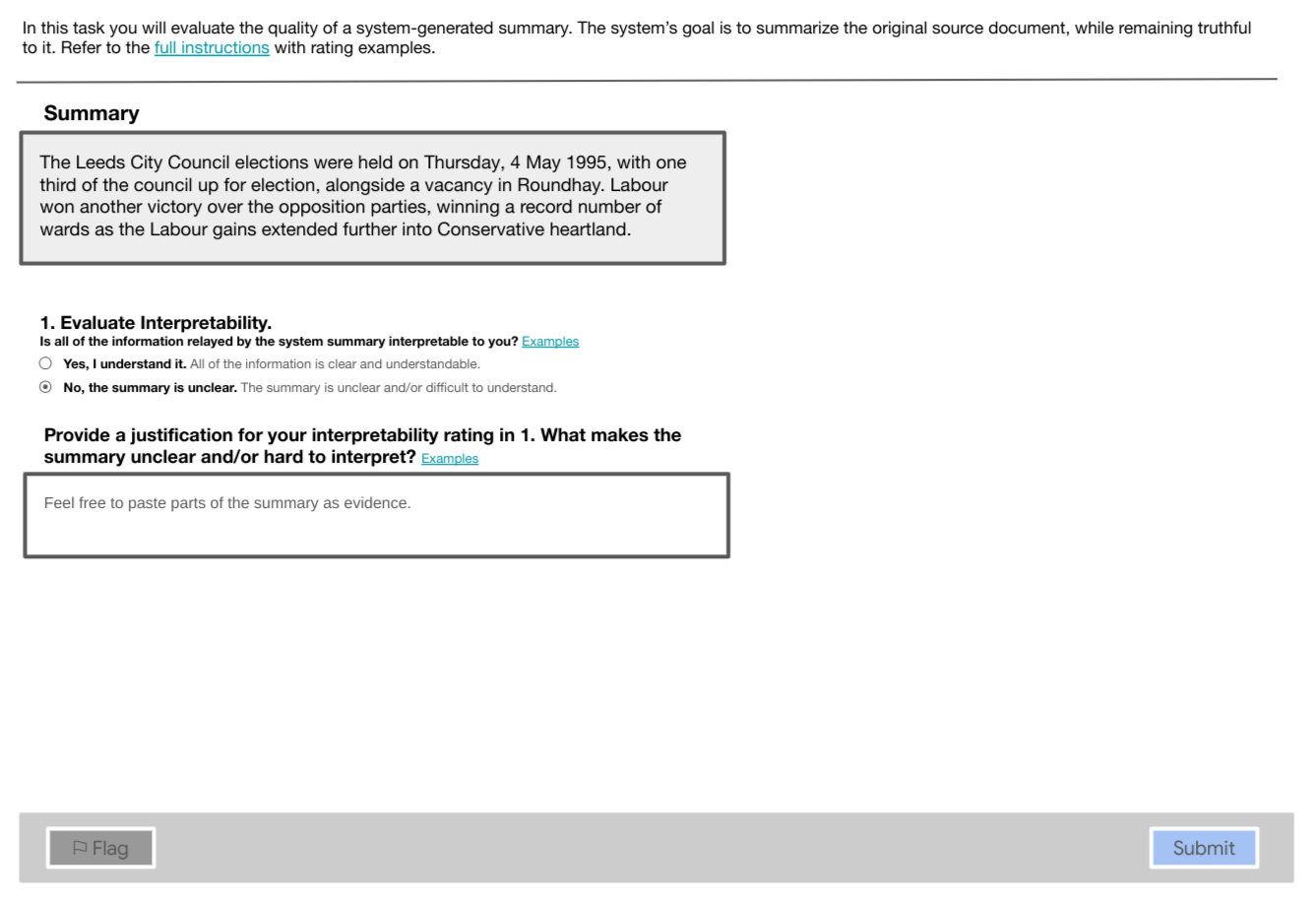} 
\caption{Interpretability stage. The source document is hidden. If the task is rated as not interpretable, the attribution stage is skipped and the annotator proceeds to the next task in the queue. During training and pilot, the justification element is shown only if the task is rated as not interpretable.}
\label{fig:AIS_Summ_UI_1}
\end{figure}
\begin{figure}[H]
\centering
\includegraphics[width=0.8\linewidth]{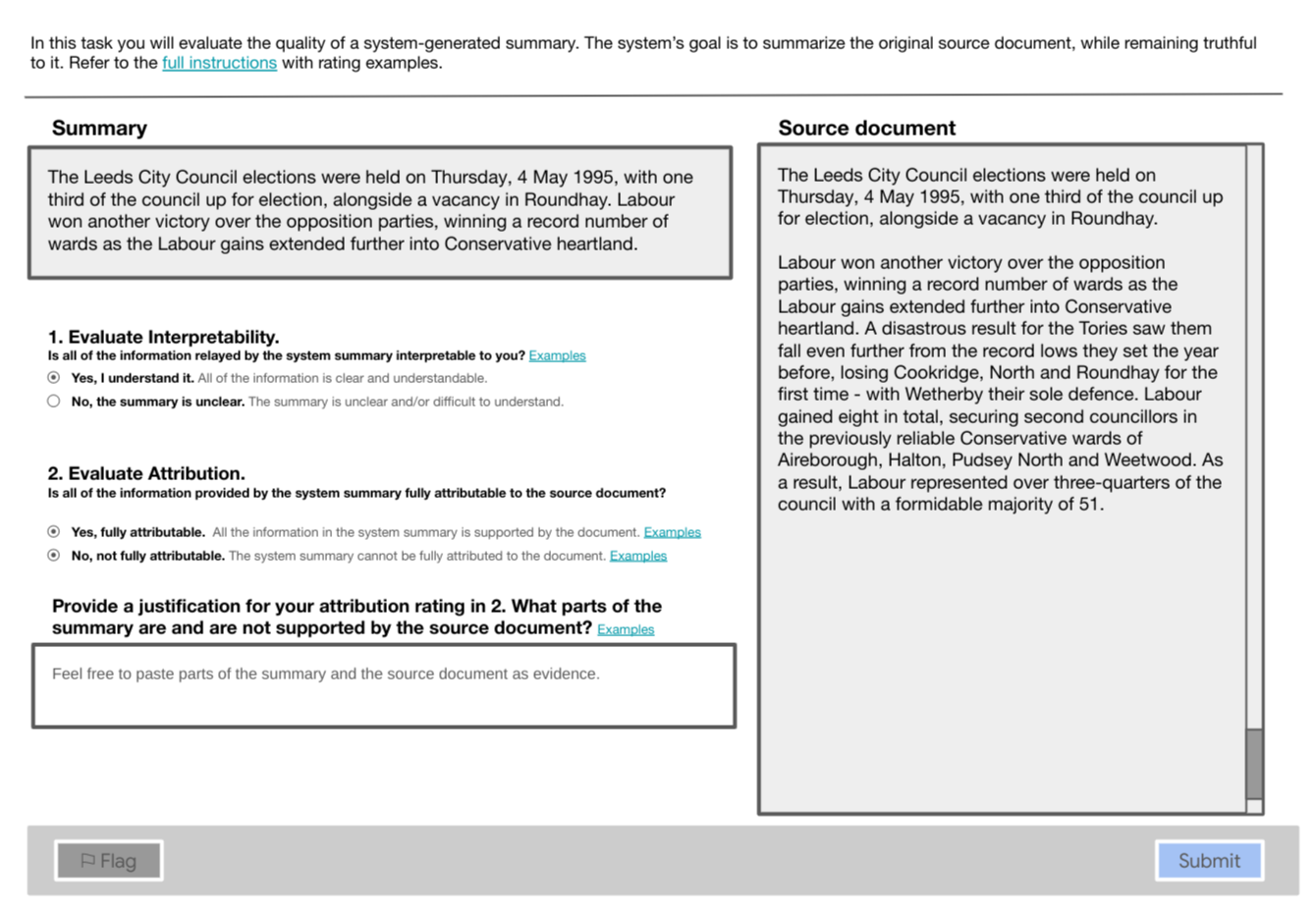}
\caption{Attribution stage. The source document is shown. During training and pilot, the justification element is required for all ratings.}
\label{fig:AIS_Summ_UI_2}
\end{figure}

\appendixsection{Annotator Interface for Table-to-Text Tasks}
\begin{figure}[H]
\centering
\includegraphics[width=0.8\linewidth]{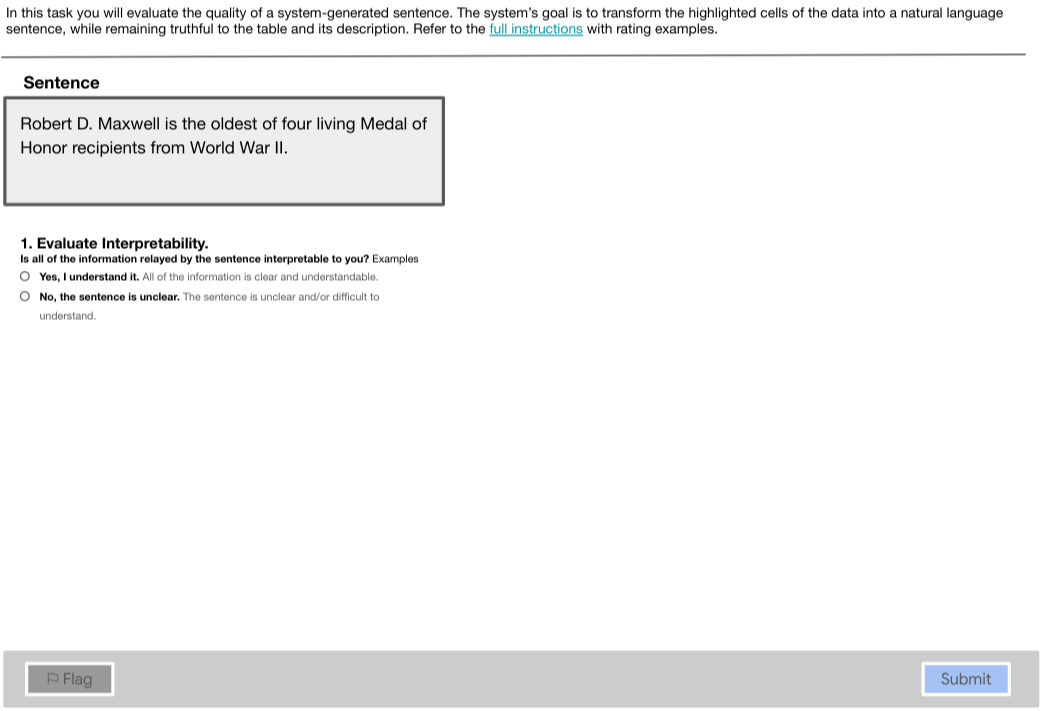} 
\caption{Interpretability stage. The source table and its description are hidden. If the task is rated as not interpretable, the attribution stage is skipped and the annotator proceeds to the next task in the queue. During training and pilot, the justification element is shown only if the task is rated as not interpretable.}
\label{fig:AIS_T2T_UI_1}
\end{figure}
\begin{figure}[H]
\centering
\includegraphics[width=0.8\linewidth]{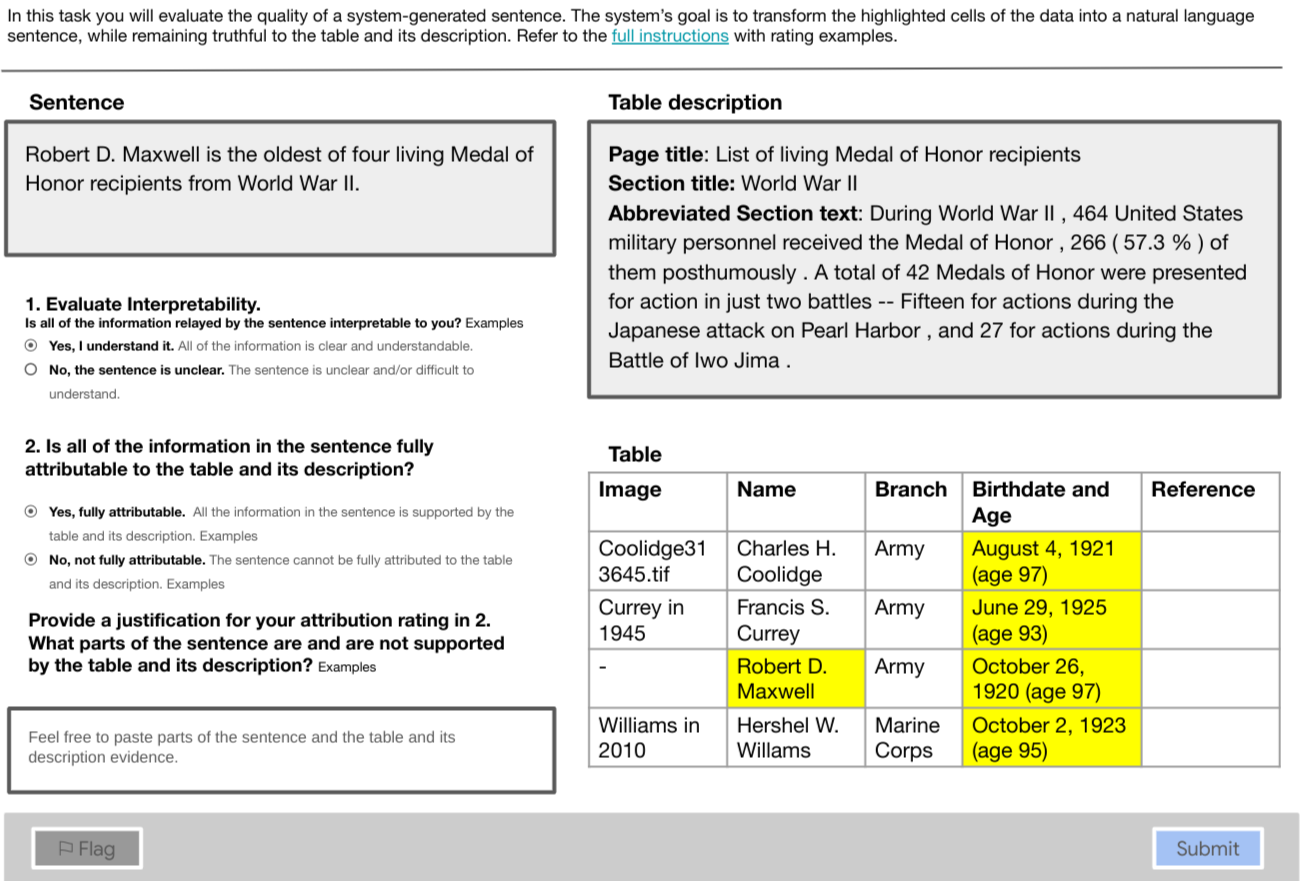}
\caption{Attribution stage. The source table its description are shown. The rendering preserves highlighted cells from the ToTTo data. During training and pilot, the justification element is shown for all ratings in the second stage.}
\label{fig:AIS_T2T_UI_2}
\end{figure}

\starttwocolumn
\acknowledgments
The authors would like to thank Roee Aharoni, Sebastian Gehrmann, Mirella Lapata, Hongrae Lee, Shashi Narayan, Ankur Parikh, Fernando Pereira, and Idan Szpektor for their detailed feedback on the manuscript and throughout the project. We would also like to thank Ashwin Kakarla and his team for making the human evaluation study possible and Alejandra Molina and Kristen Olson for their guidance on the task interface design. We are thankful to the larger Google Research community for the many discussions during the course of this project.
\bibliography{ais}

\begin{thebibliography}{41}
\expandafter\ifx\csname natexlab\endcsname\relax\def\natexlab#1{#1}\fi

\bibitem[{Adiwardana et~al.(2020)Adiwardana, Luong, So, Hall, Fiedel,
  Thoppilan, Yang, Kulshreshtha, Nemade, Lu et~al.}]{adiwardana2020towards}
Adiwardana, Daniel, Minh-Thang Luong, David~R So, Jamie Hall, Noah Fiedel,
  Romal Thoppilan, Zi~Yang, Apoorv Kulshreshtha, Gaurav Nemade, Yifeng Lu,
  et~al. 2020.
\newblock Towards a human-like open-domain chatbot.
\newblock \emph{arXiv preprint arXiv:2001.09977}.

\bibitem[{Anantha et~al.(2021)Anantha, Vakulenko, Tu, Longpre, Pulman, and
  Chappidi}]{qrecc}
Anantha, Raviteja, Svitlana Vakulenko, Zhucheng Tu, Shayne Longpre, Stephen
  Pulman, and Srinivas Chappidi. 2021.
\newblock Open-domain question answering goes conversational via question
  rewriting.
\newblock In \emph{Proceedings of NAACL}.

\bibitem[{Belz, Mille, and Howcroft(2020)}]{belz2020disentangling}
Belz, Anja, Simon Mille, and David~M Howcroft. 2020.
\newblock Disentangling the properties of human evaluation methods: A
  classification system to support comparability, meta-evaluation and
  reproducibility testing.
\newblock In \emph{Proceedings of the 13th International Conference on Natural
  Language Generation}, pages 183--194.

\bibitem[{Carston(1988)}]{carston1988explicature}
Carston, Robyn. 1988.
\newblock Implicature, explicature, and truth-theoretic semantics.
\newblock In Ruth Kempson, editor, \emph{Mental Representations: The Interface
  Between Language and Reality}. Cambridge University Press, pages 155--181.

\bibitem[{Choi et~al.(2018)Choi, He, Iyyer, Yatskar, Yih, Choi, Liang, and
  Zettlemoyer}]{quacppr}
Choi, Eunsol, He~He, Mohit Iyyer, Mark Yatskar, Wen-tau Yih, Yejin Choi, Percy
  Liang, and Luke Zettlemoyer. 2018.
\newblock {Q}u{AC}: Question answering in context.
\newblock In \emph{Proceedings of EMNLP}.

\bibitem[{Choi et~al.(2021)Choi, Palomaki, Lamm, Kwiatkowski, Das, and
  Collins}]{choi2021decontext}
Choi, Eunsol, Jennimaria Palomaki, Matthew Lamm, Tom Kwiatkowski, Dipanjan Das,
  and Michael Collins. 2021.
\newblock {Decontextualization: Making Sentences Stand-Alone}.
\newblock \emph{Transactions of the Association for Computational Linguistics},
  9:447--461.

\bibitem[{Dalton et~al.(2020)Dalton, Xiong, Kumar, and Callan}]{castppr}
Dalton, Jeffrey, Chenyan Xiong, Vaibhav Kumar, and Jamie Callan. 2020.
\newblock Cast-19: A dataset for conversational information seeking.
\newblock \emph{Proceedings of SIGIR}.

\bibitem[{Dinan et~al.(2019)Dinan, Roller, Shuster, Fan, Auli, and
  Weston}]{dinan2019wizard}
Dinan, Emily, Stephen Roller, Kurt Shuster, Angela Fan, Michael Auli, and Jason
  Weston. 2019.
\newblock Wizard of wikipedia: Knowledge-powered conversational agents.
\newblock In \emph{Proceedings of {ICLR}}.

\bibitem[{Durmus, He, and Diab(2020)}]{durmus-etal-2020-feqa}
Durmus, Esin, He~He, and Mona Diab. 2020.
\newblock {FEQA}: A question answering evaluation framework for faithfulness
  assessment in abstractive summarization.
\newblock In \emph{Proceedings of ACL}.

\bibitem[{Dziri et~al.(2021)Dziri, Rashkin, Linzen, and
  Reitter}]{Dziri2021EvaluatingGI}
Dziri, Nouha, Hannah Rashkin, Tal Linzen, and David Reitter. 2021.
\newblock Evaluating groundedness in dialogue systems: The {BEGIN} benchmark.
\newblock \emph{ArXiv}, abs/2105.00071.

\bibitem[{Gehrmann et~al.(2021)Gehrmann, Adewumi, Aggarwal, Ammanamanchi,
  Aremu, Bosselut, Chandu, Clinciu, Das, Dhole, Du, Durmus, Du{\v{s}}ek,
  Emezue, Gangal, Garbacea, Hashimoto, Hou, Jernite, Jhamtani, Ji, Jolly, Kale,
  Kumar, Ladhak, Madaan, Maddela, Mahajan, Mahamood, Majumder, Martins,
  McMillan-Major, Mille, van Miltenburg, Nadeem, Narayan, Nikolaev,
  Niyongabo~Rubungo, Osei, Parikh, Perez-Beltrachini, Rao, Raunak, Rodriguez,
  Santhanam, Sedoc, Sellam, Shaikh, Shimorina, Sobrevilla~Cabezudo, Strobelt,
  Subramani, Xu, Yang, Yerukola, and Zhou}]{gehrmann-etal-2021-gem}
Gehrmann, Sebastian, Tosin Adewumi, Karmanya Aggarwal, Pawan~Sasanka
  Ammanamanchi, Anuoluwapo Aremu, Antoine Bosselut, Khyathi~Raghavi Chandu,
  Miruna-Adriana Clinciu, Dipanjan Das, Kaustubh Dhole, Wanyu Du, Esin Durmus,
  Ond{\v{r}}ej Du{\v{s}}ek, Chris~Chinenye Emezue, Varun Gangal, Cristina
  Garbacea, Tatsunori Hashimoto, Yufang Hou, Yacine Jernite, Harsh Jhamtani,
  Yangfeng Ji, Shailza Jolly, Mihir Kale, Dhruv Kumar, Faisal Ladhak, Aman
  Madaan, Mounica Maddela, Khyati Mahajan, Saad Mahamood, Bodhisattwa~Prasad
  Majumder, Pedro~Henrique Martins, Angelina McMillan-Major, Simon Mille, Emiel
  van Miltenburg, Moin Nadeem, Shashi Narayan, Vitaly Nikolaev, Andre
  Niyongabo~Rubungo, Salomey Osei, Ankur Parikh, Laura Perez-Beltrachini,
  Niranjan~Ramesh Rao, Vikas Raunak, Juan~Diego Rodriguez, Sashank Santhanam,
  Jo{\~a}o Sedoc, Thibault Sellam, Samira Shaikh, Anastasia Shimorina,
  Marco~Antonio Sobrevilla~Cabezudo, Hendrik Strobelt, Nishant Subramani, Wei
  Xu, Diyi Yang, Akhila Yerukola, and Jiawei Zhou. 2021.
\newblock The {GEM} benchmark: Natural language generation, its evaluation and
  metrics.
\newblock In \emph{Proceedings of the 1st Workshop on Natural Language
  Generation, Evaluation, and Metrics}.

\bibitem[{Gopnik and Wellman(1992)}]{gopnik1992child}
Gopnik, Alison and Henry~M. Wellman. 1992.
\newblock Why the child's theory of mind really is a theory.
\newblock \emph{Mind \& Language}, 7(1-2):145--171.

\bibitem[{Grice(1975)}]{grice1975logic}
Grice, Herbert~P. 1975.
\newblock Logic and conversation.
\newblock In \emph{Speech acts}. Brill, pages 41--58.

\bibitem[{Gupta et~al.(2021)Gupta, Wu, Liu, and Xiong}]{Gupta2021DialFactAB}
Gupta, Prakhar, Chien-Sheng Wu, Wenhao Liu, and Caiming Xiong. 2021.
\newblock Dialfact: A benchmark for fact-checking in dialogue.
\newblock \emph{ArXiv}, abs/2110.08222.

\bibitem[{Harrington et~al.(1985)Harrington, Morley, {\v{S}}cedrov, and
  Simpson}]{harrington1985harvey}
Harrington, Leo~A, Michael~D Morley, A~{\v{S}}cedrov, and Stephen~G Simpson.
  1985.
\newblock \emph{Harvey Friedman's research on the foundations of mathematics}.
\newblock Elsevier.

\bibitem[{Honovich et~al.(2021)Honovich, Choshen, Aharoni, Neeman, Szpektor,
  and Abend}]{Honovich2021Q2EF}
Honovich, Or, Leshem Choshen, Roee Aharoni, Ella Neeman, Idan Szpektor, and
  Omri Abend. 2021.
\newblock $q^{2}$: {E}valuating factual consistency in knowledge-grounded
  dialogues via question generation and question answering.
\newblock In \emph{Proceedings of EMNLP}.

\bibitem[{Howcroft et~al.(2020)Howcroft, Belz, Clinciu, Gkatzia, Hasan,
  Mahamood, Mille, van Miltenburg, Santhanam, and Rieser}]{howcroft2020twenty}
Howcroft, David~M, Anja Belz, Miruna-Adriana Clinciu, Dimitra Gkatzia, Sadid~A
  Hasan, Saad Mahamood, Simon Mille, Emiel van Miltenburg, Sashank Santhanam,
  and Verena Rieser. 2020.
\newblock Twenty years of confusion in human evaluation: Nlg needs evaluation
  sheets and standardised definitions.
\newblock In \emph{Proceedings of the 13th International Conference on Natural
  Language Generation}, pages 169--182.

\bibitem[{Kwiatkowski et~al.(2019)Kwiatkowski, Palomaki, Redfield, Collins,
  Parikh, Alberti, Epstein, Polosukhin, Devlin, Lee, Toutanova, Jones, Kelcey,
  Chang, Dai, Uszkoreit, Le, and Petrov}]{NQppr}
Kwiatkowski, Tom, Jennimaria Palomaki, Olivia Redfield, Michael Collins, Ankur
  Parikh, Chris Alberti, Danielle Epstein, Illia Polosukhin, Jacob Devlin,
  Kenton Lee, Kristina Toutanova, Llion Jones, Matthew Kelcey, Ming-Wei Chang,
  Andrew~M. Dai, Jakob Uszkoreit, Quoc Le, and Slav Petrov. 2019.
\newblock Natural questions: A benchmark for question answering research.
\newblock \emph{Transactions of the Association for Computational Linguistics},
  7:452--466.

\bibitem[{Ladhak et~al.(2022)Ladhak, Durmus, He, Cardie, and
  McKeown}]{ladhak2022faithful}
Ladhak, Faisal, Esin Durmus, He~He, Claire Cardie, and Kathleen McKeown. 2022.
\newblock Faithful or extractive? on mitigating the
  faithfulness-abstractiveness trade-off in abstractive summarization.
\newblock In \emph{Proceedings of the 60th Annual Meeting of the Association
  for Computational Linguistics (Volume 1: Long Papers)}, pages 1410--1421,
  Association for Computational Linguistics, Dublin, Ireland.

\bibitem[{Maynez et~al.(2020)Maynez, Narayan, Bohnet, and
  McDonald}]{maynez2020faithfulness}
Maynez, Joshua, Shashi Narayan, Bernd Bohnet, and Ryan McDonald. 2020.
\newblock On faithfulness and factuality in abstractive summarization.
\newblock In \emph{Proceedings of ACL}.

\bibitem[{Mehri and Eskenazi(2020)}]{mehri-eskenazi-2020-usr}
Mehri, Shikib and Maxine Eskenazi. 2020.
\newblock {USR}: An unsupervised and reference free evaluation metric for
  dialog generation.
\newblock In \emph{Proceedings of ACL}, Online.

\bibitem[{Nallapati et~al.(2016)Nallapati, Zhou, dos Santos,
  G{\.{u}}l{\c{c}}ehre, and Xiang}]{cnndm}
Nallapati, Ramesh, Bowen Zhou, Cicero dos Santos, {\c{C}}a{\u{g}}lar
  G{\.{u}}l{\c{c}}ehre, and Bing Xiang. 2016.
\newblock Abstractive text summarization using sequence-to-sequence {RNN}s and
  beyond.
\newblock In \emph{Proceedings of {C}o{NLL}}.

\bibitem[{Nan et~al.(2021)Nan, Nogueira~dos Santos, Zhu, Ng, McKeown,
  Nallapati, Zhang, Wang, Arnold, and Xiang}]{nan-etal-2021-improving}
Nan, Feng, Cicero Nogueira~dos Santos, Henghui Zhu, Patrick Ng, Kathleen
  McKeown, Ramesh Nallapati, Dejiao Zhang, Zhiguo Wang, Andrew~O. Arnold, and
  Bing Xiang. 2021.
\newblock Improving factual consistency of abstractive summarization via
  question answering.
\newblock In \emph{Proceedings of ACL}.

\bibitem[{Nie, Chen, and Bansal(2019)}]{DBLP:conf/aaai/NieCB19}
Nie, Yixin, Haonan Chen, and Mohit Bansal. 2019.
\newblock Combining fact extraction and verification with neural semantic
  matching networks.
\newblock In \emph{Proceedings of AAAI}.

\bibitem[{Parikh et~al.(2020)Parikh, Wang, Gehrmann, Faruqui, Dhingra, Yang,
  and Das}]{parikh-etal-2020-totto}
Parikh, Ankur, Xuezhi Wang, Sebastian Gehrmann, Manaal Faruqui, Bhuwan Dhingra,
  Diyi Yang, and Dipanjan Das. 2020.
\newblock {ToTTo}: A controlled table-to-text generation dataset.
\newblock In \emph{Proceedings of EMNLP}.

\bibitem[{Pavlick and Kwiatkowski(2019)}]{pavlick2019inference}
Pavlick, Ellie and Tom Kwiatkowski. 2019.
\newblock {Inherent Disagreements in Human Textual Inferences}.
\newblock \emph{Transactions of the Association for Computational Linguistics},
  7:677--694.

\bibitem[{Raffel et~al.(2020)Raffel, Shazeer, Roberts, Lee, Narang, Matena,
  Zhou, Li, and Liu}]{t5ppr}
Raffel, Colin, Noam Shazeer, Adam Roberts, Katherine Lee, Sharan Narang,
  Michael Matena, Yanqi Zhou, Wei Li, and Peter~J. Liu. 2020.
\newblock Exploring the limits of transfer learning with a unified text-to-text
  transformer.
\newblock \emph{Journal of Machine Learning Research}, 21:140:1--140:67.

\bibitem[{Rashkin et~al.(2021)Rashkin, Reitter, Tomar, and Das}]{rashkinetal}
Rashkin, Hannah, David Reitter, Gaurav~Singh Tomar, and Dipanjan Das. 2021.
\newblock Increasing faithfulness in knowledge-grounded dialogue with
  controllable features.
\newblock In \emph{Proceedings of ACL}.

\bibitem[{Ribeiro, Singh, and Guestrin(2016)}]{ribeiro2016model}
Ribeiro, Marco~Tulio, Sameer Singh, and Carlos Guestrin. 2016.
\newblock Model-agnostic interpretability of machine learning.
\newblock \emph{arXiv preprint arXiv:1606.05386}.

\bibitem[{Santhanam et~al.(2021)Santhanam, Hedayatnia, Gella, Padmakumar, Kim,
  Liu, and Hakkani-Tur}]{santhanam2021rome}
Santhanam, Sashank, Behnam Hedayatnia, Spandana Gella, Aishwarya Padmakumar,
  Seokhwan Kim, Yang Liu, and Dilek Hakkani-Tur. 2021.
\newblock Rome was built in 1776: A case study on factual correctness in
  knowledge-grounded response generation.
\newblock \emph{arXiv preprint arXiv:2110.05456}.

\bibitem[{See, Liu, and Manning(2017)}]{pointernetwork}
See, Abigail, Peter~J. Liu, and Christopher~D. Manning. 2017.
\newblock Get to the point: Summarization with pointer-generator networks.
\newblock In \emph{Proceedings of the 55th Annual Meeting of the Association
  for Computational Linguistics (Volume 1: Long Papers)}, pages 1073--1083,
  Association for Computational Linguistics, Vancouver, Canada.

\bibitem[{Shuster et~al.(2020)Shuster, Ju, Roller, Dinan, Boureau, and
  Weston}]{dodeca}
Shuster, Kurt, Da~Ju, Stephen Roller, Emily Dinan, Y-Lan Boureau, and Jason
  Weston. 2020.
\newblock The dialogue dodecathlon: Open-domain knowledge and image grounded
  conversational agents.
\newblock In \emph{Proceedings of ACL}.

\bibitem[{Thorne et~al.(2021)Thorne, Glockner, Vallejo, Vlachos, and
  Gurevych}]{DBLP:journals/corr/abs-2104-00640}
Thorne, James, Max Glockner, Gisela Vallejo, Andreas Vlachos, and Iryna
  Gurevych. 2021.
\newblock Evidence-based verification for real world information needs.
\newblock \emph{CoRR}, abs/2104.00640.

\bibitem[{Thorne and Vlachos(2018)}]{Thorne2018AutomatedFC}
Thorne, James and Andreas Vlachos. 2018.
\newblock Automated fact checking: Task formulations, methods and future
  directions.
\newblock In \emph{Proceedings of COLING}.

\bibitem[{Thorne et~al.(2018)Thorne, Vlachos, Christodoulopoulos, and
  Mittal}]{thorne-etal-2018-fever}
Thorne, James, Andreas Vlachos, Christos Christodoulopoulos, and Arpit Mittal.
  2018.
\newblock {FEVER}: a large-scale dataset for fact extraction and
  {VER}ification.
\newblock In \emph{Proceedings of NAACL}.

\bibitem[{Wang, Cho, and Lewis(2020)}]{wang-etal-2020-asking}
Wang, Alex, Kyunghyun Cho, and Mike Lewis. 2020.
\newblock Asking and answering questions to evaluate the factual consistency of
  summaries.
\newblock In \emph{Proceedings of ACL}.

\bibitem[{Welleck et~al.(2019)Welleck, Weston, Szlam, and
  Cho}]{Welleck2019DialogueNL}
Welleck, Sean, Jason Weston, Arthur~D. Szlam, and Kyunghyun Cho. 2019.
\newblock Dialogue natural language inference.
\newblock In \emph{Proceedings of ACL}.

\bibitem[{Wilson and Sperber(2004)}]{wilson2002relevance}
Wilson, Deirdre and Dan Sperber. 2004.
\newblock \emph{Relevance Theory}, The Handbook of Pragmatics. Blackwell.

\bibitem[{Wiseman, Shieber, and Rush(2017)}]{wiseman-etal-2017}
Wiseman, Sam, Stuart Shieber, and Alexander Rush. 2017.
\newblock Challenges in data-to-document generation.
\newblock In \emph{Proceedings of EMNLP}.

\bibitem[{Zaheer et~al.(2020)Zaheer, Guruganesh, Dubey, Ainslie, Alberti,
  Ontanon, Pham, Ravula, Wang, Yang, and Ahmed}]{bigbird}
Zaheer, Manzil, Guru Guruganesh, Kumar~Avinava Dubey, Joshua Ainslie, Chris
  Alberti, Santiago Ontanon, Philip Pham, Anirudh Ravula, Qifan Wang, Li~Yang,
  and Amr Ahmed. 2020.
\newblock Big bird: Transformers for longer sequences.
\newblock In \emph{Advances in Neural Information Processing Systems},
  volume~33, pages 17283--17297, Curran Associates, Inc.

\bibitem[{Zhong et~al.(2020)Zhong, Liu, Chen, Wang, Qiu, and Huang}]{matchsum}
Zhong, Ming, Pengfei Liu, Yiran Chen, Danqing Wang, Xipeng Qiu, and Xuanjing
  Huang. 2020.
\newblock Extractive summarization as text matching.
\newblock In \emph{Proceedings of the 58th Annual Meeting of the Association
  for Computational Linguistics}, pages 6197--6208, Association for
  Computational Linguistics, Online.

\end{thebibliography}
\end{document}